\documentclass[preprint,12pt]{elsarticle}

\usepackage{amssymb}
\usepackage{amsmath,epsfig}
\usepackage{array}
\usepackage{gensymb}
\usepackage[latin1]{inputenc}
\usepackage{subfigure}
\usepackage{epsfig}
\usepackage{multirow}
\usepackage{makecell}
\usepackage{amssymb,amsfonts,amsmath}
\usepackage{graphicx}
\usepackage{float}
\usepackage{algorithm}
\usepackage{algorithmic}
\usepackage{array}
\usepackage{MnSymbol}
\usepackage[modulo]{lineno}
\usepackage[table,xcdraw]{xcolor}
\usepackage{color}
\usepackage{booktabs}
\usepackage{hyperref}
\usepackage[final]{pdfpages}
\usepackage{graphicx}
\usepackage{diagbox}
\usepackage{multirow}
\usepackage[edges]{forest}
\usepackage{enumitem}
\useforestlibrary{linguistics}
\forestapplylibrarydefaults{linguistics}

\DeclareMathOperator*{\argmax}{arg\,max}

\graphicspath{{figures/}}

\usepackage{array}
\newcommand{\PreserveBackslash}[1]{\let\temp=\\#1\let\\=\temp}
\newcolumntype{C}[1]{>{\PreserveBackslash\centering}p{#1}}
\newcolumntype{R}[1]{>{\PreserveBackslash\raggedleft}p{#1}}
\newcolumntype{L}[1]{>{\PreserveBackslash\raggedright}p{#1}}

\journal{Elsevier journal}

\begin{document}

\begin{frontmatter}

\title{Randomization-based Machine Learning in Renewable Energy Prediction Problems: Critical Literature Review, New Results and Perspectives}

\author[javi1,javi2]{J. Del Ser\corref{cor1}}
\author[antonio]{D. Casillas-Perez}
\author[antonio]{L. Cornejo-Bueno}
\author[luis]{L. Prieto-Godino}
\author[julia]{J.~Sanz-Justo}
\author[carlos]{C. Casanova-Mateo}
\author[sancho]{S. Salcedo-Sanz} \cortext[cor1]{Corresponding author: Javier Del Ser. ICT Division, TECNALIA, Basque Research and Technology Alliance (BRTA), 48160 Derio, Bizkaia, Spain. Ph: +34 94 643 08 50. \ead{javier.delser@tecnalia.com}}

\address[javi1]{TECNALIA, Basque Research and Technology Alliance (BRTA), \\48160 Derio, Bizkaia, Spain}
\address[javi2]{University of the Basque Country (UPV/EHU), 48013 Bilbao, Bizkaia, Spain}
\address[antonio]{Dept. of Signal Processing and Communications, Universidad Rey Juan Carlos, Fuenlabrada, Madrid, Spain.}
\address[sancho]{Dept. of Signal Processing and Communications, Universidad de Alcal\'a, Alcal\'a de Henares, Madrid, Spain}
\address[luis]{Dept. of Technical Services, Iberdrola Renewables, Madrid, Spain}
\address[julia]{LATUV, Laboratorio de Teledetecci\'on, Universidad de Valladolid, Spain}
\address[carlos]{Dept. of Civil Engineering: Construction, Infrastructure and Transport, Univ. Polit\'ecnica de Madrid, Madrid, Spain}

\begin{abstract}
Randomization-based Machine Learning methods for prediction are currently a hot topic in Artificial Intelligence, due to their excellent performance in many prediction problems, with a bounded computation time. The application of randomization-based approaches to renewable energy prediction problems has been massive in the last few years, including many different types of randomization-based approaches, their hybridization with other techniques and also the description of new versions of classical randomization-based algorithms, including deep and ensemble approaches. In this paper we review the most important characteristics of randomization-based machine learning approaches and their application to renewable energy prediction problems. We describe the most important methods and algorithms of this family of modeling methods, and perform a critical literature review, examining prediction problems related to solar, wind, marine/ocean and hydro-power renewable sources. We support our critical analysis with an extensive experimental study, comprising real-world problems related to solar, wind and hydro-power energy, where randomization-based algorithms are found to achieve superior results at a significantly lower computational cost than other modeling counterparts. We end our survey with a prospect of the most important challenges and research directions that remain open this field, along with an outlook motivating further research efforts in this exciting research field.
\end{abstract}


\end{frontmatter}

\section{Introduction}
\label{sec:intro}

Renewable energies are clean and inexhaustible sources of energy, diverse and abundant enough for their potential use on the entire planet \cite{CHEN2020109985}. Renewable energy sources do not produce greenhouse gases (associated with climate change and global warming) nor polluting emissions, harmful for humans' and animals' health. Renewable resources are increasingly competitive: the costs of renewable resources are strongly falling at a sustainable rate (the cost of solar photovoltaics has fallen over 80\% in the last 10 years, and the onshore wind over 40\% \cite{Ren21}), whereas the general cost trend for fossil fuels is in the opposite direction, in spite of their strong volatility. These well-known strong advantages collide with the most important issue associated with renewable energy sources: their intrinsic intermittency, which hinders their inclusion in the energetic mix over a certain limit. 

Nowadays, the best way of dealing with renewable energy intermittency is to carry out accurate predictions of the renewable production, at different prediction-time horizon, from short-term to medium-term, in order to  make renewable sources compatible with the power grid \cite{IMPRAM2020100539}. Therefore, the literature devoted to renewable energy prediction approaches has been huge in the last few years, including different big review papers on general approaches for prediction in renewable sources \cite{kariniotakis2017renewable,perez2016review,tsai2017models}, and also specific prediction problems and algorithms for wind energy \cite{giebel2003state,mohandes2004support,ortiz2011short,salcedo2011short}, solar energy \cite{zeng2013short,li2019renewable} or marine and ocean energy \cite{cornett2008global,cuadra2016computational,cornejo2016significant}. 

Many of these previous approaches dealing with renewable energy prediction problems discuss Machine Learning (ML) and related methods for prediction \cite{kalogirou2001artificial,salcedo2018feature,wang2019review,lai2020survey}. ML methods are currently a hot spot in renewable energy prediction problems, with hundreds of new algorithmic proposals and real-world applications \cite{perera2014machine}, including well-established artificial intelligence methods \cite{zahraee2016application}, hybridization with numerical methods \cite{salcedo2009hybridizing}, or novel trends in the field such as deep learning \cite{wang2019review}, among others.

In the last few years, there have been a massive development of certain ML methods for prediction, which share some specific characteristics of random initialization of part of their structure or parameters. They are known as {\em randomization-based ML approaches}, and include different techniques such as different kind of neural networks (Extreme Learning Machines, Random Vector Functional Link networks), reservoir computing, different ensemble methods, many of them with multi-layer and deep versions, to improve their performance in hard prediction problems. Randomization-based approaches have experimented a massive grow in the last years, mainly due to their excellent performance in many prediction problems, together with extremely fast training times, due to their random initialization and simple training schemes. The application of randomization-based algorithms to renewable energy has also been extremely important, mainly in the last 5 years, where the main body of applications can be found. 

In this paper we review the most important randomization-based ML approaches in the frame of renewable energy prediction problems. We describe the most important characteristics of these methods, and provide a comprehensive literature review of their application in prediction problems dealing with renewable energy sources of diverse nature, such as solar, wind, marine and ocean and hydro-power. Finally, we also show some specific case studies in solar energy, wind energy and hydro-power prediction problems, where randomization-based approaches are shown to outperform previous approaches in the literature, obtaining extremely competitive results within a very reduced computation time. 

The rest of the paper has been structured in the following way:  Section \ref{Rand_methods} describe the most important characteristics of randomization-based ML methods, with details on their structure, training algorithms and characteristics. Section \ref{Literature_Review} provides a comprehensive review of the application of randomization-based ML approaches to different problems in renewable energy prediction problems. Section \ref{Experiments_Results} presents some experimental results on the application of randomization-based ML algorithms to different prediction problems in solar radiation prediction, wind speed prediction and water level prediction in reservoirs for hydro-power applications. In all these problems we have applied a large battery of randomization-based approaches and compare their results with the state-of-the-art algorithms for each specific problem, and some alternative classical ML approaches which have obtained good results in these problems in the past. Finally, Section \ref{Challenges} gives some concluding remarks for this research, including the discussion of the most important challenges and research directions in this area.

\section{Randomization-based ML methods}\label{Rand_methods}

Randomization-based ML methods comprise those ML algorithms which include the randomization of some of their parts along their training process, in order to improve their accuracy, avoid overfitting or make them robust against outlier data. This randomization can be applied to different parts of the methods: in some cases, the randomization is applied to the selection of the training set as a means to induce controlled diversity in the input data. Ensemble methods based on bootstrap aggregating are a clear example of this kind of randomization-based ML methods. In other cases, the randomization is applied to some parts of the topology or structure of the ML method. For instance, Echo State Networks (ESN) randomly choose the links which form their recurrent neural topology. Finally, methods such as Extreme Learning Machines (ELM) and Random Vector Functional Link (RVFL) networks initialize at random the values of their weights, which can be conceived as another way to exploit randomization in the construction of data-based models. In the next subsection we described in detail different randomization-based ML algorithms, giving some key points of their structure and capabilities for obtaining good results in classification and regression problems in renewable energy. For extended general information on randomization-based ML methods we refer to the comprehensive reviews reported in \cite{gallicchio2017randomized}, \cite{zhang2016survey} and more recently, in \cite{gonzalez2020practical} and \cite{suganthan2021origins}.

\subsection{Ensemble methods}

Ensemble methods have the particularity of improving the predictive performance of a single learning model, based on the randomized combination of different training models \cite{zhou2012ensemble}.
This learning paradigm assumes that combinations of several base ML models can improve the prediction performance and overcome the robustness or generalization capacity of complex ML, such as the neural networks, which involve a huge number of parameters.
Ensemble methods support their efficiency delegating the prediction in simpler but coordinated model, the called learners.
Training a set of simpler models bring some important advantages, as quicker, against complex ones.
In recent years, many of them have been developed, however, bagging and boosting techniques have become particularly interested in ML problems \cite{gonzalez2020practical}, leading to very powerful and used algorithmic versions for classification and regression problems.

\subsubsection{Bagging}\label{sec:bagging}
\begin{figure}[!ht]
    \centering
    \includegraphics[draft=false, angle=0,width=10cm]{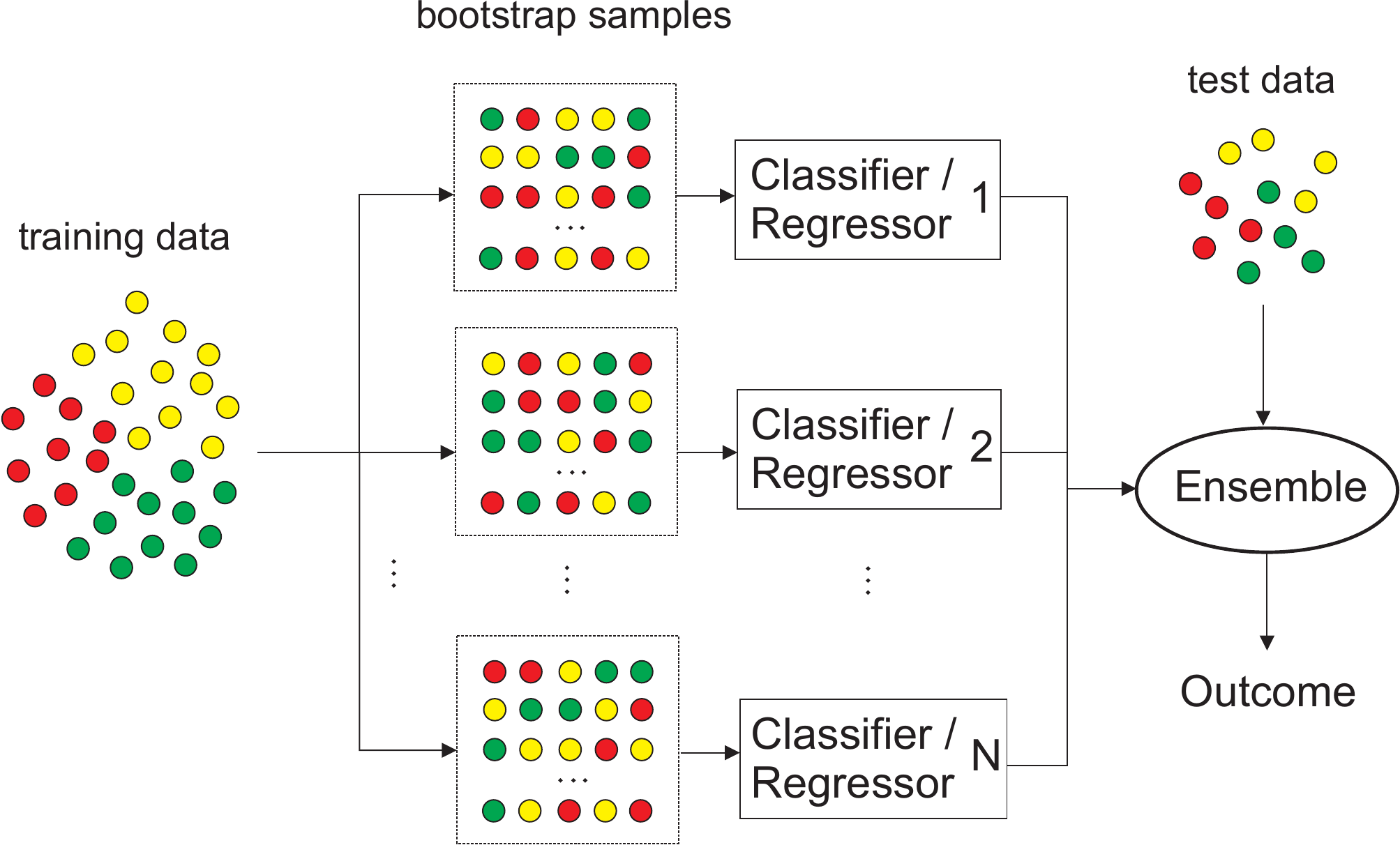}
    \caption{ \label{Bagging_Ejemplo} Diagram of a general bagging technique for classification or regression problems in ML.}
\end{figure}

We first start with the description of the ensemble technique known as \emph{bootstrap aggregating} or just {\em bagging}. The basic idea behind bagging is to train a set of simple models and combine their individual predictions as shown in Figure \ref{Bagging_Ejemplo}, in such a way that the global stability and accuracy of the ensemble surpass the individually obtained by the ML approaches which compose it. 
Bagging also reduces variance of the ML performance techniques and helps avoid overfitting, which is usually more severe in complex ML methods.
Bagging establishes that all the base ML models which composed the ensemble has the same architecture, which result in same topology, number of input-output variables and number of parameters to train. As an example, a set of decision trees trained with the bagging technique assume that all trees have same branches, with same parameters to train and same input-output variables, see Figure \ref{Bagging_Ejemplo}. The individual models of the ensemble defers in the parameters, which are trained with different training sets.

The mathematical description of the bagging technique can be done as follows: Let $\mathcal{D}=\lbrace(\mathbf{x}_i,y_i)\rbrace_{i=1}^n$ be a given training set of $n$ input-output pairs.
The procedure of bagging, shown in Figure \ref{Bagging_Ejemplo}, generates $N$ new training subsets $\mathcal{D}_{i}$,  each of size $n'$, by sampling from $\mathcal{D}$ uniformly and with replacement (note that some observations may be repeated in each $\mathcal{D}_{i}$). This kind of sampling is known as a {\em bootstrap sample}. Then, the parameters of $N$ equal models $\lbrace\mathcal{M}_i\rbrace_{i=1}^N$ are discovered training each model with these subset $\mathcal{D}_{i}$ respectively.
Finally, the ensemble model provides the output applying a decision rule which combines the individual outputs of each model: averaging their outputs (in the case of regression problems) or by voting (if dealing with classification problems) \cite{mohandes2018classifiers}.

Bagging models can be deemed the simplest way to create ensembles. Note that the base ML models are trained independently with no influence among them. This property allows to train each in parallel which drastically reduce the training time of this ensemble model.

Random Forest (RF) \cite{breiman2001random} is one of the most popular bagging-like techniques for classification and regression problems. It specifically uses decision or regression trees as learners, and it has been successfully applied to a large number of real problems and applications. It differs from the pure bagging technique in that the topology of the trees changes among them. Trees of the ensemble (the forest) may have different length, topology or use different input variables which greatly increase the variability of the learners but is contrary to the bagging paradigm from theoretical view.

Its main advantage lies in its generalisation capacity, achieved by compensating the errors obtained from the predictions of the different decision trees. Once the decision or regression trees have been generated, and each has obtained its classification or regression solution, a voting or averaging scheme is taken into account to obtain the final prediction result \cite{breiman2001random}.

The training procedure carried out takes into account the following steps. Given a set of training data $\lbrace(\textbf{x}_i,{y}_i)\rbrace_{i=1}^n$, the main parameters to be adjusted are: number of estimators (number of trees in the forest) - $N$, and maximum number of features to be considered when splitting a node - $maxDepth$ (being recommended as the square root of the total number of features). Once we have set these parameters:
\begin{enumerate}[leftmargin=*]
\item We initialize each one of the $N$ decision or regression trees for the classification or regression problem respectively.
\item For each tree $\textbf{T}_t$, we select $n_t$ samples with replacement, by using the bootstrapping technique.
\item Only a subset of maximum $maxDepth$ features shall be considered for the construction of each trees.
\item Each tree $\textbf{T}_t$ will obtain a solution.
\item The ensemble output of the random forest method will be computed by majority voting  in the case of classification:
\begin{equation}
    \hat{Y}(\mathbf{x}) = \argmax_l\sum_{t=1}^N[\textbf{T}_t(\mathbf{x})=l].
\end{equation}
or averaging for regression problems: 
\begin{equation}
    \hat{Y}(\mathbf{x}) = \frac{1}{N}\sum_{t=1}^N\alpha_t \textbf{T}_t(\mathbf{x}).
\end{equation}
\end{enumerate}

\subsubsection{Boosting}

\begin{figure}[!ht]
    \centering
    \includegraphics[draft=false, angle=0,width=10cm]{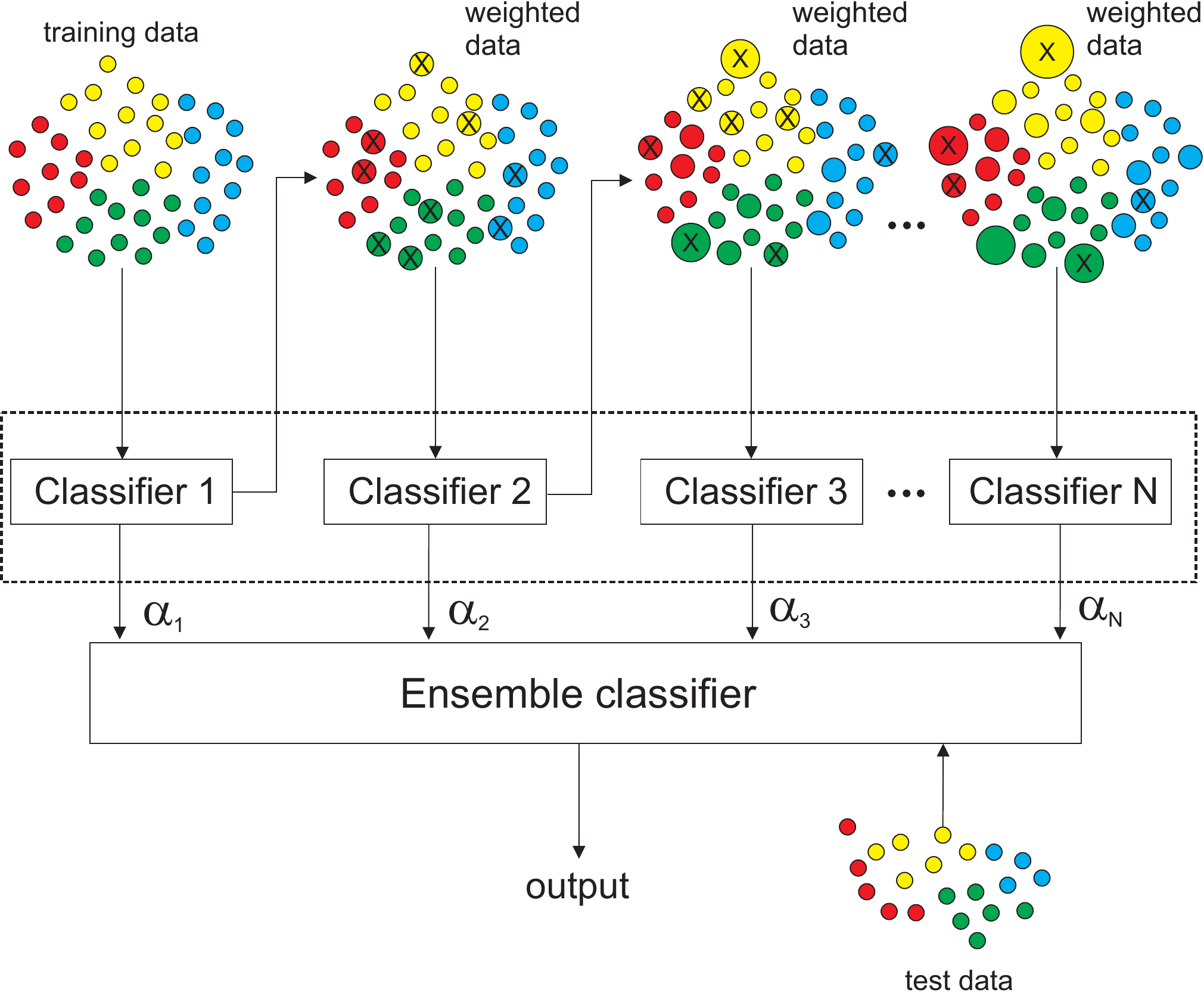}
    \caption{Diagram of the AdaBoost algorithm exemplified for multi-class classification problems. Different size circles stand for samples with more associated weight (w) due to misclassification in the previous step (marked with X).}
    \label{Boosting_Ejemplo} 
\end{figure}

Boosting approaches are an alternative family of ensemble algorithms which has obtained excellent performance in both classification and regression problems \cite{ferreira2012boosting}. As bagging, boosting follows the learning paradigm of using simple or weak ML models (classifiers/regressors), named as learners, to form a powerful final approach properly combining their outputs. Boosting also establishes the same topology for all the learners involved in the ensemble (same architecture, number of input-output variables and number of parameters to train).
However, there exist important differences. The most evident difference is located on the procedure for training the weak learners. In bagging, the weak learners are trained in parallel using different subsets $\mathcal{D}_i$ randomly sampled from the whole training dateset $\mathcal{D}$ as explained in the previous section~\ref{sec:bagging}. In boosting, the learners are trained sequentially, in such a way that each new learner requires that the previous learner had been trained before, see Figure~\ref{Boosting_Ejemplo}. In this way, learners are dependent among them, contrary to the bagging method. In boosting, all the learners use the whole set of training dataset for computing their parameters, i.e, there is no bootstrap sample step.

Other important difference is that in bagging, all input-output pairs are equally weighted to train each learner and also each learner equally contributes to determine the final output the ensemble model.
In boosting, training input-output pairs are weighting according to the accuracy for being predicted by the previous learner (except for the first learner in the queue which uses the equally weighted samples). Consequently, learners are more specialized as soon as they are placed into the final locations along the queue. Besides, the contribution of each learner to the output of the ensemble is usually weighed according to its accuracy, which does not happen in the bagging.
This is the general scheme for all boosting methods. There exist different boosting strategies which concern with the kind of weighting policy that methods apply to each training sample, and/or the output of the each learner. 

One of the most widely used boosting techniques is the Adaptive Boosting (AdaBoost) algorithm. AdaBoost proposes to train each of these machines iteratively, in such a way that each base learner focuses on the data that was misclassified by its predecessor, to iteratively adapt its parameters and achieve better results \cite{ferreira2012boosting,gonzalez2020practical}. We can find multiple variants of the AdaBoost algorithm, starting from the original one \cite{freund1997decision} designed to tackle binary classification problems, regression or multi-class classification options. Figure \ref{Boosting_Ejemplo} shows an outline of the Adaboost algorithm for multi-class classification. In the following lines, the AdaBoost pseudocode is explained:
\begin{enumerate}[leftmargin=*]
\item  Given a set of training data $\mathcal{D}=\lbrace(\textbf{x}_i,y_i)\rbrace_{i=1}^n$, we proceed with the initialisation of each base learner $\lbrace \textbf{T}_t\;|\;1\leq t \leq N\rbrace$, and assigning the set of sample weights $\lbrace {w}_i\;|\;1\leq i \leq n\rbrace$ corresponding to the input-output pairs $\lbrace(\textbf{x}_i,y_i)\rbrace_{i=1}^n$ according to the uniform distribution: ${w}_i = \frac{1}{n}$.
\item  For each base learner $\textbf{T}_t$, we use the training dataset with the distribution of weights ${w}_i$ for training it.
\item  After this training process, for each base learner $\textbf{T}_t$, we calculate the estimation error $\epsilon_t$ computed as: \begin{equation}
    \epsilon_t=\sum_{\textbf{T}_t(\textbf{x}_i)\neq {y}_i} \frac{w_i}{\sum_{\textbf{x}_i} w_i},\quad 1\leq i\leq n
\end{equation}
\item  Based on this error we obtain the weight of the current base learner to the ensemble output $\alpha_t$:
\begin{equation}
    \alpha_t = \log \frac{1-\epsilon _t}{\epsilon _t}
\end{equation}
\item  Finally, the distribution of the weights ${w}_i$ corresponding to each $\textbf{x}_i$, which will be used the next learner, is proportionally adjusted to the probability that a sample is correctly estimated, and inversely proportional to the error of the learner $\epsilon_t$.
\item The final output, provided by the algorithm globally, will be: 
\begin{equation}
    \hat{Y}(\mathbf{x}) = \argmax_l\sum_{t=1}^N[\alpha_t \cdot (\textbf{T}_t(\mathbf{x})=l)].
\end{equation}
This final function refers to the boosting method for classification problems, which simply integrates the weighted output of individual learners by voting. In regression problems, the output consists on computing a weighted average of the outputs:
\begin{equation}
    \hat{Y}(\mathbf{x}) = \frac{1}{N}\sum_{t=1}^N\alpha_t \textbf{T}_t(\mathbf{x}).
\end{equation}
\end{enumerate}
The main difference of this algorithm with the multi-class variant AdaBoost.M1 \cite{freund1997decision}, is that only the weights of the correctly classified samples are decreased ($w_i = w_i \frac{\epsilon _t}{1-\epsilon _t}$).

\subsection{Extreme Learning Machines (ELM)}\label{sec.ELM}
\begin{figure}[!ht]
    \centering
    \includegraphics[width=0.7\columnwidth]{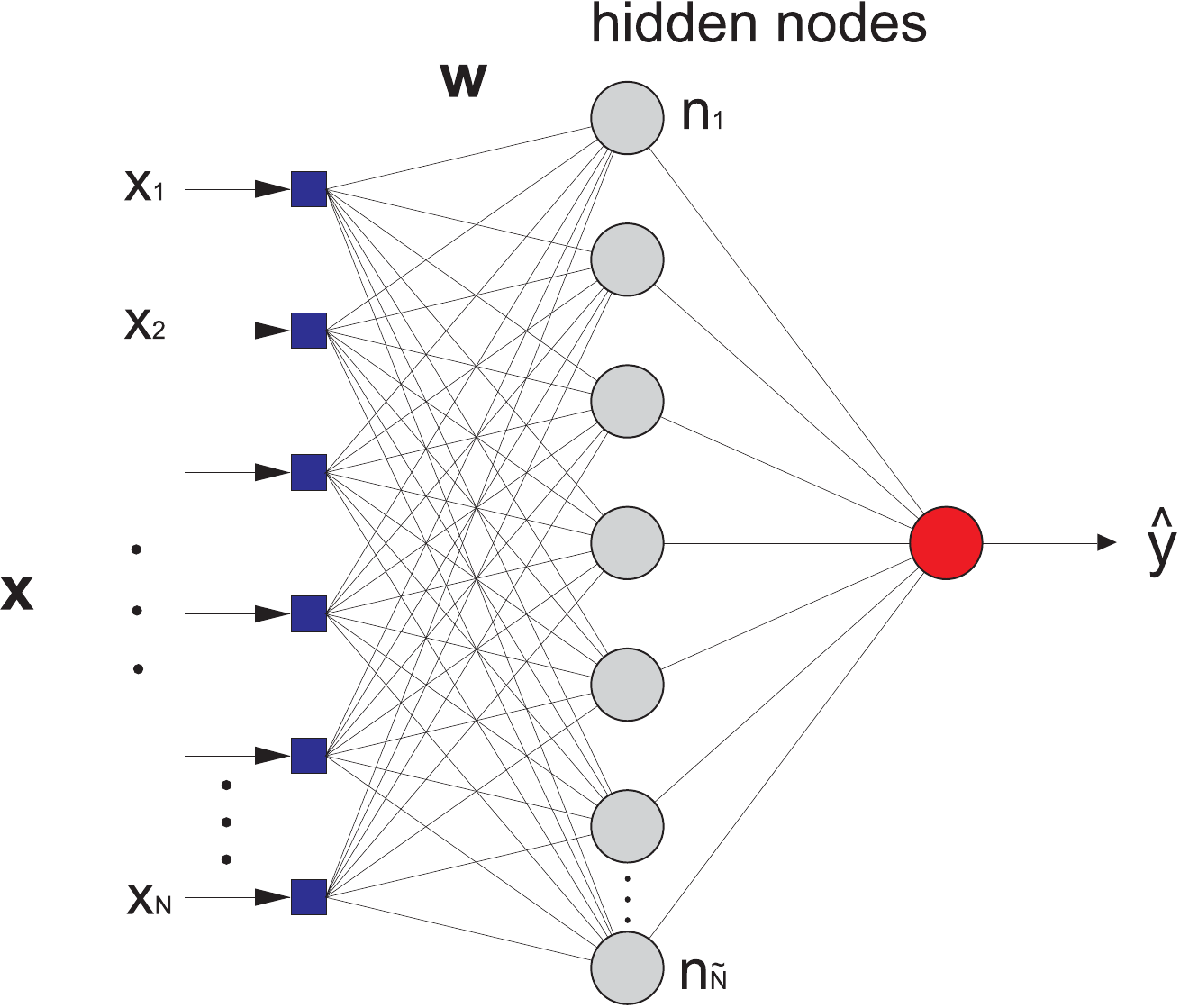}
    \caption{ \label{ELM_Ejemplo} Multi-layer perceptron structure considered in the ELM algorithm.}
\end{figure}

An extreme-learning machine \cite{Huang06} is a fast training method mainly used for feed-forward multi-layer perceptron structures, see Figure \ref{ELM_Ejemplo}. In the ELM algorithm the network weights of the first layer are randomly set, usually using an uniform probability distribution. Then, it establishes the output matrix of the hidden layer and computes the Moore-Penrose pseudo-inverse of this matrix. The optimal weights of the output layer are directly obtained by multiplying the computed pseudo-inverse matrix with the target, that is, the weights of the output layer which fit best with the objective values (see \cite{Huang12} for details). This method obtains competitive results with respect to other classical training methods, and the training computation efficiency overcomes multi-layer perceptrons (MLPs), or even Support Vector Machine algorithms \cite{Huang12}.

Mathematically, the ELM algorithm considers a training set
$\lbrace(\textbf{x}_i,y_i)\rbrace_{i=1}^n$
to fit the weights $(\beta_k)$ associated to
each hidden nodes $\tilde{N}$ to optimally compose an output which minimum mean squared error. It applies the following steps:
\begin{enumerate}[leftmargin=*]
\item Randomly assign input weights $\textbf{w}_k$ and the bias $b_k$, where $k = 1, \ldots ,\tilde{N}$, using a uniform probability distribution in $[-1,1]$.
\item Calculate the hidden-layer output matrix $H$, defined as follows:
\begin{equation}
\textbf{H} = \left[\begin{array}{ccc}
g( \textbf{w}_1 \textbf{x}_1 + b_1) & \cdots & g(\textbf{w}_{\tilde{N}} \textbf{x}_1 + b_{\tilde{N}}) \\
\vdots & \cdots & \vdots \\
g(\textbf{w}_1 \textbf{x}_N + b_1) & \cdots & g(\textbf{w}_{\tilde{N}} \textbf{x}_N + b_{\tilde{N}})
\end{array}
\right]_{\tilde{N}}
\end{equation}
where $g(\cdot)$ is an activation function.
\item The training problem is reduced to a $\boldsymbol\beta$ parameter optimization problem, which can be defined as:
\begin{equation}\label{eq.opt.ELM}
\min\limits_{\boldsymbol\beta} \lVert \textbf{H} \boldsymbol\beta-\textbf{Y}\rVert,
\end{equation}
\item Finally, calculate the output weight vector $\boldsymbol\beta$ as follows:
\begin{equation}\label{eq.elm}
    \boldsymbol\beta = \textbf{H}^\dagger \textbf{Y}^T,
\end{equation}
where $\textbf{H}^\dagger$ is the Moore-Penrose inverse of the matrix $\textbf{H}$ \cite{Huang06}, and $\textbf{Y}^T$ is the transpose training output vector, $\textbf{Y}=[y_1,\ldots,y_n]$.
\item Then, the predicted or classified output is obtained as: $\hat{Y}(\mathbf{x}) = \textbf{H} \boldsymbol\beta$.
\end{enumerate}

Note that the number of hidden nodes $\tilde{N}$ is a free parameter to be set before the training of the ELM algorithm, and must be estimated for obtaining good results by scanning a range of $\tilde{N}$ in a validation phase. 

\subsection{Multihidden-Layer ELMs (ML-ELM)}
One of the main problems we can find in Multilayer Neural Networks (ML-ELM) is the poor performance they present when trained with the Backpropagation algorithm, mainly caused by the vanishing gradient problem. In contrast to deep networks of this type, the ML-ELM algorithm does not require iterative adjustment of the weights of the hidden layers, and the construction of this algorithm is simpler \cite{xiao2017multiple}. First, given a training set $\lbrace(\textbf{x}_i,y_i)\rbrace_{i=1}^n$, the number of hidden layers $\tilde{N}$ and the activation function $g$, the learning process will be similar to the one explained in the section \ref{sec.ELM}, but with some differences:
\begin{enumerate}[leftmargin=*]
\item Randomly assign input weights $\textbf{w}_k$ and the bias $b_k$, where $k = 1, \ldots ,\tilde{N}$, between the input and the first hidden layer.
\item Calculate the first hidden-layer output matrix $\textbf{H}$, defined as follows: $\textbf{H} = g(\textbf{w}_{ki}\textbf{x}_i + b_k)$.
\item Calculate the output weights $\boldsymbol\beta$ between the hidden layer and output layer as: $\boldsymbol\beta = \textbf{H}^\dagger \textbf{Y}^T$ (these weights will be updated as we obtain the outputs of the rest of the layers, as indicated in the following points).
\item Calculate the expected output of the second hidden layer: $\textbf{H}_1 = \textbf{Y}\boldsymbol\beta$ , where the matrix $\textbf{Y}$ is the training output vector.
\item Then, we have to calculate the weights ($\textbf{w}_{HE}$) between the first hidden layer and the second hidden layer, and the bias of the second hidden neurons ($b_1$), using the previous calculation of $\textbf{H}_1$ \cite{xiao2017multiple}.
\item Now, we can update the actual output of the second hidden layer: $\textbf{H}_2 = g(\textbf{w}_{HE}\textbf{H}_E + b_1)$.
\item And also update the weights ($\boldsymbol\beta$) between the previous hidden layer ($\textbf{H}_2$) and the output layer: $\boldsymbol\beta _{new} = \textbf{H}_2^\dagger \textbf{Y}^T$.
\item If the number of hidden layers is $K$, we have to execute steps 4 to 7 by $(K-1)$ times.
\item Finally, we can obtain the output of the algorithm as: $\hat{Y}(\mathbf{x}) = \textbf{H}_k \boldsymbol\beta$.
\end{enumerate}

\subsection{Random Vector Functional Link Networks (RVFL)}\label{sec.rvlf}
\begin{figure}[!ht]
    \centering
    \includegraphics[width=0.8\columnwidth]{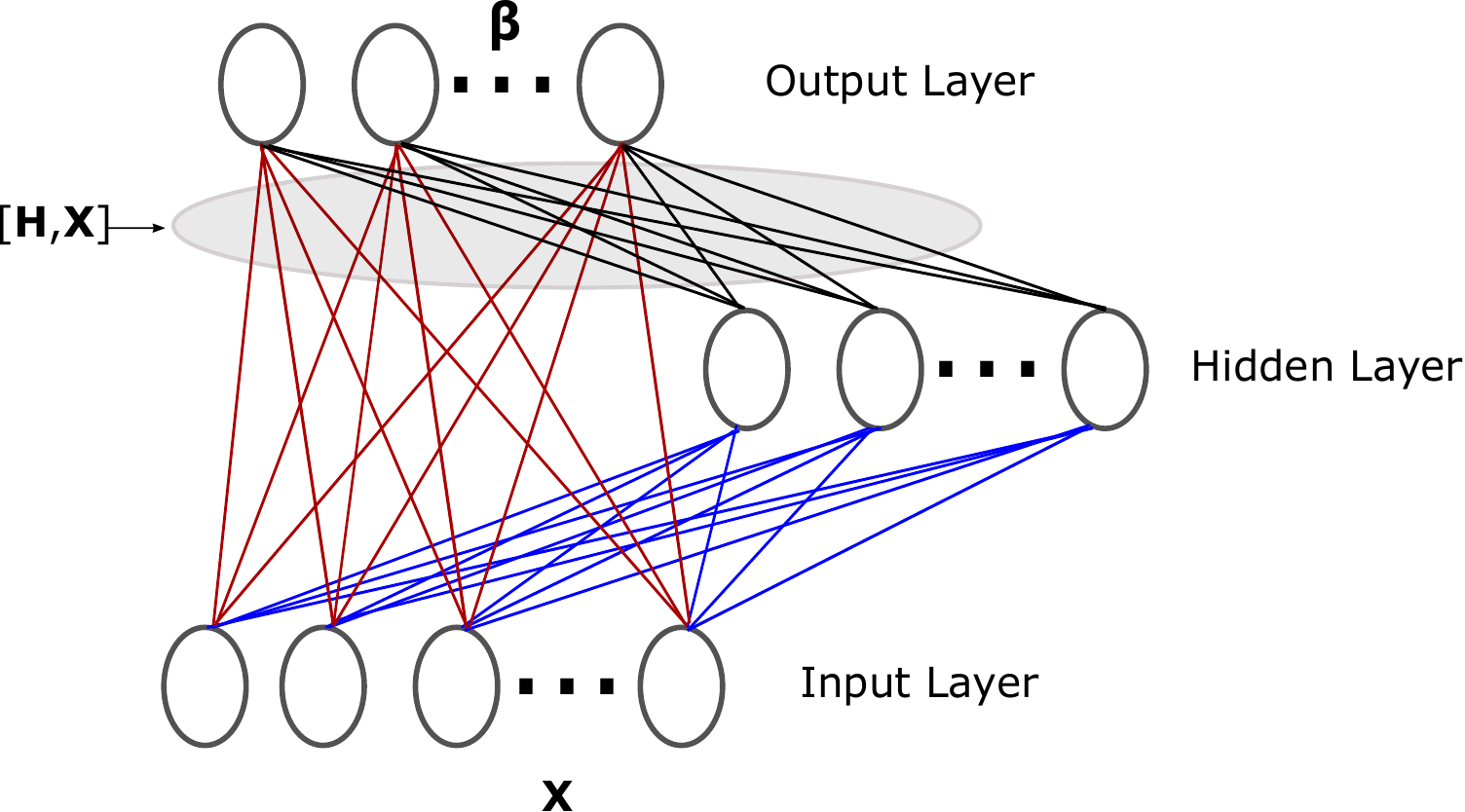}
    \caption{Exemplifying diagram of a Random Vector Functional Link Network.}
    \label{RVFL} 
\end{figure}

Random Vector Functional Link Networks (RVFL) \cite{katuwal2019random} are based on the structure of SLFNs, with the particularity that the weights and biases of the neurons in the hidden layer are initialised randomly, and their values kept fixed throughout the training phase. Only the weights of the output layer need to be updated to achieve the lowest estimation error. As in the previous sections, we are going to delve a little deeper into the construction of this type of algorithms. Similar to the operation of ELMs, RVFL does not require iterative adjustment of the hidden layer weights, but the main difference with respect to the ELM algorithm is in the way we obtain the final output. Figure \ref{RVFL} shows the RVFL architecture.
Given a training set $\lbrace(\textbf{x}_i,y_i)\rbrace_{i=1}^n$, and $\tilde{N}$ hidden nodes in the hidden layer:
\begin{enumerate}[leftmargin=*]
\item Randomly assign input weights $\textbf{w}_k$ and the bias $b_k$, where $k = 1, \ldots ,\tilde{N}$, between the input and the hidden layer.
\item Calculate the hidden-layer output matrix $\textbf{H}$, which consists of a non-linear transformation of the input features.
\item The inputs of the output layer will consist of the matrix $\textbf{H}$ as well as the original features $\textbf{x}_i$. Therefore if $n$ is the total number of input features and $\tilde{N}$ the number of neurons in the hidden layer, we will have a total of $n+\tilde{N}$ inputs for each output node (the main difference we mentioned above with respect to the ELM algorithm).
\item During the training process, the parameters of the hidden layer kept fixed, so our work will mainly focus on obtaining the output weights ($\boldsymbol\beta$). For this we have to solve an optimization problem given by the following expression:
\begin{equation}\label{eq.opt.RVFL}
\min\limits_{\boldsymbol\beta} ||\textbf{P} \boldsymbol\beta-\textbf{Y}||^{2} + \lambda ||{\boldsymbol\beta}||^{2},
\end{equation}
where $\textbf{P} = [\textbf{H};\textbf{X}]$ is the concatenation of the matrix $\textbf{H}$ and the input training-set features $\textbf{X}$, and $\textbf{Y}$ is the training output vector. 
\item Equation \eqref{eq.opt.RVFL} can be solved using ridge regression or Moore-Penrose pseudoinverse (for classification). Using Moore-Penrose pseudo-inverse, the solution is obtained by: $\boldsymbol\beta = \textbf{P}^\dagger \textbf{Y}$, formula that we always use in the ELM algorithm to obtain the solution, see equation \eqref{eq.elm}. And to solve the equation using the regularized least squares (or ridge regression), the solution is given by equations \eqref{eq.primal}  \cite{suganthan2021origins}:
\begin{equation}\label{eq.primal}
 \boldsymbol\beta = (\textbf{P}^T\textbf{P}+\lambda \textbf{I})^{-1}\textbf{P}^T\textbf{Y} .
\end{equation}

\item The final output is given by: $\hat{Y}(\mathbf{x}) = \textbf{P} \boldsymbol\beta$.

\end{enumerate}

At this point, it is straightforward to see that there are small differences between ELMs and RVFLs algorithms. Both methods have a similar internal structure, except that in the case of ELMs the direct links between input and output are omitted. This fact has been at the core of a controversial debate about the novelty and pioneering contribution of these techniques \cite{wang2008comments,huang2008reply}. In fact, both methods are interesting in terms of learning capability. Thanks to their good performance in different problems, it is only a matter of deciding which of the two are better suited for an specific task. ELMs, for instance, have shown an extremely good performance to construct hybrid approaches by merging them with evolutionary computation techniques, due to their extremely good computational overhead.

\subsection{Deep RVFL (dRVFL)}
The Deep Random Vector Functional Link (dRVFL) network is an extension of the RVFL network \cite{katuwal2019random}. This type of network is characterized by a set of hidden layers, where the input to each layer is formed by the output of the predecessor layer. In each of the hidden layers, an internal representation of the input data will be carried out. These structures present great flexibility in the design, being able to fix the size of the network, as well as the number of neurons in each layer with the value we want. For the sake of simplicity when explaining how the algorithm works, we will consider $K$ hidden layers, which will contain the same number of neurons $\tilde{N}$. Following the procedure in the section \ref{sec.rvlf}, and given a training set $\lbrace(\textbf{x}_i,y_i)\rbrace_{i=1}^n$ as well as the number of layers, $K$ and the number of neurons in each layer $\tilde{N}$:
\begin{enumerate}[leftmargin=*]
\item Randomly assign input weights $\textbf{W}^{(1)},\ldots, \textbf{W}^{(K)}$ and biases $\textbf{B}^{(1)}, \ldots, \textbf{B}^{(K)}$, between the input and the first hidden layer, and inter hidden layers (for simplicity of notation we will omit the bias of each layer). These parameters are kept fixed during the training.
\item Calculate the first hidden-layer output matrix $\textbf{H}^{(1)}$, which consists of a non-linear transformation of the input features, as: $\textbf{H}^{(1)} = g(\textbf{X}\textbf{W}^{(1)})$; where $\textbf{X}$ is the input training-data matrix and $g(\cdot)$ is the activation function of the neurons.
\item As long as the number of hidden layers is greater than 1, the outputs of each hidden layer shall be obtained as follows: $\textbf{H}^{(K)} = g(\textbf{H}^{(K-1)}\textbf{W}^{(K)})$.
\item The input to the output layer is $\textbf{I} = [\textbf{H}^{(1)};\textbf{H}^{(2)};\ldots;\textbf{H}^{(K-1)};\textbf{H}^{(K)};\textbf{X}]$.
\item Only the output weights ($\boldsymbol\beta$) are obtained during the training process \cite{katuwal2019random}.
\item The final output of the algorithm is defined as follows: $\hat{Y}(\mathbf{x}) = \textbf{I}\boldsymbol\beta $
\end{enumerate}

\subsection{Ensemble dRVFL (edRVFL)}

Sometimes the use of dRVFL network implies excessive computational load in time and memory to obtain the output model ($\boldsymbol\beta$), especially if we work with large amounts of data \cite{zhang2015divide}. A possible solution to this problem can be found in edRVFL networks. Based on the structure of the dRVFL network, this strategy allows us to decompose the $\boldsymbol\beta$ output model into smaller submodels, $\boldsymbol\beta_{ed}$.  Each $\boldsymbol\beta_{ed}$ model is obtained independently, as if they were different learning models, with the final output being the average of all the outputs (in a regression problem), or the one obtained by majority voting (in a classification problem). The input of each hidden layer will be formed by a non-linear transformation of the output of the predecessor layer, as in dRVFL, with the particularity that now, in each layer, the original input features are also taken into account, as in the standard RVFL algorithm.
We can follow the same procedure as in previous section to explain the operation of edRVFL networks. Given a training set $\lbrace(\textbf{x}_i,y_i)\rbrace_{i=1}^n$ as well as the number of layers, $K$ and the number of neurons in each layer $\tilde{N}$:
\begin{enumerate}[leftmargin=*]
\item Randomly assign input weights $\textbf{W}^{(1)},\ldots, \textbf{W}^{(K)}$ and the bias $\textbf{B}^{(1)}, \ldots, \textbf{B}^{(K)}$, between the input and the first hidden layer, and inter hidden layers (for simplicity of notation we will omit the bias of each layer). These parameters kept fixed during the training.
\item The output of the first hidden-layer is defined as: $\textbf{H}^{(1)}=g(\textbf{X}\textbf{W}^{(1)})$. Then, for the rest of the layers, the outputs are: $\textbf{H}^{(K)}=g([\textbf{H}^{K-1}\mathbf{X}]\textbf{W}^{(K)})$. 
\item The outputs weights $(\boldsymbol\beta _{ed})$ are solved independently \cite{katuwal2019random}.
\item Finally, the output can be obtained by averaging the intermediate outputs or by majority voting, depending on whether we are dealing with a regression or a classification problem, respectively.  
\end{enumerate}

\subsection{Echo State Network (ESN)}

Reservoir computing (RC) models are based on Recurrent Neural Networks (RNN), whose architecture has a part called a reservoir that is responsible for projecting a sequence of input data to a set of states. One of the main problems we can find in RNNs is related to vanishing gradient, so \cite{jaeger2002tutorial} designed a type of recurrent reservoir-based architecture, called Echo State Network. The main parts of a standard ESN are: the input layer, the hidden layer or reservoir, which is recurrent and random, and the output layer. The hidden layer is made up of interconnected dynamic neurons, which are activated by a non-linear function, which as in previous sections can correspond to the sigmoid function, or hyperbolic tangent, among others. This hidden layer allows us to increase the dimensions of the inputs. Predictions are obtained by a product between the reservoir states and the output weights \cite{chouikhi2018genesis}, allowing to model long-term relationships between the input and output sequences without suffering from the aforementioned downsides of backpropagation-based neural architectures.

\begin{figure}[!ht]
    \centering
        \includegraphics[width=\columnwidth]{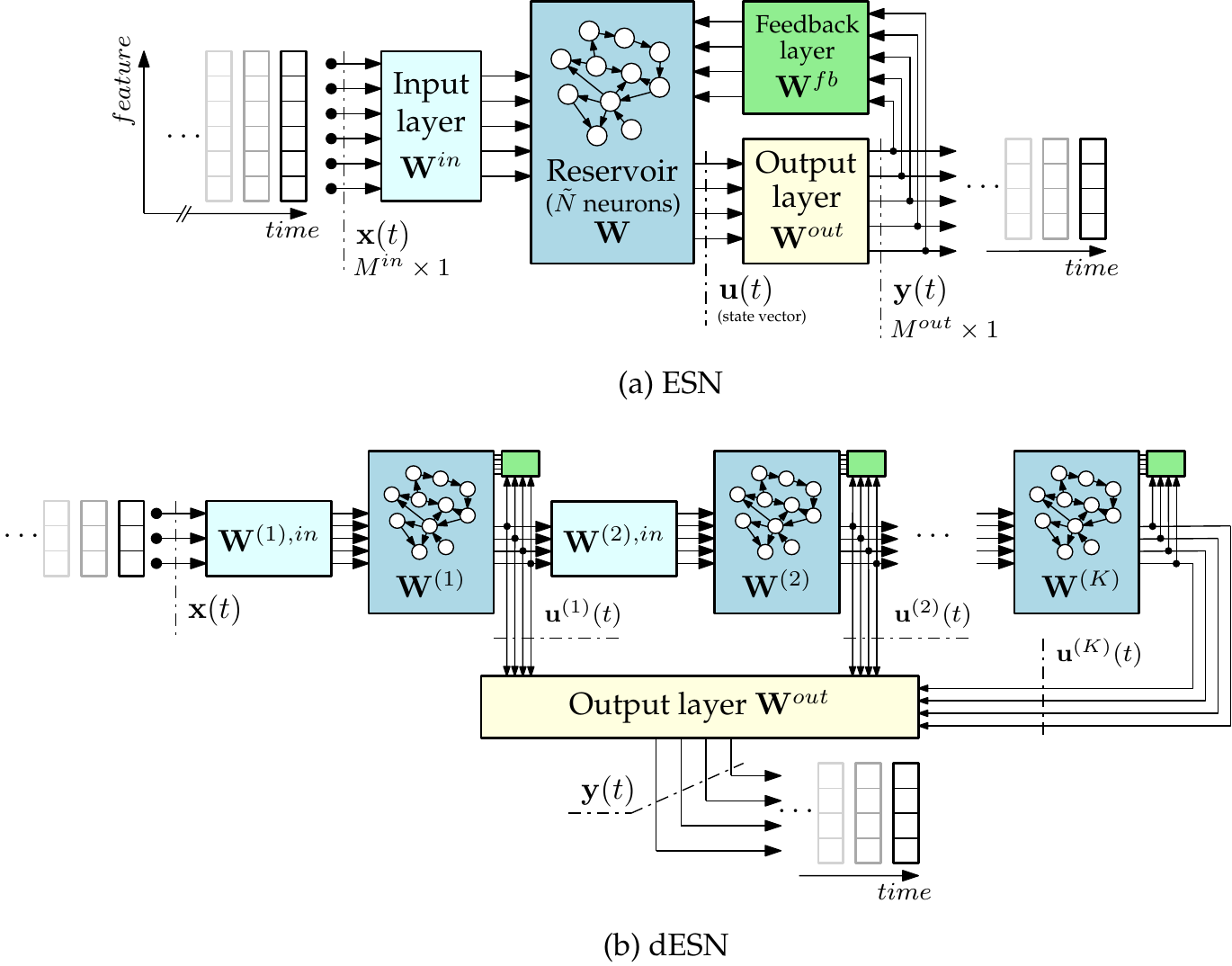}
    \caption{Diagram of (a) an Echo State Network; (b) its deep variant, Deep ESN.}
    \label{ESN} 
\end{figure}

Figure \ref{ESN}.a shows the ESN architecture of a generic ESN. We now describe the whole procedure carried out to obtain the final output of the model in a standard ESN architecture. The training and inference processes can be described as follows:
\begin{enumerate}[leftmargin=*]
\item Vectors $\textbf{x}(t)$, $\textbf{u}(t)$ and $\textbf{y}(t)$ represent the inputs, hidden states and outputs of the network for time $t$ over the sequence, respectively. As such, we assume that the input is a sequence of $M_{in}\times 1$ vectors, where $M_{in}$ denotes the number of features of every input example $\mathbf{x}(t)$. Likewise, the output is composed by a sequence of $M_{out}\times 1$ instances $\mathbf{y}(t)$ that represent the target variable to be predicted.

\item The dynamics of ESN to obtain the state vector at time $t+1$ (recurrent update) are dictated by the following recurrence: 
\begin{equation} \label{eq.rec.ESN}
\textbf{u}(t+1) = \alpha \cdot g\left(\textbf{W}^{in}\textbf{x}(t+1)+\textbf{W}\textbf{u}(t)+\textbf{W}^{fb}\mathbf{y}(t)\right) + (1-\alpha)\mathbf{u}(t),
\end{equation}
where $\mathbf{W}^{in}$ is a $\tilde{N}\times M^{in}$ matrix of input weights; $\textbf{W}$ is a $\tilde{N}\times \tilde{N}$ matrix of hidden weights that map the output of the input weight matrix to the set of hidden states $\mathbf{u}(t)$; and $\mathbf{W}^{fb}$ is a $\tilde{N}\times M^{out}$ feedback matrix. It is important to note that the weights in these matrices are initialized at random (ensuring the fulfillment of certain properties that we will later describe), and kept fixed during the rest of the training process. We partly inherit the notation from preceding sections by denoting as $\tilde{N}$ the size of the reservoir (i.e. its number of recurrently connected neurons). In the above recurrence, $g(\cdot)$ is an activation function that is set beforehand as another hyper-parameter of the model. Finally, $\alpha\in\mathbb{R}[0,1]$ is the so-called \emph{leaky rate} parameter of the reservoir, that permit to adjust the speed at which the state vector of the reservoir \emph{reacts} against changes in the input dynamics.
\item The output of the ESN at time $t+1$ is given by: 
\begin{equation} \label{eq.out.ESN}
\textbf{y}(t+1) = h(\textbf{W}^{out}[\textbf{x}(t+1);\textbf{u}(t+1)]), 
\end{equation}
where $[\cdot;\cdot]$ denotes row-wise concatenation, $\mathbf{W}^{out}$ denotes a $M^{out}\times M^{in}+\tilde{N}$ containing the weights of the output layer, and $h(\cdot)$ is the activation function of the output layer (normally set to be a linear activation, as opposed to $g(\cdot)$, which is usually selected to be a non-linear function \cite{chouikhi2018genesis}). The weights in the output layer $\textbf{W}^{out}$ are the only ones that are trained during the training phase. For this purpose a Moore-Penrose pseudo-inverse matrix or a regularized least squares method is performed to solve an optimization problem similar to those of RVFL (Expression \eqref{eq.opt.RVFL}) and ELM (Expression \eqref{eq.opt.ELM}).
\item Once output weights are adjusted based on training input and output sequences $\{\mathbf{x}(t)\}_{t=1}^T$ and $\{\mathbf{y}(t)\}_{t=1}^T$, we can obtain the predicted output for every new input sequence point $\mathbf{x}(t')$ (with $t'>T$) by applying Expressions \eqref{eq.rec.ESN} and \eqref{eq.out.ESN} \cite{del2020deep}.
\end{enumerate}

To maintain good model performance, it is necessary that the hidden layer or reservoir is well designed, as it is the key to the correct operation of the ESN architecture. Both the number of neurons and the connectivity rate $\lambda$ between them (i.e. the number of non-zero entries in $\mathbf{W}$) are often set by intuitive recommendations: the use of a high number of neurons contributes to the good performance of the model, at the cost of an increased complexity of the network. Choosing the parameters wisely can lead to a proper balance between the accuracy of the ESN model and its training complexity \cite{chouikhi2018genesis}. Furthermore, maintaining stable dynamics is related to the sparsity degree of the hidden neurons $\lambda$. To ensure this stability, $\textbf{W}$ is scaled to meet the so-called \emph{echo state property}, that imposes that the effect on the output of the input of the reservoir should gradually fade over time \cite{yildiz2012re}. This is done by imposing certain algebraic conditions in terms of singular value over the reservoir weight matrix and the leaking rate $\alpha$ that depend on another parameter (spectral radius $\rho_{max}<1$), which establishes the confidence under which the working regime of the reservoir is ensured to comply with the echo state property mentioned previously.

\subsection{Deep ESN (dESN)}

The success of many research projects resorting to multi-layered structures of data-based modeling techniques \cite{tang2015extreme,zhao2015investigation} has encouraged the upsurge of an alternative version of Echo State Networks (dESN), where standard ESNs are stacked to form an ensemble of multiple reservoirs, capable of capturing patterns at different temporal scales. Indeed, the presence of multiple reservoirs is the main difference between a standard ESNs and dESNs, as they increase the level of non-linearity and granularity by which the model is capable of representing feature mappings over sequences with high non-linear complexity \cite{chouikhi2018genesis}. Put differently, dESNs can be understood as a concatenation of interconnected reservoirs, where successively the outputs of one reservoir will form the inputs of the next reservoir. Both the input and output layers in dESNs remain the same as in the standard model. 

Considering the notation and reservoir dynamics explained in the previous section, we now detail how dESN can be trained, supported by the diagram of a generic dESN shown in Figure \ref{ESN}.b: 
\begin{enumerate}[leftmargin=*]
\item We extend the notation of input and hidden reservoir matrices with superscript $(k)$, so that $\textbf{u}^{(k)}(t)$ denotes the hidden state vector of the $k^{th}$ reservoir for time $t$ time samples. Similarly, $\textbf{W}^{(k),in}$, $\textbf{W}^{(k)}$ and $\textbf{W}^{(k),fb}$ denote the input weight matrix, the hidden recurrent weight matrix and the feedback matrix of reservoir $k$. We assume that $K$ reservoirs are stacked one after another, so that $k\in\{1,\ldots,K\}$.
\item The dynamics of dESN to obtain each hidden state vector at time $t$ (recurrent update) are given, for $k=1$, by:
\end{enumerate}
\begin{align}
&\textbf{u}^{(1)}(t+1) = \alpha^{(k)} h(\textbf{W}^{(1),in}\textbf{x}(t+1)+\textbf{W}^{(1)}\textbf{u}^{(1)}(t))+ (1-\alpha^{(1)})\mathbf{u}^{(1)}(t),\label{eq.deepstate1}
\end{align}
\begin{enumerate}[leftmargin=*]
\item[] whereas, for $1<k\leq K$:
\end{enumerate}
\begin{align}
&\textbf{u}^{(k)}(t\hspace{-0.75mm}+\hspace{-0.75mm}1)\hspace{-0.75mm}=\hspace{-0.75mm}\alpha^{(k)} h(\textbf{W}^{(k),in}\textbf{u}^{(k-1)}(t\hspace{-0.75mm}+\hspace{-0.75mm}1)\hspace{-0.75mm}+\hspace{-0.75mm}\textbf{W}^{(k)}\textbf{u}^{(k)}(t))+ (1\hspace{-0.75mm}-\hspace{-0.75mm}\alpha^{(k)})\mathbf{u}^{(k)}(t), \label{eq.deepstate2}
\end{align}
\begin{enumerate}[leftmargin=*]
\item []where $h(\cdot)$ is the activation function of the reservoirs.

\item[3.] In the same way, the output at time $t+1$ is obtained by \cite{chouikhi2018genesis}: 
\begin{equation}
\textbf{y}(t) = g(\textbf{W}^{out}[\mathbf{u}^{(1)}(t);\mathbf{u}^{(2)}(t);\ldots;\mathbf{u}^{(K)}(t)]),
\end{equation}
where $g(\cdot)$ is the activation function of the outputlayer, and the amount of information from the reservoirs mapped to the output sequence is the concatenation of all hidden state vectors of the stacked hierarchy of reservoirs. Here, the output weight matrix $\mathbf{W}^{out}$ is computed in a similar fashion to the case of a single-layered ESN model. We also note that other alternative formulations of the output layer for dESN can be found in the literature where e.g. only the hidden state vector $\mathbf{u}^{(K)}(t)$ of the last reservoir is considered.
\end{enumerate}

As data propagates through the reservoirs, instances over the input sequence undergo non-linear transformations that allow extracting features that were not previously visible in their original form, which are further captured over different time instants by virtue of the recurrent nature of the reservoirs. The fact that such patterns are captured at different granularity scales is the main advantage of dESNs over standard ESNs.

\section{Literature review}\label{Literature_Review}

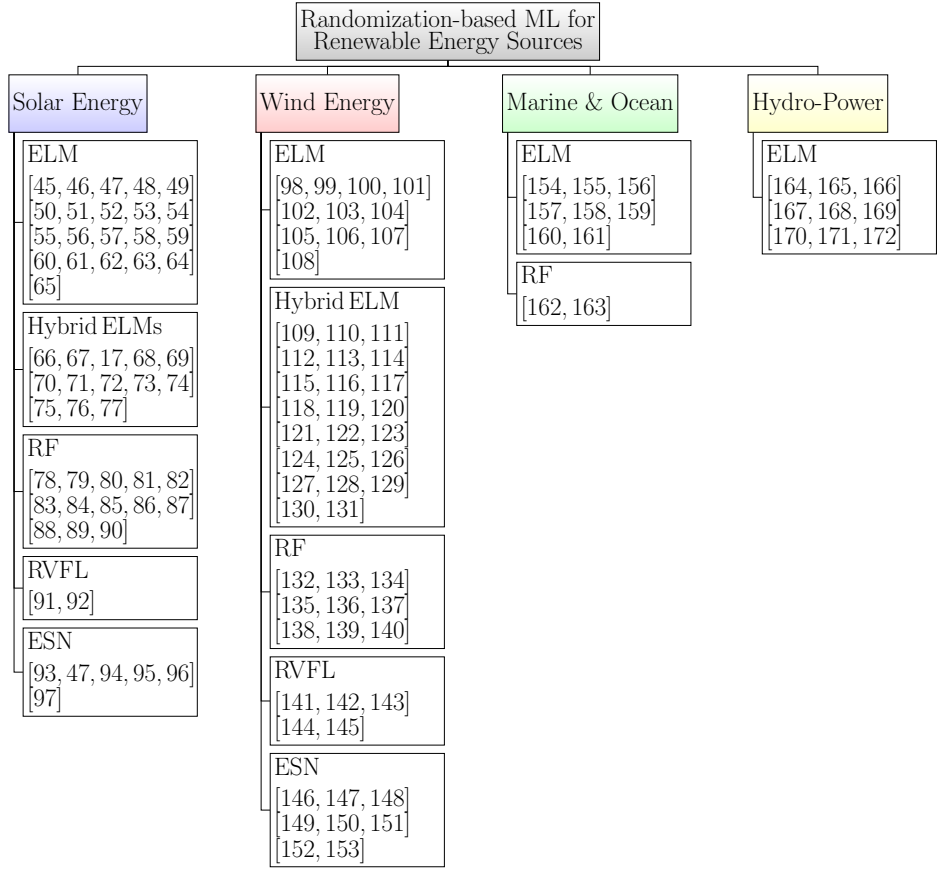
\begin{figure}[h!]
	\resizebox{0.9\columnwidth}{!}{\begin{forest}
			upper style/.style = {draw, rectangle, top color=white, bottom color=black!20},
			lower style/.style = {draw, thin, text width=14em,align=left},
			s sep'+=50pt,
			forked edges,
			where level<=1{%
				upper style
			}{
				lower style,
			},
			where level<=0{%
				parent anchor=children,
				child anchor=parent,
				if={isodd(n_children())}{%
					calign=child edge,
					calign primary child/.process={
						O+nw+n{n children}{(#1+1)/2}
					},
				}{%
					calign=edge midpoint,
				},
			}{
				folder,
				grow'=0,
			},
			[{{\Huge Randomization-based ML for}\\{\Huge Renewable Energy Sources}}, for tree={fill=white,minimum size=2cm}
			[{{\huge Solar Energy}}, bottom color=blue!20
			[{{\huge ELM}} \vspace{3mm}\\
			{\huge \cite{csahin2014application,abadi2014extreme,jayawardene2014comparison,Shamshirband15_2,Shamshirband15}}\\
			{\huge \cite{hou2018global,Deo19,Cornejo19,Karaman21,teo2015forecasting}}\\
			{\huge \cite{li2015day,burianek2016solar,Tang16,Hossain17,Chen17}}\\
			{\huge \cite{al2018extreme,du2018designing,Behera18,Behera20,pani2019short}}\\
			{\huge \cite{pani2019forecasting}}\\
			]
			[{{\huge Hybrid ELMs}} \vspace{3mm}\\
			{\huge \cite{Salcedo14,Aybar16,salcedo2018feature,Salcedo17,Bouzgou17}}\\
			{\huge \cite{hosseini2017comparison,Ghimire18,Salcedo18,majumder2018variational,zhang2019hybrid}}\\
			{\huge \cite{Zhou20,Liu20,Feng20}}\\
			]
			[{{\huge RF}} \vspace{3mm}\\
			{\huge \cite{Sun16,Ibrahim17,Torres19,Ahmad18,Srivastava19}}\\
			{\huge \cite{Babar20,Benali19,Prasad19,Prasad20,Niu20}}\\
			{\huge \cite{Liu19,Zeng19,abuella2017random}}\\
			]
			[{{\huge RVFL}} \vspace{3mm}\\
			{\huge \cite{majumder2019short,aggarwal2018short}}\\
			]
			[{{\huge ESN}} \vspace{3mm}\\
			{\huge \cite{ruffing2009short,jayawardene2014comparison,Yao19,Li20EchoMulti,wu2020multi}}\\
			{\huge \cite{Li20EchoSpatio}}\\
			]
			] 
			[{{\huge Wind Energy}}, bottom color=red!20
			[{{\huge ELM}} \vspace{3mm}\\
			{\huge \cite{wu2013extreme,Mohamidi15,lazarevska2016wind,Nikolic16}}\\
			{\huge \cite{Yin17,Deng19,Xiao21}}\\
			{\huge \cite{luo2018short,tian2018short,li2019wind}}\\
			{\huge \cite{tian2020prediction}}\\
			]
			[{{\huge Hybrid ELM}} \vspace{3mm}\\
			{\huge \cite{Salcedo14b,Liu15,Wang15}}\\
			{\huge \cite{Mladenovic16,li2016short,li2016wind}}\\
			{\huge \cite{huang2016hybrid,qolipour2019prediction,Zheng17}}\\
			{\huge \cite{Peng17,TPENG17,mi2017wind}}\\
			{\huge \cite{Yang19,Zhang19,Manohar18}}\\
			{\huge \cite{Wang18,Mahmoud18,Liu18}}\\
			{\huge \cite{Hu18,wang2018research,zhou2018short}}\\
			{\huge \cite{sun2018adaptive,tian2019artificial}}\\
			]
			[{{\huge RF}} \vspace{3mm}\\
			{\huge \cite{Lahouar17,shi2018improved,alonso2015random}}\\
			{\huge \cite{sun2018multistep,lin2015seasonal,niu2018ultra}}\\
			{\huge \cite{kaya2018hybrid,natarajan2020wind,vassallo2020analysis}}\\
			]
			[{{\huge RVFL}} \vspace{3mm}\\
			{\huge \cite{ren2015detecting,nhabangue2018wind,Ahmed18}}\\
			{\huge \cite{mishra2019short,jalli2020prediction}}\\
			]
			[{{\huge ESN}} \vspace{3mm}\\
			{\huge \cite{Liu15Echo,dorado2017robust,lopez2018wind}}\\
			{\huge \cite{chitsazan2019wind,Wang19Echo,wang2019novel}}\\
			{\huge \cite{Chen19Echo,Hu20Echo}}\\
			]
			] 
			[{{\huge Marine \& Ocean}}, bottom color=green!20
			[{{\huge ELM}} \vspace{3mm}\\
			{\huge \cite{Alexandre15,Cornejo16,cornejo2016grouping}}\\
			{\huge \cite{Cornejo18Bayes,Kumar18,Ali19}}\\
			{\huge \cite{Kaloop20,shamshirband2020prediction}}\\
			]
			[{{\huge RF}} \vspace{3mm}\\
			{\huge \cite{Serras19,callens2020using}}\\
			]
			] 
			[{{\huge Hydro-Power}}, bottom color=yellow!20
			[{{\huge ELM}} \vspace{3mm}\\
			{\huge \cite{li2014monthly,atiquzzaman2018robustness,Yaseen19}}\\
			{\huge \cite{Feng19,Wang20,Ribeiro20}}\\
			{\huge \cite{niu2019comparison,lian2020trend,Chen20}}\\
			]
			] 
			] 
	\end{forest}}   
	\centering
	\caption{Summary of recent works dealing with randomization-based ML approaches in renewable energy prediction problems.}
	\label{fig:taxonomy}
\end{figure}

This section reviews the most significant works in the literature on the application of randomization-based ML algorithms to renewable energy prediction problems. We have structured the section by energy resources (solar, wind, marine/ocean and hydro-power), and then by type of randomization-based algorithm. Note that ensemble methods have been included in the corresponding randomization-based algorithms which form the ensemble. The discussion of the works in each case has a chronological structure. Figure \ref{fig:taxonomy} shows a summary of the references discussed in this section, which may serve as guide of the structure of the following subsections. Before further proceeding with the analysis of the existing literature, we first describe the methodology followed for properly collecting all contributions published so far on randomization-based algorithms in renewable energy prediction problems, in Section \ref{methodology_bbl}. We will close this literature review with a critical analysis of the trends unveiled during our review of the literature, which can be found in Section \ref{critical_literature_analysis}.

\subsection{Bibliographic methodology for the literature review}\label{methodology_bbl}

A large number of search queries was performed in well-known scientific publication databases, including Scopus, Web of Science, Google Scholar. A variety of query strings was utilized for a systematic discovery of published works that are related to the topic of this survey, including the name of the main randomization-based methods (e.g. \texttt{Random Forest}, \texttt{Extreme Learning Machines} or \texttt{Echo State Network}) plus \texttt{solar prediction}, \texttt{wind prediction}, \texttt{marine prediction} or \texttt{hydro-power prediction}, among other terms linked to renewable and sustainable energy. Once all results were retrieved from the aforementioned databases, we removed duplicates and performed an exhaustive analysis on a paper by paper basis, towards ascertaining their alignment with the topic under study. This systematic review process gave rise to the literature review and analysis that we present in the subsequent sections.

\subsection{Solar energy}

\subsubsection{ELMs in solar radiation prediction problems}
In the last years, the application of ELMs to solar radiation prediction and estimation problems has been massive. Some of the first works dealing with ELMs in solar radiation prediction were proposed from 2014 in the most important journals and conferences on renewable energy.  In \cite{csahin2014application} the ELM was applied to a problem of solar radiation estimation in Turkey. In this case, satellite data from 20 different locations over Turkey were used. The predictive variables included land surface temperature, altitude, latitude, longitude, month, and city location. The ELM performance was compare to that of a traditional neural network with back-propagation training, obtaining significant improvements. In \cite{abadi2014extreme} also an ELM and a multi-layer feed-forward network with back propagation are implemented to estimate hourly solar radiation on horizontal surface in Surabaya, Java, Indonesia. Predicted variables included meteorological data such as temperature, humidity, wind speed, and direction of speed as inputs for the prediction model. This study reported a good performance of the ELM in comparison with the multi-layer network with back-propagation training in this solar radiation estimation problem. In \cite{jayawardene2014comparison} a comparison of the solar energy estimation performance obtained by an ELM with an ESN is carried out. The prediction is based on input variables such as current solar irradiance, temperature and PV plant power output. In \cite{Shamshirband15_2} the capacity of the ELM for obtaining good predictions of global solar radiations is evaluated. A comparison with SVR and Multi-layer perceptrons in data from Iran is carried out. In a similar work, \cite{Shamshirband15} presents a kernel-based ELM (KELM) applied to a problem of daily horizontal global solar radiation, considering the maximum and minimum air temperatures as input variables. Experiments in data from a city of Southern Iran (Bandar Abass), with great solar potential, showed the goodness the KELM in the prediction of daily horizontal global solar radiation, improving the results obtained with a Support Vector Regression approach. In \cite{hou2018global} a novel regularized online sequential ELM which is proposed to a problem of daily solar radiation prediction. The model integrates a variable forgetting factor (FOS-ELM) to predict global solar radiation at Bur Dedougou (Burkina Faso). The Bayesian information criterion is applied as a feature selection system to obtain the best inputs for the prediction. In \cite{Deo19} this paper presents a study that designed a regionally adaptable and predictively efficient ELM model to forecast long-term incident solar radiation over Australia. The relevant satellite-based input data extracted from the Moderate Resolution Imaging Spectro-radiometer has been considered. In \cite{Cornejo19} different ML regressors, have been tested in a problem of solar radiation estimation from satellite data. Specifically, ELMs, Multi-Layer perceptrons, SVR and GPR have been tested and compared. Analysis of the results obtained in a real problem of solar radiation estimation at Toledo, Spain, has been carried out. In \cite{Karaman21} the authors compare standard ELMs vs ANNs for solar radiation estimation. Networks are trained with data acquired from Karaman province (Turkey) in the period 2010-2018. The results obtained showed better estimations for ELM than MLPs tested with different activation functions.

\subsubsection{ELMs in PV power systems production prediction}

In \cite{teo2015forecasting} the ELM was applied to the solar energy production of PV systems in Germany, considering different input variables such as ambience temperature, temperature of the PV panels, accumulated irradiance or daily energy. Similarly \cite{li2015day} also tackles a problem of PV systems production prediction with ELMs. Specifically, 1-day-ahead hourly forecasting of PV power output in Shanghai, China, considering different models depending on the weather conditions (sunny days, cloudy days, and rainy days). Another work dealing with PV systems production prediction with ELMs is \cite{burianek2016solar}, in this case meteorological input variables, mainly cloudiness, are considered. In \cite{Tang16} a hybrid approach involving entropy method and ELMs were applied to a prediction problem of short-term PV power generation. The entropy method is used to carry out a first processing of the data, using the ELM to obtain a final forecast of electricity generation in the PV system. A comparison with generalized regression neural network and radial basis function neural networks is carried out. In \cite{Hossain17} a day-ahead and 1-hour-ahead mean PV output power prediction problem has been tackled, by applying an ELM approach. Meteorological input variables recorded in three grid-connected PV systems at University of Malaya, Malaysia, are used to train the ELM. A comparison with a SVR and MLPs is used to evaluate the performance of the proposed approach. In \cite{Chen17} a KELM approach optimized with the Nelder-Mead Simplex optimization method is employed in a problem of PV arrays fault diagnosis. Experiments are carried out considering data from a laboratory PV array installed on the Physics and Information Engineering in Fuzhou University, China. In \cite{al2018extreme} an ELM is used to provide accurate 24 h-ahead solar PV power production predictions. The proposed ELM model is applied to a real case study of 264 kWp solar PV system installed at the Applied Science Private University, Amman, Jordan. In \cite{du2018designing} a Maximum Power Point Tracker for PV systems is designed, supported by a supervised weather-type classification system using a fuzzy-weighted ELM. A comparison with a SVM is carried out to show the performance of the proposed system. In \cite{Behera18} a PV power forecasting system leading by an optimized ELM technique is proposed. In this case, the ELM weights are modified with a Particle Swarm Optimization (PSO) algorithm to optimize its performance. Experiments in the Indian state of Orissa are carried out, in which the performance of the optimized ELM is compared with a MLP network. The same PV system's data from Orissa are used in \cite{Behera20}, where a short-term PV power forecasting is proposed by a 3-steps approach, formed by combining empirical mode decomposition technique, sine cosine algorithm , and ELMs. In \cite{pani2019short} a Pruned-ELM (P-ELM) approach is proposed for a problem of short-term PV power prediction. P-ELM shows statistical ways to compute the significance of inner nodes in the network structure. Starting from an initial large number of inner nodes, inappropriate nodes are then pruned by taking into account the appropriateness to the forecasting problem. In this case, the performance of the P-ELM is further improved by modifying the weights of input layer with a PSO algorithm. In \cite{pani2019forecasting} a Wavelet KELM hybridized with a Gravitational Search Algorithm is proposed for a problem of solar irradiance forecasting. Experiments have been carried out in data from a real power plant of $1$MW in India.

\subsubsection{Hybrid ELMs techniques in solar radiation prediction problems}
In \cite{Salcedo14}, a hybrid meta-heuristic -- ELM approach is proposed, in which the Coral Reefs Optimization (CRO) algorithm was used to assign the ELM weights to improve the solar radiation prediction performance of the algorithm. Experiments in data from the radiometric observatory at Toledo, Spain, confirmed the good performance of this proposal.
In \cite{Aybar16} a hybrid Grouping Genetic Algorithm (GGA) and ELM (GGA-ELM) is proposed for solar radiation estimation from numerical weather models. The GGA is used to select groups of input variables that maximizes the prediction performance, given by an ELM. In this case the hybrid approach forms a wrapper feature selection approach \cite{salcedo2018feature}, where the ELM is used due to its good properties of performance and low computational cost. Results in data from the Radiometric Observatory of Toledo (Spain), shows the good performance of this approach. \cite{Salcedo17} discusses a similar application of feature selection in a problem of solar radiation estimation. In this case a hybrid approach mixing a CRO algorithm for selecting the features and an ELM for obtaining the solar radiation prediction is proposed. In \cite{Bouzgou17} a hybrid approach mixing mutual information and ELM is proposed in a problem of time series solar irradiance prediction. The prediction model includes different scenarios, such as long windows, short windows, standard Principal Components Analysis (PCA) and clear-sky model inclusion. Results in time series from different parts of the World have shown the good performance of this approach.  In \cite{hosseini2017comparison} a comparison among different ELMs for predicting daily horizontal diffuse solar radiation in a region of southern Iran is carried out. The work discusses hybrid methods based on ELMs, such as complex ELM (C-ELM), self-adaptive evolutionary ELM (SaE-ELM), and online sequential ELM (OS-ELM). The precision of the C-ELM, SaE-ELM, OS-ELM, and ELM models is evaluated for different Iran regions, including data sets of southern Iranian cities (Yazd, Shiraz, Bandar Abbas, Bushehr, and Zahedan). In \cite{Ghimire18} a new approach to predict the monthly mean daily solar radiation based on a hybrid ELM model is proposed. Specifically a self-adaptive differential evolutionary ELM is proposed, using a swarm-based ant colony optimization for feature selection. This hybrid ELM model has been integrated with the MODIS-based satellite data and the European Center for Medium Range Weather Forecasting (ECMWF) Reanalysis data. Experiments in Australian solar rich cities (Brisbane and Townsville) are carried out. In a similar approach, \cite{Salcedo18} has proposed a hybrid CRO with ELM algorithm for solar radiation estimation in Sunshine Coast and Brisbane, Australia, obtaining excellent results in terms of solar prediction error. In \cite{majumder2018variational}, a new hybrid method combining Variational Mode Decomposition (VMD) and a KELM for solar irradiation forecasting is proposed. In this case the VMD decompose a solar radiation signal into different modes, which are processed by the KELM to obtain the prediction. Experiments with data from the Indian city of Tangi, Odisha, have been carried out. In \cite{zhang2019hybrid} a hybrid model that predicts solar radiation through ELM optimized by the bat algorithm based on wavelet transform and PCA is proposed. Results in data from four different cities in North America, Asia and Australia illustrate the performance of this approach. In \cite{Zhou20}, the authors propose a new hybrid ML model composed by an ELM, a genetic algorithm (GA) and customized similar day analysis, for PV power prediction. They show than single ML does not have stable prediction performances compared to the combination of models. The SDA-GA-ELM is evaluated on the real-world dataset from Desert Knowledge Australia Solar Center (DKASC). In \cite{Liu20}, the authors propose a short-term photovoltaic power regressor based on an ELM model, the ICSO-ELM. The model input is computed from the correlation coefficient model and the convergence is accelerated using a class of optimized based on swarm optimization. This special optimizer is also used to compute the weights and bias of the ELM. The authors test the proposed method in DKASC database. In \cite{Feng20} a new hybrid PSO and ELM approach, is proposed to accurately predict daily solar radiation. A complete comparison with alternative ML approaches is carried out using long-term solar radiation data in the period 1961-2016 from seven stations located on the Loess Plateau of China.

\subsubsection{RF approaches in solar radiation prediction} 

In \cite{Sun16} the potential of the air pollution index for estimating solar radiation is evaluated. Meteorological data, solar radiation, and air pollution index data from three sites, with different air pollution index conditions are used to develop RF models. Then, RF models with and without considering air pollution index data are compared. The results show that the performance of random forest models with air pollution index data is better than that of the empirical methodologies. In \cite{Ibrahim17} a novel hybrid model for predicting hourly global solar radiation using RF and firefly algorithm is proposed. In this case, the firefly algorithm is used to optimize the RF technique by finding the best number of trees and leaves per tree in the algorithm. This hybrid approach has been tested in the prediction of hourly global solar radiation at Klang Valley, Malaysia.  In \cite{abuella2017random} an ensemble prediction model for solar power output of a solar PV system is proposed. The ensemble model is formed by a set of support vector machines, which generate the forecasts and a RF which acts as an ensemble learning method to combine the forecasts. The authors in \cite{Ahmad18} elaborate a comparison among ML models, RF and extra trees, decision trees and SVR, for solar thermal energy systems. All methods are compared based on their generalisation ability (stability), accuracy and computational cost. They found that the best accuracy were obtain by RF and ET method in case of study located in Chamb\'ery (France). In \cite{Torres19}, the authors elaborate a comparison among different ensemble models, specifically, RF regression, Gradient Boosted Regression (GBR) and Extreme Gradient Boosting (XGB) for global and local wind energy prediction and solar radiation. Experiments with real data prove that these ensemble methods improve on SVR for individual wind farm energy prediction, and show that GBR and XGB are competitive predicting wind energy in a much larger geographical scale. In solar energy prediction, both gradient-based ensemble methods improve the performance of the SVR. The authors consider two scenarios to evaluate the performance of the randomized ensemble models in wind energy prediction: predicting the output of a single farm, the Sotavento wind farm, situated in Galicia (Spain) and predicting the total wind energy production of peninsular Spain. In the case of the prediction of daily aggregated incoming solar energy, a total of 98 {\em Mesonet} weather stations covering the state of Oklahoma were considered. The authors in \cite{Srivastava19} propose several ML methods for solar radiation forecasting for a few days in advance, which result critical for solar power plants. Specifically, in this study, the authors evaluate MARS, CART, M5 and RF models for 1-day-ahead to 6-day-ahead hourly solar radiation forecasting. Data has been collected at the cite of Gorakhpur, India. The study concludes that RF models obtain the best predictions among all the methods compared. The work in \cite{Benali19} proposes to study ANNs and RF methods for solar radiation forecasting, devoting special attention on the normal beam, horizontal diffuse and global components. The dataset for training the methods was acquired on the site of Odeillo (France), a region characterized for having a high meteorological variability. The objective is to predict hourly solar irradiations for time horizons from h+1 to h+6. The RF method reached the best accurate prediction for all measured parameters in all time horizons tested.  In \cite{Prasad19} a multi-stage multivariate empirical mode decomposition hybridized with ant colony optimization and RF has been applied to a problem of monthly solar radiation prediction for three locations of the Queensland state, Australia. A comparison with RF, M5tree and mini-max probability machine regression has shown the goodness of the proposed hybrid approach. In \cite{Liu19} PCA and K-means clustering algorithm, combined with RF optimized by Differential Evolution Grey Wolf Optimizer have been applied to a problem of PV power generation prediction. A comparison of the proposed method against SVM, Neural Networks, Decision tree and Gaussian regression model has been carried out. In \cite{Zeng19} a RF model has been applied to construct high-density network of daily global solar radiation and its spatio-temporal variations in China. Observational data from $2.379$ meteorological stations and solar radiation observations, have been used to feed the RF model. The work in \cite{Babar20} focuses on applying RF regression for improving the solar irradiance maps at high latitudes. Data from meteorological reanalyses and retrievals may present severe bias, especially in high latitude regions. Specifically, data from the ECMWF Reanalysis 5 (ERA5) and the Cloud, Albedo, Radiation dataset Edition 2 were used as inputs for a RF regression model. The proposed RF with reanalysis inputs was tested on five Swedish locations, where it was found to improve surface solar irradiance estimates from previous approaches.
In \cite{Prasad20} the empirical mode decomposition method and the singular value decomposition algorithm have been coupled with an RF for obtaining an accurate weekly solar radiation prediction model. Results in four different location at Queensland state, Australia, are reported. In \cite{Niu20} a hybrid forecasting model for PV power generation composed by a large number of algorithms was proposed. Specifically, the proposal included the combination of RF, improved grey ideal value approximation, complementary ensemble empirical mode decomposition, a PSO algorithm based on dynamic inertia factor, and a back-propagation neural network. Experiments have been carried out at the Desert Garden PV station in Canada. 

\subsubsection{ESNs in solar radiation prediction}

Interestingly, the first work dealing with an ESN in a solar energy prediction problem is earlier than other approaches such as ELM or RF. Specifically, it is \cite{ruffing2009short}, where an ESN is proposed to give multi-step predictions of solar irradiance, from 30 to 270 minutes time-horizons into the future. The ESN is trained and tested using solar data from the National Renewable Energy Laboratory Solar Radiation Research Laboratory in Golden, Colorado. The proposed approach overcomes the conventional Recurrent Neuronal Networks which find difficulties to forecast in multi-step time horizons. After this approach, ESNs were not applied to solar radiation prediction problems until \cite{jayawardene2014comparison}, which presents a comparison between ESN and ELMs for photovoltaic power prediction. The trained models provide multiple time predictions of a large photovoltaic plant at eight different time steps varying from a few seconds to a minute plus. Forecasting is developed using the current solar irradiance, temperature and photovoltaic plant power output as input variables. Both ESNs and ELMs have been designed to be executed in real-time and tested on a real dataset. In the experiments carried out, it is shown that ESNs strongly surpasses ELM in terms of mean-squared error and correlation. No further ESN approaches for solar radiation prediction problems have been proposed until very recently, in \cite{Yao19}, where a novel forecasting model based on multiple reservoirs echo state network (MR-ESN) is proposed to forecast the power output of photovoltaic generation systems. The work, first extracts features from the input through the unsupervised learning algorithm of restricted Boltzmann machine. Then, using the forecasting performance evaluation criteria, a principal component analysis is used to extract the main features. Finally,  the Davidon-Fletcher-Powell quasi-Newton algorithm is used for iteratively finding the optimal weights. The method is tested in a PV plant power output time series at Twentynine Palms (USA). In \cite{Li20EchoMulti}, a MR-ESN is proposed for accurate solar irradiance prediction in renewable energy systems. MR-ESN has the high quality efficiency of ESNs with the advantages of deep learning. The approach receives the multiple reservoirs data in series which are transformed into a richer state representation. Various prediction time-horizons including 1-hour-ahead and several-hour-ahead prediction are tested. The results obtained shown that the MR-ESN outperforms the traditional ESN and Elman neural networks performance. In \cite{wu2020multi}, an ESN is proposed to multi-timescale forecast of solar irradiance, based on multi-task learning. The ESN simultaneously predicts solar energy generation on different timescales for hierarchical decisions. The paper uses the multitasking learning paradigm to study the multi-timescale forecasting which its best contribution. The ESN has been trained and tested with data from 6 different stations from the California Irrigation Management Information System. In \cite{Li20EchoSpatio}, a chain-structure ESN (CESN) for enhancing the scalability, robustness and computational efficiency in spatio-temporal solar irradiance prediction is proposed. The number of ESN modules in the CESN is determined by a spatial autocorrelation analysis. This autocorrelation analysis is also developed on the temporal information of each spatial variable to provide appropriate inputs for each ESN module. Finally, the ensemble spatio-temporal solar irradiance prediction model is built based on CESN. The paper shows that CESN achieves more accurate predictions than Elman neural networks, or classical ESNs, in datasets from California Irrigation Management Information System.

\subsubsection{RVFL networks in solar energy prediction}

There are not many works dealing with RVFL in solar energy prediction problems, in spite that this is a very promising technique for these applications, as we will show later on in the experimental section of this work. We have only found two references applying RVFL approaches to solar energy prediction problems. The work in \cite{aggarwal2018short} proposes a RVFL network for short-term solar power forecasting, and compares it with other two artificial neural systems: the single shrouded layer feed-forward neural network (SLFN) and the random weight single shrouded layer feed-forward neural network (RWSLFN). Data for training and evaluating were adquired from the solar power data of Sydney, Australia. The results obtained in this training data prove that RVFL outperforms the RWSLFNs or SLFNs.
In turn, in \cite{majumder2019short} a new hybrid model combining kernel functions along with the RVFL network is proposed for short-term solar power prediction. This new method shares with the original RVFL its fast training speed, reduced architecture and good generalization skills. The proposal consists of using specific local and global kernel functions in both direct links and hidden nodes to improve the prediction accuracy. For retrieving the optimal kernel, the work applies a metaheuristic evaporation-based water cycle algorithm. The validation of the model has been carried out using data from two solar power plants located in the states of New York and California (USA). 

\subsection{Wind energy}

\subsubsection{ELMs in wind speed prediction}
In \cite{wu2013extreme} an ELM for wind speed estimation and sensorless control for wind turbine power generation systems, is proposed. The ELM model is designed to run in real time. Wind speed estimations are used to determine the optimal rotational speed command for maximum power point tracking. The method is tested with synthetic and real data. In \cite{Mohamidi15} a wind speed prediction model based on ELM is presented to estimate the wind power density. Data training have been obtained from two accurate Weibull methods of standard deviation and power density. The validity of the ELM model is verified by comparing its predictions with SVR, MLP and a GP approach. The wind powers predicted by all approaches are compared with those calculated using measured data, obtaining good results. In \cite{lazarevska2016wind} an approach to forecasting the wind speed based on ELMs is discussed. In this case the wind speed is modelled by means of available meteorological data such as solar radiation, air temperature, humidity, pressure, etc. In \cite{Nikolic16} a model based on ELMs for sensor-less estimation of wind speed based on wind turbine parameters is proposed. The inputs for estimating the wind speed are then the wind turbine power coefficient, blade pitch angle, and rotational speed. In order to validate the results, the authors compared the prediction capabilities of the ELM model with that of the GP, MLP and SVR. In \cite{Yin17} an effective secondary decomposition model is developed for wind power forecasting using ELM trained using a crisscross optimization. During the pre-processing step, the input time series is decomposed into several Intrinsic Mode Functions (IMFs), applying wavelet packet decomposition (WPD). During the training phase, the transformed sub-series are predicted with the ELM. In \cite{luo2018short} the authors propose a ML-based approach for facing some the difficulties to handle the large-scale dataset generated in a wind forecasting application. The approach is based on a combination of an ELM and a deep-learning model. In \cite{tian2018short} a wind power prediction algorithm based on EMD and ELM is proposed. First, wind power time series is decomposed into several components with different frequency by EMD, for reducing the non-stationary of time series. Then, and improved ELM is applied to obtain the prediction of the wind power. In \cite{Deng19} a sensorless wind speed estimation algorithm based on the unknown input disturbance observer and the ELM for the variable-speed wind turbine. Specifically, the idea is to estimate the aerodynamic characteristics of the wind turbine with the ELM, and from them, to estimate wind speed also based on the ELM model, using the previously estimated aerodynamic torque by the unknown input disturbance observer, together with the measured rotor speed and pitch angle. In \cite{li2019wind} a wind power prediction method based on ELM with kernel mean p-power error loss is proposed. The prediction capabilities of this approach are compared with the traditional BP neural network. In \cite{tian2020prediction} a prediction approach for short-term wind speed using ensemble EMD-permutation entropy and regularized ELM is proposed. First, wind speed time series is decomposed into several components with different frequency by the EMD decomposition. The permutation entropy value for each component is used to analyze its complexity. The components can be recombined to obtain a set of new subsequences, which are used as inputs of different prediction models based on regularized ELMs to obtain the wind speed prediction. Finally, in \cite{Xiao21} a self-adaptive KELM is designed for short-term wind speed forecasting. The self-adaptive KELM could simultaneously obsolete old data and learn from new data by reserving overlapped information between the updated and old training datasets. Experiments in wind speed data from three different stations at Penglai region (China) are employed as a numerical experiment.

\subsubsection{Hybrid ELM models for wind speed prediction}
In \cite{Salcedo14b} a hybrid ELM-CRO approach is proposed for accurate wind speed prediction. The proposal is a wrapper approach for feature selection, in which the CRO is used to select the best set of features which minimize the error of an ELM in a validation set. The final set of features is then used to train another ELM for test the performance of the proposed approach. Experiments in data from a real wind farm in Portland, Oregon, USA, have shown the good performance of this approach. In \cite{Liu15} a hybrid method is proposed for accurate wind speed  forecasting. Specifically, four different hybrid models are presented by combining signal decomposing algorithms (WD, WPD, EMD and fast ensemble EMD) and ELMs. The results obtained with these hybrid approaches shown that the ELM is a robust method for wind speed prediction, and all the proposed hybrid algorithms have better performance than the single ELM, showing that using decomposition methods improves the information processing carried out by the ELM approach. In \cite{Wang15} an hybrid ensemble model for probabilistic short-term wind speed forecasting is proposed. The ensemble starts with a Empirical Wavelet Transform, which is employed to extract meaningful information from wind speed series. A Gaussian Process Regression (GPR) model is then used to combine the predictions of different forecasting engines (Autoregressive Integrated Moving Average (ARIMA), ELM, SVR and Least Square SVR), and obtaining this way a final wind speed prediction. The effectiveness of the proposed ensemble approach is demonstrated with wind speed data from two wind farms in China. In \cite{Mladenovic16} an ELM coupled with wavelet transform (ELM-WAVELET) is used for the prediction of wind turbine wake effect in wind farms. Estimation and prediction results of ELM-WAVELET model are compared with those by ELM on its own, GP, SVR and MLPs in this problem of wake estimation. In \cite{li2016short} a combined approach based on ELMs and an error correction model based on persistence is proposed to predict wind power in the short-term time horizons. First, an ELM is used to forecast the short-term wind power, and then this prediction is further processed using a short-term forecasting error by applying a persistence method. In \cite{li2016wind} an optimized KELM (O-KELM) method with evolutionary computation strategy is proposed for a problem of wind power prediction from time series. In this approach he structure and the parameters of the KELM are optimized by applying three different optimization meta-heuristics: GAs, Differential Evolution (DE) and Simulated Annealing. The results obtained showed that the GA-KELM algorithm outperformed the other two O-KELM approaches at future 10-minutes, 20-minutes and 40-minutes ahead prediction in terms of the RMSE value, whereas the DE-KELM obtained the best result in the 30-minutes ahead prediction problem. In \cite{huang2016hybrid} a hybrid wind speed forecasting model based on VMD, partial autocorrelation function (PACF), and weighted regularized ELM (WR-ELM), is proposed for a problem wind speed forecasting. First, the historic wind speed time series is decomposed into several IMF with the VMD. Second, the partial correlation of each IMF sequence is analyzed using PACF to select the optimal subfeature set for particular predictors of each IMF. Then, the predictors of each IMF are constructed in order to enhance its strength using the WR-ELM, and the wind speed prediction is finally obtained by adding up all the predictors. In \cite{Zheng17} a hybrid approach for wind speed prediction based on composite quantile regression Outlier Robust ELM (OR-ELM), with feature selection and parameter optimization using a hybrid population-based algorithm is developed. The hybrid methodology includes the combination of PSO and gravitational search algorithm for fine-tuning the optimal value of weights and bias of the ELM network structure, and also to carry out a feature selection to obtain the most relevant input features for the model. The effectiveness of this proposal is evaluated in two problems of wind speed forecasting at two locations of the National Renewable Energy Laboratory (NREL). 
In \cite{Peng17} a novel probabilistic wind speed forecasting model based on the combination of an OR-ELM and the time-varying mixture copula function is proposed. Case studies using the real wind speed data from the NREL are used to evaluate the performance of this hybrid approach. In \cite{TPENG17} a novel hybrid model for wind speed forecasting is proposed based on a two-stage decomposition algorithm: the complementary ensemble empirical mode decomposition with adaptive noise (CEEMDAN) and the VMD which surpass the non-linearities characteristic of wind speed time series. The former decomposes the original wind speed series into a series of IMFs with different frequencies. The latter is employed to re-decompose the IMF with the highest frequency using CEEMDAN into a number of modes successively. Then a modified AdaBoost.RT algorithm is coupled with an ELM to forecast all the decomposed modes using CEEMDAN and VMD. The developed model is tested in  four wind datasets and compared with the results of alternative methods such as a bagging approach, the partial least squares model, a MLP and a SVR, surpassing all of them in this problem. In \cite{mi2017wind} an ELM is trained for wind speed forecasting combined with wavelets methods for a first treatment of the wind speed signal. Specifically, this work presents a new hybrid method for the multi-step wind speed prediction based on wavelet domain denoising, WPD, EMD, Autorregressive Moving Average (ARMA) and the OR-ELM. The proposed approach is compared with different methods for wind speed prediction, such as ARMA, MLPs and the ELM, showing the best forecasting performance. In \cite{Yang19} a hybrid model formed by EMD, stacked auto-encoders, and ELMs for wind speed forecast is proposed. Experiments in real wind speed time series data collected from five different airports in the United Kingdom is used to evaluate the performance of this approach, comparing its results with those by other deep learning models. In \cite{Zhang19} a model based on hybrid mode decomposition (HMD) method and online sequential OR-ELM for short-term wind speed prediction. In data pre-processing period, wind speed is deeply decomposed by the HMD, which is in turn formed by a VMD, sample entropy and WPD. The crisscross algorithm is then applied to optimize the input-weights and hidden layer biases for proposed online sequential OR-ELM. In \cite{Manohar18} a discrete wavelet transform and ELM has been proposed for mode detection, fault detection/classification and section identification of wind turbines. The main objective of this approach is to protect wind turbines against wind speed intermittency due to the uncertainty in wind speed, which significantly affects the voltage-current profile. The effectiveness of the proposed approach has been validated using different statistical indices and compared with reported techniques for varying fault scenarios. In \cite{Wang18} an approach that combines ELMs with improved complementary ensemble EMD with adaptive noise and ARIMA models is proposed for short wind speed prediction problems. The ELM model is employed to obtain short-term wind speed predictions, while the ARIMA model is used to determine the best input variables for the network. Experiments in real data from three location in the Zhangye regions of the Hexi Corridor of China were used to show the goodness of the approach.
In \cite{Mahmoud18} a self-adaptive evolutionary ELM for wind power generation intervals forecasting is proposed. This hybrid approach is composed by an ELM, which constitutes the core algorithm, whose parameters evolve to become the network self-adaptative. The method estimates the potential uncertainties that may result in risk facing the power system planning, economical operation, and control. The authors compare this method in different case studies using Australian real wind farms and makes a comparison with ANNs, SVMs, and bootstrap method. In \cite{Liu18}, a new hybrid wind speed forecasting models using multi-decomposing strategy and ELM algorithm is presented. First, the wavelet packet decomposition method decomposes the input wind speed series into several sub-layers. The EMD method is used to obtain the low, high frequency sub-layers, and finally, the ELM is used to complete the wind speed predicting computation for these decomposed wind speed sub-layers. In \cite{Hu18}, a nonlinear hybrid wind speed forecasting model using LSTM network, hysteretic ELM and DE algorithm is developed. First, the performance of ELM is enhanced embedding a hysteresis process into their neuron activation function. Second, a DE approach is introduced to optimize the number of hidden layers in LSTM and number of neurons in the hidden layer. Finally, the prediction results of both regressors are aggregated by a novel nonlinear combined mechanism. This new hybrid method has been evaluated in data from a wind farm in Inner Mongolia region, China. In \cite{wang2018research} an ELM is trained for wind energy forecasting. The approach includes a data preprocesing step for estimating input data missing and use the cuckoo search algorithm, proposed to optimize the approach as a hybrid algorithm. In \cite{zhou2018short} a hybrid model combining extreme-point symmetric mode decomposition (ESMD), ELM and PSO is proposed for a problem of short-term wind power forecasting. The ESMD is applied to decompose wind power into several IMFs and then the PSO-ELM is applied to predict each IMF. Finally, the predicted values of these components are assembled into the final forecast value compared with the original wind power. Experimental results in real time series from a location in Yunnan, China, has shown the goodness of this approach. In \cite{sun2018adaptive} a hybrid model formed by an adaptive secondary decomposition, a leave-one-out cross-validation-based regularized extreme learning machine and the backtracking search algorithm, is proposed for a problem of multi-step wind speed forecasting. The proposed model has been tested on thirteen benchmark models with different time-horizon forecasting. The main goal in \cite{qolipour2019prediction} is to predict the long-term wind speed behavior, and the $24$h predictions of changes in wind speed. The proposed hybrid method combines ELM together with a Grey model $(1,1)$ as a method of Grey systems theory. Long-term wind speed are obtained using three-year data (2013-2016) for the Zanjan city (Iran), and $24$h wind speed forecast is obtained using 10-year data (2005-2015) from this city. Finally, in \cite{tian2019artificial} a short-term wind speed prediction model based on artificial bee colony algorithm optimized error minimized ELM is proposed. In this case the ELM provides the short-term wind speed prediction, with properties of fast learning speed and strong generalization ability, and the artificial bee colony algorithm is introduced to optimize the parameters of the hidden layer nodes such as the number of useless neurons, optimizing this way the efficiency of the algorithm.

\subsubsection{RF for wind speed prediction problems} 

In \cite{alonso2015random}, both RF and Gradient Boosting are proposed for wind energy prediction. This work experimentally show that both ensemble methods can improve the performance of SVR for individual wind farm energy prediction. Also, it proves that GBR is competitive for predicting wind energy in a larger geographical scale. Experiments on data from wind farms of Spain have been carried out. In \cite{lin2015seasonal} a seasonal analysis and prediction of wind energy is carried out by using RF and autorregresive with inputs (ARX) model structures. The study shows important accuracy prediction improvements by incorporating seasonal effects into the model, by including routinely measured variables,
such as radiation and pressure, and by separately predicting wind speed and direction. In \cite{Lahouar17}, RF is used for an hour-ahead wind power forecasting. First, the approach carefully select the meteorological data needed as input variables according to to a correlation and a study of importance measures. Then, a RF is trained to build an hour-ahead wind power predictor. The proposed method is tested with a real dataset taken from Sidi Daoud wind farm in Tunisia. Results show an interesting improvement of forecast accuracy using the proposed model, as well as an important reduction of the different error criteria compared to classical neural network prediction. In \cite{shi2018improved} an improved RF version is proposed for short-term wind power forecasting. Stacking at local optima and overfitting problems are avoided by proposing a 2-stage feature selection and a supervised RF. First, redundant features are removed by a variable importance measure method which selects those features with less correlation between the inputs and the output. Second, an improved RF is applied to obtain the wind power forecasting. A comparison against a back propagation neural network, Bayesian network, and SVR is carried out to show the good performance of this approach. In \cite{sun2018multistep} a novel hybrid method is proposed based on RF and ANN for wind speed and power prediction. First, the method trains an ANN to perform short-term wind speed prediction. Then, the output of the ANN is transformed into supplementary input features in the prediction process. Second, the RF is used as supervised prediction model using all these new inputs from the ANN.
In \cite{niu2018ultra} a weighted RF trained with the niche immune lion algorithm for short term wind power forecasting is proposed. The approach first develops a wavelet decomposition and use their components as inputs of a weighted RF optimized by the niche immune lion algorithm. Two empirical analyzes are carried out to prove the accuracy of the model. The first one with data acquired from a wind farm of Inner Mongolia. The second one from a wind farm on the south-east coast of China. In \cite{kaya2018hybrid} an RF for regression approach improved by EMD is proposed for a problem of hourly wind energy production forecasting. The performance of the proposed method is tested on real wind power data from an energy company in Turkey. In \cite{natarajan2020wind} a RF algorithm is tested in a problem of wind power forecasting. The prediction results obtained are close with the actual wind power generated at the wind farms considered, and it is shown that RF obtains better performance in this problem than ANNs. In \cite{vassallo2020analysis} three different RF-based approaches have been tested and analyzed in a problem of wind speed prediction. First, using a standalone RF model versus using RF as a correction mechanism for the persistence approach, second utilizing a recursive versus direct multi-step forecasting strategy, and third training data availability on model forecasting accuracy from one to six hours ahead. These approaches are tested on data from the FINO1 offshore platform and Atmospheric Radiation Measurement Southern Great Plains C1 site. The obtained results have been compared in this case with the persistence method.

\subsubsection{ESNs for wind speed prediction problems} 

In \cite{Liu15Echo}, a short-term wind speed predictor is proposed based on spectral clustering and optimised ESN. First, a wavelet transformation is used to decompose the wind speed into multiple series which analyzed with PCA for avoiding redundant information. Then, an ESN is used to simultaneously predict multiple outputs, and a genetic algorithm was employed to optimise the ESN parameters. Data from a wind farm in northern China is used for the training. 
In \cite{dorado2017robust} the prediction of ramp-events in wind farms is taken into account, by means of an ESN. Specifically, 6-h and 24-h binary (ramp/non-ramp) prediction based on an ESN is proposed. Experiments in three different wind farms in Spain show the potential of ESN approaches in this problem of wind ramps events prediction.
In \cite{lopez2018wind}, a hybrid approach is proposed based on combining an ESN and a LSTM for wind power prediction.
The hybrid architecture is quite similar to an ESN, but using LSTM blocks as units in the hidden layer. The training process has two steps: First, for each epoch, the hidden layer is trained with a gradient descend method. Second, the output layer is tuned with a quantile regression. The approach has been test on data from the Klim Fjordholme wind farm. An ESN for wind speed and direction prediction is developed in \cite{chitsazan2019wind} introducing nonlinear functions. This paper proposes two methods for wind speed and direction forecasting based on the nonlinear relations between the internal states of ESNs. The methods decrease the number of internal states, and reduce the computational load compared to classical ESNs with fixed sizes and topologies by reducing the orders of the weight matrices. The design is simple with high learning capability and prediction accuracy, and does not require extensive training, parameter tuning, or complex optimization. The methods are tested with wind speed and direction data provided by the Nevada department of transportation roadway weather stations in Reno, NV, USA. To demonstrate the efficiency of the proposed methods, they are compared with classical ESN and with adaptive neuro-fuzzy inference system (ANFIS). The results of the new methods compare favorably with both ESN and ANFIS. In \cite{Wang19Echo} a new wind power prediction method with high accuracy based on ESNs is proposed. The method is a hybrid of wavelet transform, ESN and ensemble technique. First, a wavelet transform is used to decompose raw wind power time series data into several time series with different frequencies. Then, ESN is applied to automatically learn the nonlinear mapping relationship between input and output in each frequency. Later, an ensemble technique is applied to deal with the model misspecification and data noise problems. Experiments in wind power data from real wind farms in Belgium and China have been used to verify the performance of the method. In \cite{wang2019novel} a hybrid method for wind speed forecasting based on a preprocessing step, and an ESN optimized with a multi-objective optimization approach is proposed. The preprocessing method can obtain a smoother input by decomposing and reconstructing the original wind speed series. Then, the ESN, optimized by a multi-objective optimization algorithm is developed as a forecasting method. Finally, eight datasets with real data of wind speed obtained at different points of the world are used to validate the performance of the proposed system. In \cite{Chen19Echo} a novel combined model for wind speed forecasting is proposed. Different decomposition algorithms and an ESN form the proposal. The goodness of this approach is tested in wind speed data at different heights from the M2 tower from the National Wind Power Technology Center, United States, are utilized as case studies. The experimental results show that this combined method performs better than other conventional methods for wind speed prediction. Finally, in \cite{Hu20Echo} a dESN for forecasting energy consumption and wind power generation by introducing the deep learning framework into the basic ESN. dESN combines the powerful nonlinear time series modeling ability of echo state network and the efficient learning ability of the deep learning framework. Examples in energy consumption prediction in China and wind energy generation in the region of Inner Mongolia are considered to show the performance of this approach. 

\subsubsection{RVFL for wind speed prediction problems} 

In \cite{ren2015detecting} an algorithm to detect the wind power ramps in a certain forecasting horizon based on RVFL is presented. This method is tested in a real world wind power data set, comparing the results obtained with that of RF, SVM and MLP with classical training. In \cite{nhabangue2018wind} a RVFL algorithm for wind speed prediction is presented. The model applies the Chebyshev expansion to transform the direct links of the RVFL. An ensembled version of this proposal is also discussed, in which the EMD is applied to the input signal. Two wind speed datasets are used for performance comparison with other models. In \cite{Ahmed18} a method for the development of multi-step wind forecasting models based on RVFL is proposed. Experiments in real wind data from Dodge city, Kansas, USA, are carried out to test the performance of the RVFL in this particular problem of wind speed prediction. In \cite{mishra2019short} a hybrid pseudo-inverse Legendre neural network (a RVFL-based approach) with and adaptive firefly algorithm is proposed for short-term wind power forecasting. Hidden layers are configured with radial basis function units and the random input weights between the expanded input layer using Legendre polynomials, and the RBF units in the hidden layer are optimized with the metaheuristic firefly algorithm for error minimization. Data from wind farms of the states of Wyoming and California (USA) and Spain are used as case of study. In \cite{jalli2020prediction} a hybrid short-term wind speed forecasting approach is proposed combining the empirical mode decomposition technique with RVFL. The empirical mode decomposition divides the wind speed input data into a fixed number of intrinsic mode function. These intrinsic features feed the RVFL which is trained with the Levenberg-Marquart algorithm.

\subsection{Marine and ocean energy}

\subsubsection{ELMs in marine and ocean energy}

In \cite{Alexandre15} the problem of locally reconstructing significant wave height at a given point by using wave parameters from nearby buoys is tackled. The proposed approach is based on the spatial correlation among values at neighboring buoy locations. A hybrid Genetic algorithm with a ELM is proposed to solve this problem. Experiments at two different locations of the Caribbean Sea and West Atlantic showed the good performance of this hybrid approach. In \cite{Cornejo16} and \cite{cornejo2016grouping} another hybrid approach for feature selection is proposed and analyzed in two different relevant problems for marine energy applications: significant wave height and wave energy flux prediction. Specifically, a hybrid GGA-ELM is proposed, in such a way that the GGA searches for several subsets of features, and the ELM provides the prediction of the significant wave height and energy flux. Experiments in data from buoys in California, USA, are carried out. In \cite{Cornejo18Bayes} an updated of the previously described hybrid GGA-ELM approach with Bayesian Optimization was proposed. The idea is that the Bayesian approach was used to obtain the optimal parameters of the prediction system (GGA-ELM), such as probability of crossover, probability of mutation, number of neurons in the hidden layer, regularization constant, etc. The results obtained showed an improvement of over 6\% in the quality of the solutions found by the GGA-ELM with respect to the same algorithm without Bayesian optimization. In \cite{Kumar18} an ensemble of ELMs is proposed to predict the daily wave height at different stations. The approach consists of constructing an Ensemble of ELM, with the parameters of each ELM initialized in distinct regions of the input space. Results in different stations of Gulf of Mexico, Brazil and Korean region have been reported. In \cite{Ali19}, an ELM model integrated with an improved ensemble empirical mode decomposition is presented to predict significant wave height. This ELM is trained with data from the eastern coastal zones of Australia. The improved ensemble empirical mode decomposition method incorporates the historical lagged series of significant heights as the model's predictor to forecast future significant  heights. In \cite{Kaloop20} a hybrid algorithm formed by wavelet, PSO and ELM methods is proposed for estimating significant wave height in coastal and deep-sea stations. Data obtained from buoys situated off the south-east coast of the US are used to test the behaviour of the algorithm. A comparative analysis with ELM, KELM, and PSO-ELM models with and without wavelet integration was carried out. Finally, in \cite{shamshirband2020prediction} the efficiency of three ML methods, MLPs, ELMs and SVR in a problem of wave height modeling is compared. The results show that the ELM slightly outperforms MLPs and SVRs in this prediction problem.

\subsubsection{RF in marine and ocean energy}

There are only a few randomization based approaches based on RF for marine energy prediction problems. A combined RF and physics-based models is presented in  \cite{Serras19} to predict the electricity generated by ocean waves. The work proposes RF models combined with physics-based models to forecast the electricity output of the Mutriku wave farm on the Bay of Biscay (Spain). The RF is train with three sets of inputs: the electricity generated, the wave energy flux at the nearest gridpoint, and ocean and atmospheric data. In \cite{callens2020using}, RFs and Gradient boosting trees are used to improve significant wave height forecasting. Hyperparameter values of these ML algorithms are tuned using bayesian optimization. The ML methods are trained with data from Biarritz (France). The predictions made by Rf and gradient boosting trees provide better results than the ANNs.

\subsection{Hydro-power}

\subsubsection{ELMs in hydro-power prediction}

In \cite{li2014monthly}, a combination of wavelet neural networks with ELMs for 1-month-ahead discharge forecasting is proposed. The ELM is combined in this case with feed-forward neural networks to improve their prediction capabilities. The model has been tested in data acquired from two reservoirs in southwestern China. The work in \cite{atiquzzaman2018robustness} focuses on the robustness of ELMs for hydrological flow series forecasting. The work compares ELMs for hydrological flow series modeling with other ML methods, such as genetic programming and evolutionary computation based SVM. The robustness is evaluated varying number of lagged input variables, the number of hidden nodes and input parameters. In \cite{Yaseen19} an ELM is trained for river flow forecasting. The approach applies an orthogonal decomposition to tune the output-hidden layer of the ELM model. The proposed method is tested with data from Kelantan River, Malaysia, selected as a case study. The problem of operation rule derivations for optimal scheduling of hydro-power reservoir is studied in \cite{Feng19} using the k-means clustering method and ELM optimized with a PSO. First, a k-means clustering method is adopted to split all the influential factors into several disjointed sub-regions. Then, an enhanced ELM optimized with a PSO is proposed to classify these regions. The proposal is tested over two water reservoirs in China. In \cite{Wang20} an ELM is proposed for production capacity forecasting of hydro-power industries. The work trains an ELM based on the data expansion generated by a Monte Carlo algorithm. This method is proposed to deal with the common lack of data presented in the production capacity forecasting of hydro-power. In \cite{Ribeiro20} a multi-objective ensemble of ESNs and ELMs are used for streamflow series forecasting in planning electric energy production for hydro-electric plants. The work proposes the development of ensembles of unorganized machines, specifically ELMs and ESNs, which work as simple learners of a bagging method. The approach employs multi-objective optimization to select and adjust the weights of the ensembles models, taking into account the trade-off between bias and variance. The proposed method is tested with data from five real-world Brazilian hydro-electric plants.
The authors in \cite{niu2019comparison} make a comparison among different ML methods for deriving operation rule of hydro-power. Specifically, multiple linear regression, ANNs, ELMs, and SVMs are evaluated. Data acquired from Hongjiadu reservoir of China between 1952 to 2015 are used for training and tested the methods. In \cite{lian2020trend} a trend-guided ELM model for hydro-power prediction is proposed, which enhanced the physical mechanism of the input-output of the traditional ELM. First, the model smooths and repairs abnormal output data of the small hydro-power group before extracting the trend of power change. This trend constitutes the input of the ELM. Data for training and tested the approach has been extracted from hydro-power groups in Guangxi. Finally, in \cite{Chen20} an adaptive KELMs is proposed for prediction and monitoring of dam leakage flow. The parameters of this KELM are adaptable, optimized by parallel multi-population Jaya algorithm. Besides, a sensitivity analysis is used to evaluate the relative importance of each input variable. The KELM model has been trained and evaluated in data from the Tryggevaelde (Denmark) and Mississippi (USA) rivers.

\subsection{Critical literature analysis}\label{critical_literature_analysis}

As a final result of this literature review, we herein highlight some important trends and patterns found in the comprehensive literature analysis carried out in preceding sections:
\begin{itemize}[leftmargin=*]
    \item First, it is quite remarkable that the large majority of works dealing with randomization-based algorithms in renewable energy were proposed from 2014 onward. There are very few works prior to 2014 dealing with randomization-based techniques in renewable energy prediction problems, even though the majority of these techniques were proposed far before. This indicates an uprising interest in the last years in the application of these techniques, mainly fostered by their good balance between predictive performance and associated computational cost.
    
    \item Clear trends exist regarding the application of specific randomization-based techniques to renewable energy prediction problems, but the subset of studied techniques largely depends on the resource under consideration. As such, ELM-based approaches clearly emerges as the most utilized randomization-based approach, laying at the core of different applications in all the renewable resources reviewed in our survey. However, their prominence is specially notable in solar and wind, where ELMs have been massively applied in the last years, either on its own, in comparison with alternative neural computation techniques, or as a part of hybrid approaches combined with other algorithms such as evolutionary algorithms or signal decomposition techniques (e.g. EMD).
    
    \item RF has also been recurrently utilized in the majority of prediction problems formulated for different renewable resources. It has been very often applied to prediction problems in solar and wind energy, and also in some applications related to marine energy prediction problems. The robustness of this class of ensemble algorithms to overfitting and their easy hyper-parametric configuration are the main cause for the success of RF in these problems.
    
    \item Other randomization-based algorithms such as RVFL and ESN have been applied to different prediction problems, mainly in solar radiation and wind resources. These modeling approaches have received growing attention and praise in the last years once different contributions have reported a superior performance of these approaches in the aforementioned renewable energy domains. It is worth mentioning that the persistence of wind and solar radiation lays a favorable scenario for the recurrent modeling capabilities of ESN models, which we will later showcase in the experimental part of our study.
    
    \item Deep and ensemble versions of different randomization-based techniques have also been very recently applied with success in prediction problems associated to renewable energy sources. In the experimental part of this paper we show the performance of the deep version of RVFL in comparison with randomization-based techniques and alternative ML approaches, offering informed evidence that stacking more layers/reservoirs can lead to better performance scores without major computational penalties.
\end{itemize}

Besides the specific modeling and application trends exposed above, the literature corpus analyzed in this section has revealed that many studies undergo methodological or very little novelty added over the state of the art. In some cases, similar modeling approaches are just applied to different datasets, without further advances in terms of problem/techniques rather than a na\"ive experimentation of existing models to new data. In this same line of reasoning, very few works include a rigorous analysis of the statistical significance of the performance gaps reported among the techniques in the benchmark, which should be mandatory when part of the models under comparison is drawn at random. Finally, most published studies are completely driven towards finding the method that performs best (as per predictive performance, e.g. precision of forecasts). This leaves aside other factors of utmost importance for making decisions in practice based on the predictions issued by the models. Alternative aspects such as the explanations of the predictions (i.e. what input variables impacted most on the output, what should have occurred for the output to change to a certain level), the compliance of the output with the physics governing the scenario where the learning problem is formulated, the computational efficiency of every model, or the estimation of the epistemic uncertainty induced by the model in its output are largely overseen in the literature. We will later revolve around these noted lacks in the literature, which in our vision should be among the research priorities of the community in years to come.

In what follows we complement our critical literature review with a robust experimental evaluation of randomization-based techniques in diverse prediction problems involving different resources prediction tasks. As shown through the experiments, we go beyond the discussion of the predictive performance of the models under consideration towards verifying the relevance of the performance gaps based on statistical tests and the computational efficiency of the models. Reporting also on these figures should be enforced for future contributions leveraging the benefits of randomization-based machine learning approaches.

\section{Experiments and results}\label{Experiments_Results}

In this section we show the potential of randomized-based ML approaches in several prediction problems in renewable energy. We first tackle a problem of global solar radiation estimation from satellite data. We also deal with a problem of hourly wind speed prediction in three different wind farms, and finally we tackle a problem of dammed water level prediction using data from a real hydro-power installation. Table \ref{tab:randomizationConf} lists the values of the hyper-parameters explored for every randomization-based ML model considered in the experiments. Hyper-parameter values for other non-randomization based modeling counterparts are retrieved from the original study where they were first proposed, which is cited in the description and discussion of every problem.
\begin{table}[H]
\centering
\caption{Hyper-parameter values considered for every randomization-based ML model considered in the experiments.}
\vspace{0.3cm}
\label{tab:randomizationConf}
\resizebox{\columnwidth}{!}{\begin{tabular}{cccc}
\toprule
Model & Hyper-parameter & Meaning & Values \\
\midrule
\multirow{2}{*}{\makecell{RF, AdaBoost}} & N & Number of weak learners (trees) & [10,30,50,100,200,300] \\
 & maxDepth & Maximum allowed depth for trees & [3,5,7,9,unbounded] \\
\midrule
\multirow{2}{*}{\makecell{ELM, ML-ELM, RVFL,\\multiRVFL, dRVFL, edRVFL}} & K & Number of layers & [1,2,3,4,5] \\
 & $\tilde{N}$ & Number of neurons per layer & [10,20,30,50,100,200,300] \\
\midrule
\multirow{4}{*}{ESN, dESN} & K & Number of reservoirs & [1,2,3,4] \\
 & $\tilde{N}$ & Number of neurons per reservoir & [10,20,50,100,200,300] \\
 & $\rho_{max}$ & Spectral radius & [0.7,0.8,0.9,0.99] \\
 & $\lambda$ & Connectivity of reservoir neurons & [0.5,0.7,0.9,1.0] \\
\bottomrule
\end{tabular}}
\end{table}

\subsection{Solar radiation estimation}

The first problem we tackle is a problem of solar radiation estimation from satellite and numerical models data. Here we present novel results of randomization-based regression approaches, comparing the results obtained with those of previous approaches dealing with the same problem in \cite{Cornejo19} (alternative regression techniques)  and \cite{guijo2020evolutionary} (evolutionary neural networks). We first describe the problem, including the dataset available (objective and input variables), and then we discuss the results obtained with the randomization-based techniques and algorithms for comparison.

\subsubsection{Dataset description}\label{sec:dataset_solar}
We consider real data of solar radiation at the radiometric station of Toledo, Spain ($39^\circ$ 53'N, $4^\circ$ 02'W). As objective variable for the prediction, one year of hourly global solar radiation data was available for this study (from the 1st of May 2013 to the 30th of April 2014) measured with a Kipp \& Zonen CM-11 pyranometer. Regarding the predictive (input) variables, we have selected the following:
\begin{itemize}[leftmargin=*]
\item Reflectivity information acquired from the Spinning Enhanced Visible and Infrared Image (SEVIRI) visible spectral channels 0.6 $\mu$m (VIS 0.6) and 0.8 $\mu$m (VIS 0.8). SEVIRI \cite{Schmid00} is the main instrument on board Meteosat Second Generation Satellites. This instrument has a nominal spatial resolution at the sub-satellite point is 1 km$^2$ for the high-resolution channel, and 3 km$^2$ for the other channels \cite{Aminou02,Schmetz02}. Reflectivity values are obtained at LATUV Remote Sensing Laboratory (Universidad de Valladolid, Spain) from Level 1.5 image data considering the Sun's irradiance, the Sun's zenith angles and the Earth-Sun distance for each day.

After applying a closeness criterion (minimum Euclidean distance considering latitude and longitude values) to determine which pixel is the closest to Toledo's station, we have extracted the reflectivity information from the nearest pixel to this location as well as the reflectivity information from the eight pixels surrounding the central one. This way, 18 reflectivity values (9 for the VIS 0.6 and 9 for the VIS 0.8 channels) every 15 minutes were available. Since global solar radiation data at Toledo's station were only available at 1-hour temporal resolution (from 05:00h to 20:00h), we calculated the mean hourly value for each of the 18 values considered. Finally, with the VIS 0.6 and 0.8 reflectivity information we have calculated the {\em cloud index} following the Heliosat-2 method \cite{Rigollier04}, as shown in \cite{Cornejo19}.

\item Information from the Copernicus Atmosphere Monitoring Service (CAMS) \cite{Schroedter16} (freely available from the Internet). This model is mainly based on two state-of-the-art methodologies: a novel clear sky model developed by \cite{Lefevre13} to calculate both the global surface irradiance and the beam/direct surface irradiance for clear sky on a horizontal plane, and an improved version of the previously described Heliosat methodology, named Heliosat-4 \cite{Qu17}, to estimate the downwelling shortwave irradiance received at ground level in all sky conditions.

\end{itemize}

The complete list of input and target variables are summarized in Table \ref{tab:databaseSets}. Note that a total of 39 input values (18 reflectivity values, 18 values for the cloud index, the clear-sky value for Toledo station, the Heliosat-2 and CAMS-based models) are considered in this prediction problem.

\begin{table}[H]
\begin{center}
\caption{Predictive and target variables considered for the solar radiation prediction problem. Reflectivity and cloud index are calculated for 9 points surrounding the measuring station, for VIS 0.6 and VIS 0.8 channels (18 input variables each).}
\vspace{0.3cm}
\label{tab:databaseSets}
\resizebox{0.6\columnwidth}{!}{\begin{tabular}{cc}
\toprule
	 Predictive variables		& Units \\
\midrule
Reflectivity (VIS 0.6 and VIS 0.8 channels)           & [\%]\\
Cloud index\           & [\%]\\
Clear sky radiance\    & [W/m$^2$]\\
CAMS solar radiation  \ & [W/m$^2$]\\
SolarGIS solar radiation   & [W/m$^2$]\\
(Target) Global solar radiation &[W/m$^2$]\\
\bottomrule
\end{tabular}}
\end{center}
\end{table}

\subsubsection{Experimental results}\label{sec:Results_solar}

Table \ref{tab:Results_Solar} shows the results obtained in the solar radiation prediction problem with the randomization-based techniques discussed in this work. Results are reported in terms of the Mean Bias Error (MBE), the Mean Absolute Error (MAE), the Root Mean Square Error (RMSE) and the Pearson's Correlation Coefficient ($R^2$). These are widely utilized metrics for gauging the performance of regression models, and quantify different aspects of the error incurred by the model over a test set. To begin with, MAE and RMSE link directly with the average  amplitude of the error. By contrast, the MBE metric links to the direction of the error bias, such that a negative MBE value informs that the prediction is in general biased towards smaller values than the real values of the target variable. Finally, $R^2$ indicates the extent to which the predicted values and the real values of the target variable are linearly correlated to each other.

A comparison with alternative regression techniques in previous works (\cite{Cornejo19} and \cite{guijo2020evolutionary}) is carried out. We adopt the notation used in this latter reference when denoting the different evolutionary neural approaches: A-B, where A denotes the type of activation function used in the hidden layer of the neural network, and B that selected for the output layer. Three combinations were evaluated in this work: SU-LO, RBF-LO and SU-PU, where SU stands for \emph{Sigmoid Unit}, RBF denotes \emph{Radial Basis Function}, PU is \emph{Product Unit} and LO corresponds to \emph{Linear Output}. Further details can be found in \cite{guijo2020evolutionary}. 

As can be seen in the table, the best result over all is obtained by the dRVFL approach, with an excellent MAE value of $36.82~W/m^2$  and a $R^{2}=0.968$. This result outperforms all the rest of randomization-based approaches, and all alternative regression used for comparison, including statistical ML regressors such as SVR or GPR, neural networks and evolutionary neural approaches, i.e.  Sigmoid Unit- Radial Basis Function - Product Unit. Note that, in general, all randomization-based approaches tested work really well in this problem, obtaining values of MAE under $40~W/m^2$ with $R^{2}>0.96$, but the Adaboost regression (AdaBoost) approach, which showed a slightly poorer performance than the other randomization-based approaches. These results show that the randomization-based approaches are fully competitive against evolutionary neural networks approaches, which usually show a much larger computation time due to their training process difficulty. The randomization-based algorithms clearly outperformed statistical learning theory regressors such as SVR and GPR.
\begin{table}[h!]
\centering
\caption{\label{tab:Results_Solar} Results obtained in the solar radiation prediction problem considered with the randomization-based techniques discussed in this paper, compared with alternative regression techniques in \cite{Cornejo19} and \cite{guijo2020evolutionary}.}
\vspace{0.3cm}
\resizebox{0.9\columnwidth}{!}{
\begin{tabular}{lccccc}
\toprule
\multirow{2}{*}{Model} & \multicolumn{4}{c}{Best RMSE configuration} & \multirow{2}{*}{\makecell{$\overline{\text{RMSE}}$ \\ {[}$W/m^2${]}}} \\
\cmidrule{2-5}
& MBE [$W/m^2$] & MAE [$W/m^2$] & RMSE [$W/m^2$] & $R^{2}$ & \\
\midrule
GPR (\cite{Cornejo19})    &-8.66   &41.22 \     &74.45   &0.9413 &-   \\
MLP (\cite{Cornejo19})  &-2.47        &55.29 \    &79.09     &0.9326 & -   \\
SVR (\cite{Cornejo19})    &11.22        &56.68 \     &85.04    &0.9258& -    \\
SU-LO (\cite{guijo2020evolutionary}, Conf. 1) & 2.26 & 39.17 & 59.06 & 0.9622 & 60.73 $\pm$ 0.85 \\
RBF-LO (\cite{guijo2020evolutionary}, Conf. 1) & 2.70 & 37.33 & 57.07 & 0.9648 & 62.31 $\pm$ 4.79 \\
SU-PU (\cite{guijo2020evolutionary}, Conf. 1) & 1.58 & 38.64 & 58.41 & 0.9630 & 59.99 $\pm$ 0.90 \\
\midrule
RFR & 5.43 & 37.87 & 59.73 & 0.9613 & 61.38 $\pm$ 0.81 \\
AdaBoost & 7.44 & 58.75 & 77.15 & 0.9354 & 80.06 $\pm$ 1.24 \\
ELM & 3.14 & 36.04 & 57.26 & 0.9645 & 60.78 $\pm$ 2.56 \\
ML-ELM & 0.41 & 38.35 & 56.26 & 0.9656 & 59.93 $\pm$ 1.45\\
RVFL & -0.81 & 39.43 & 56.09 & 0.9659 & 58.44 $\pm$ 1.21\\
multiRVFL & 0.87 & 37.93 & 55.34 & 0.9668 & 58.06 $\pm$ 1.14\\
dRVFL & 0.70 & 36.82 & 54.36 & 0.9679 & 56.55 $\pm$ 1.13 \\
edRVFL & 0.01 & 38.05 & 55.76 & 0.9663 & 57.45 $\pm$ 0.79 \\
\bottomrule
\end{tabular}}
\end{table}

\begin{figure}[!ht]
    \centering
      \subfigure[]{
        \label{fig:RFRscatter}
        \includegraphics[width=0.31\textwidth]{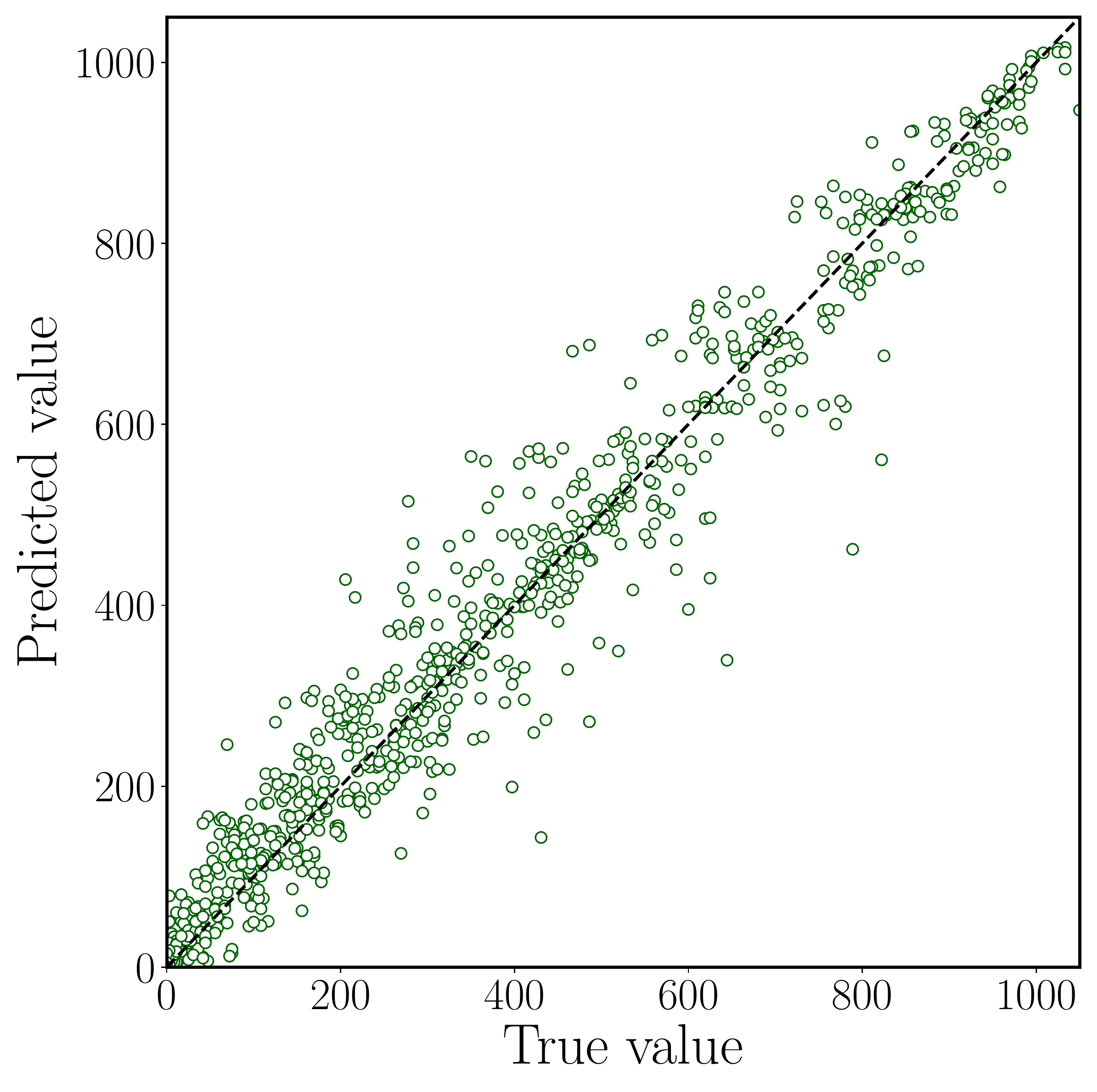}}
      \subfigure[]{
        \label{fig:multiELMscatter}
        \includegraphics[width=0.31\textwidth]{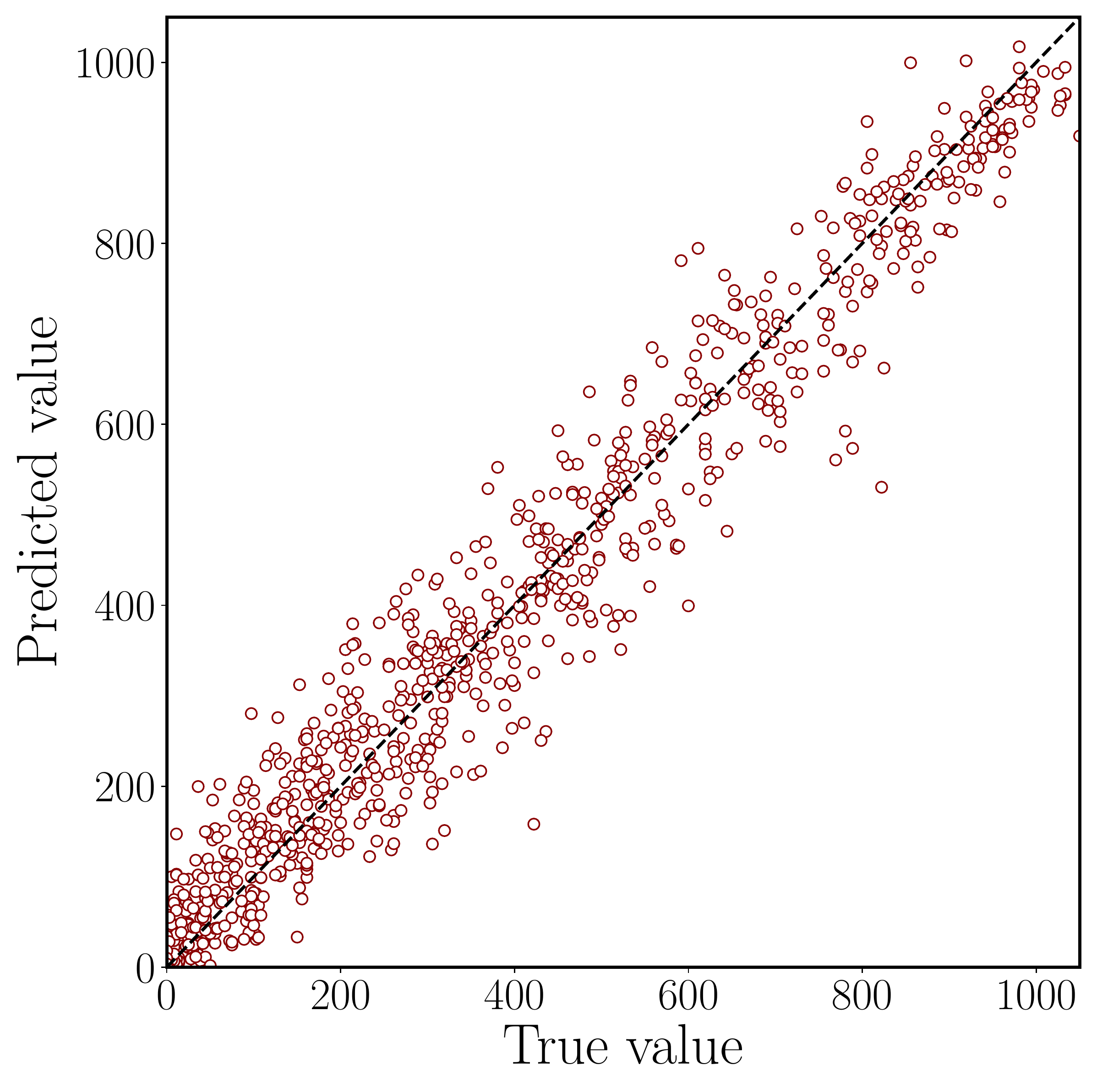}}
      \subfigure[]{
        \label{fig:DeepRVFLscatter}
        \includegraphics[width=0.31\textwidth]{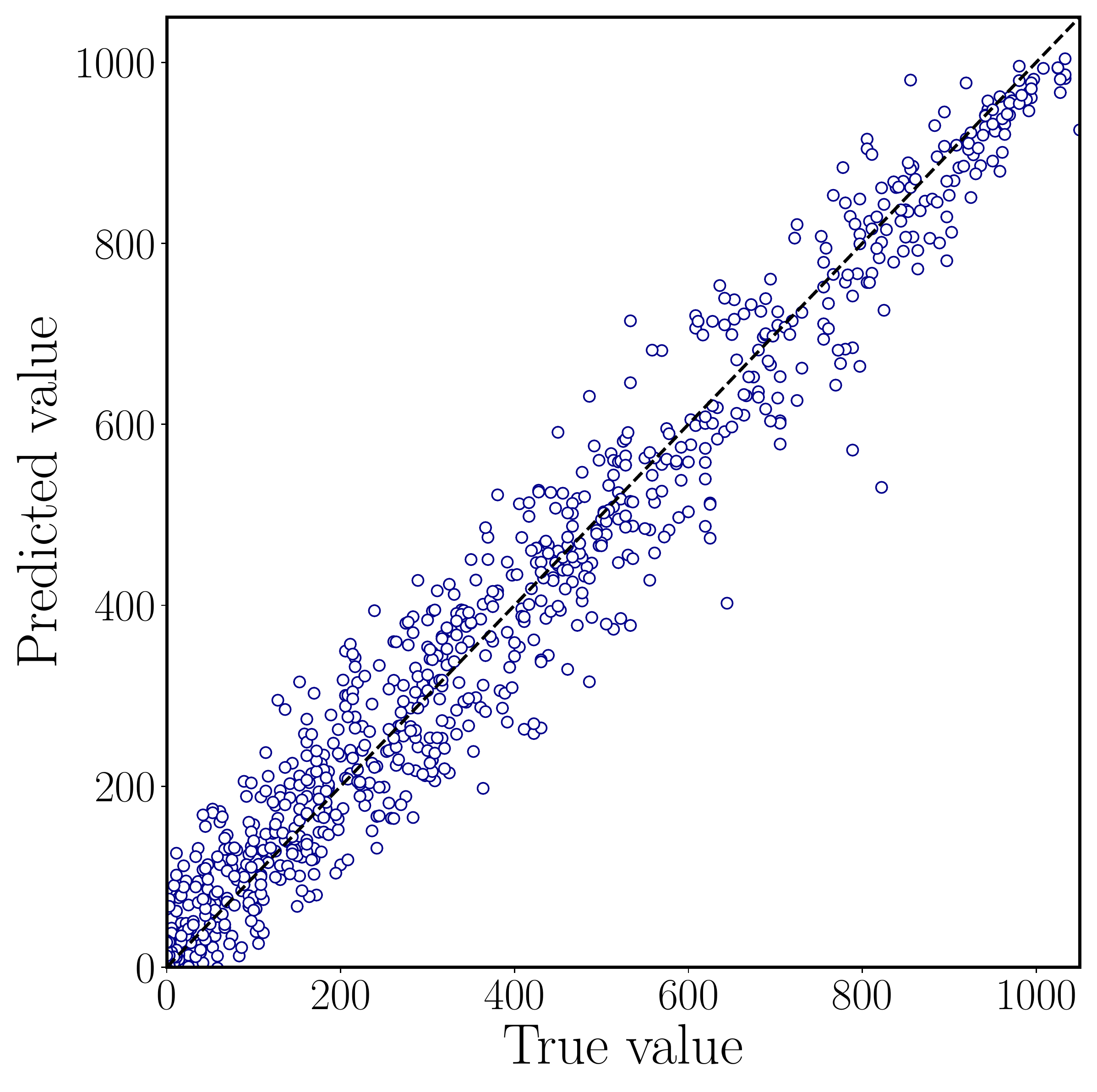}}
    \caption{Scatter plots of true versus predicted values corresponding to the best RMSE configuration of models RFR; (b) ML-ELM; (c) dRVFL.}
    \label{fig:scatterSolar}
\end{figure}

Figure \ref{fig:scatterSolar} shows an example of prediction result by the best algorithms tested in this work (RFR, ML-ELM and dRVFL). Specifically, scatter plots (ground truth vs. predicted value) are shown to illustrate the good performance of the randomization-based approaches discussed in this paper. As can be seen, the dRVFL obtains a scatter plot slightly more compact that their counterparts, leading to a very good prediction in general, and to obtain improvements over the RFR and ML-ELM approaches in specific samples.

To complete this study, Table \ref{tab:Wilcoxon_solar} shows the p-values of the Wilcoxon Rank-Sum tests between the RMSE values of every couple of randomization-based algorithms considered in this paper. Lack of statistical significance at $\alpha = 0.05$ (meaning that the algorithms performance cannot be considered as different) is highlighted as shadowed values. As can be seen, there are statistically significant differences between the performance of majority of the algorithms in this problem, but RVFL with ELM and multiRVFL with RVFL, indicating in these cases that the performance of these pairs of algorithms is statistically similar. Therefore, these results indicate that the dRVFL approach is the best-over-all randomization-based algorithm tested in this problem, obtaining statistically significant different with the performance of the rest of algorithms tested in this problem.

\begin{table}[!ht]
\centering
\caption{\label{tab:Wilcoxon_solar} p-values of the Wilcoxon Rank-Sum tests between the RMSE values of every couple of randomization-based algorithms considered in this paper. Lack of statistical significance at $\alpha=0.05$ is highlighted as shadowed cells in the table.}
\vspace{0.3cm}
\resizebox{\columnwidth}{!}{
\begin{tabular}{ccccccccc}
\toprule
p-value &
RFR & AdaBoost & ELM & ML-ELM & RVFL & multiRVFL & dRVFL & edRVFL\\
\midrule
RFR & --- & --- & --- & --- & --- & --- & --- & --- \\
AdaBoost & 2.872e-11 & --- & --- & --- & --- & --- & --- & --- \\
ELM & 2.872e-11 & 2.872e-11 & --- & --- & --- & --- & --- & --- \\
ML-ELM & 2.872e-11 & 2.872e-11 & 2.760e-02 & --- & --- & --- & --- & --- \\
RVFL & 2.872e-11 & 2.872e-11 & \cellcolor[gray]{0.8}1.515e-01 & 8.795e-04 & --- & --- & --- & --- \\
multiRVFL & 2.872e-11 & 2.872e-11 & 3.847e-02 & 1.286e-04 & \cellcolor[gray]{0.8}7.007e-01 & --- & --- & --- \\
dRVFL & 2.053e-10 & 2.872e-11 & 1.369e-08 & 6.415e-10 & 4.915e-06 & 3.974e-06 & --- & --- \\
edRVFL & 2.872e-11 & 2.872e-11 & 1.662e-07 & 4.373e-09 & 7.101e-04 & 6.036e-04 & 8.135e-03 & --- \\
\bottomrule
\end{tabular}}
\end{table}

\subsection{Wind speed forecasting}

The second problem tackled in this paper is devoted to wind speed prediction from time series. As previously mentioned, the intermittent and volatile nature of wind speed makes extremely important its prediction in large generation facilities such as wind farms \cite{tascikaraoglu2014review}. Long-term wind speed prediction usually involves numerical weather methods to take into account the atmospheric state and dynamics. However, in short-term prediction problems (up to 8 hours prediction time-horizon), statistical methods such as ML regression can be applied with good results at a very reduced computational cost. In this case study we test the performance of randomized-based regression technique in a short-term wind speed prediction in real data from three wind farms in Spain. We first describe the data available, and then the results obtained with the discussed randomization-based techniques, in comparison with alternative ML regression techniques.

\subsubsection{Dataset description}

\begin{figure}[!ht]
\begin{center}
\includegraphics[draft=false, angle=0,width=8cm]{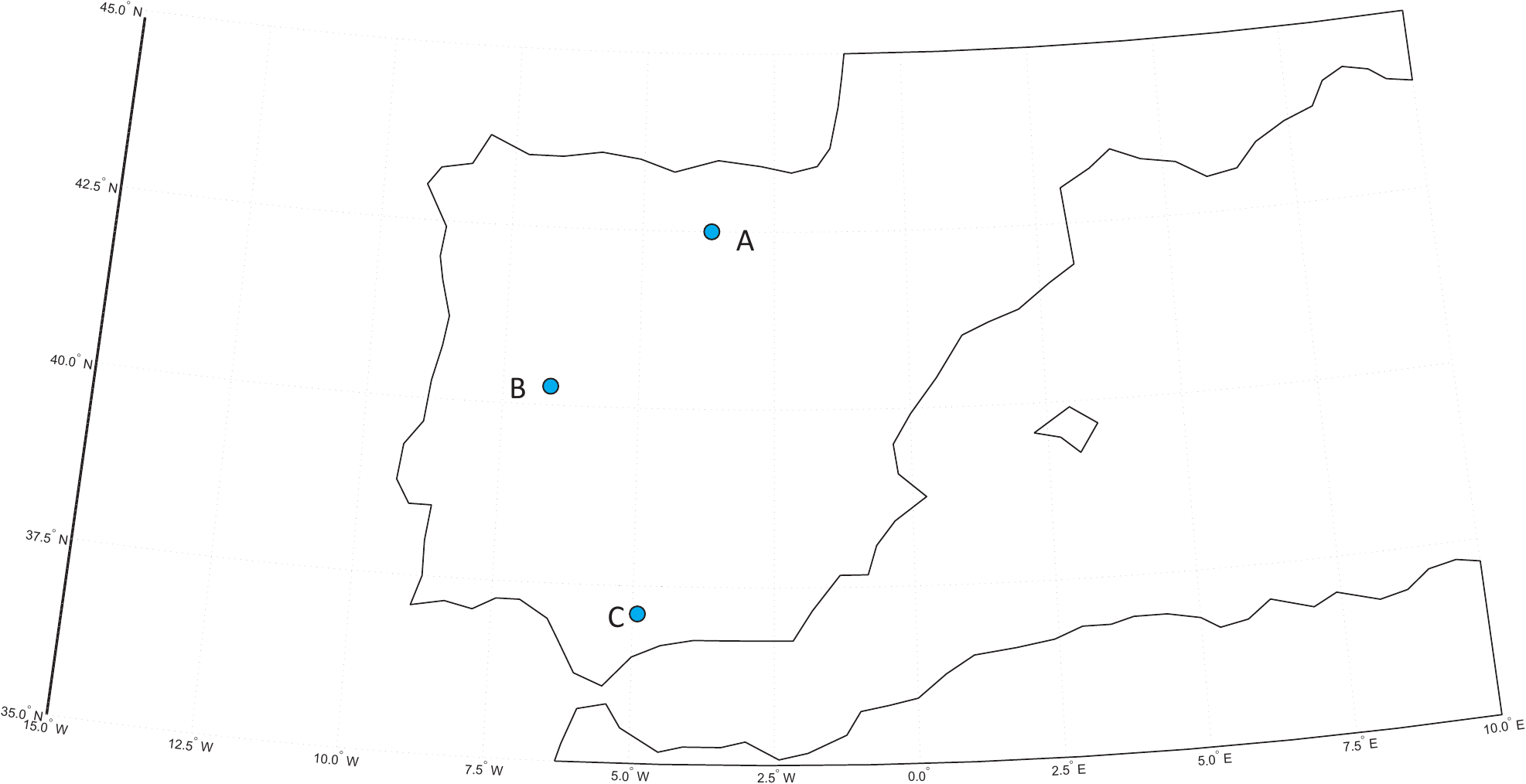}
\end{center}
\caption{\label{fig:maps_WF} Location of the three wind farms considered in Spain.}
\end{figure}

Hourly wind speed data from three wind farms in Spain (A, B and C in Figure \ref{fig:maps_WF}) have been considered in this study. Note that the wind farms selected cover different parts of Spain, north, center and south, characterized by different wind regimes. Different data lengths are available for each wind farm: in wind farm A, data from 11/01/2002 to 29/10/2012, in wind farm B, from 23/11/2000 to 17/02/2013 and in wind farm C from 02/03/2002 to 30/06/2013. Note that in all cases they are very long hourly wind speed time series, what allows testing different ML algorithms for wind speed prediction in different forecasting time-horizons. 

\begin{figure}[!ht]
    \centering
      \subfigure[]{
        \label{Mon_mod}
        \includegraphics[width=0.31\textwidth]{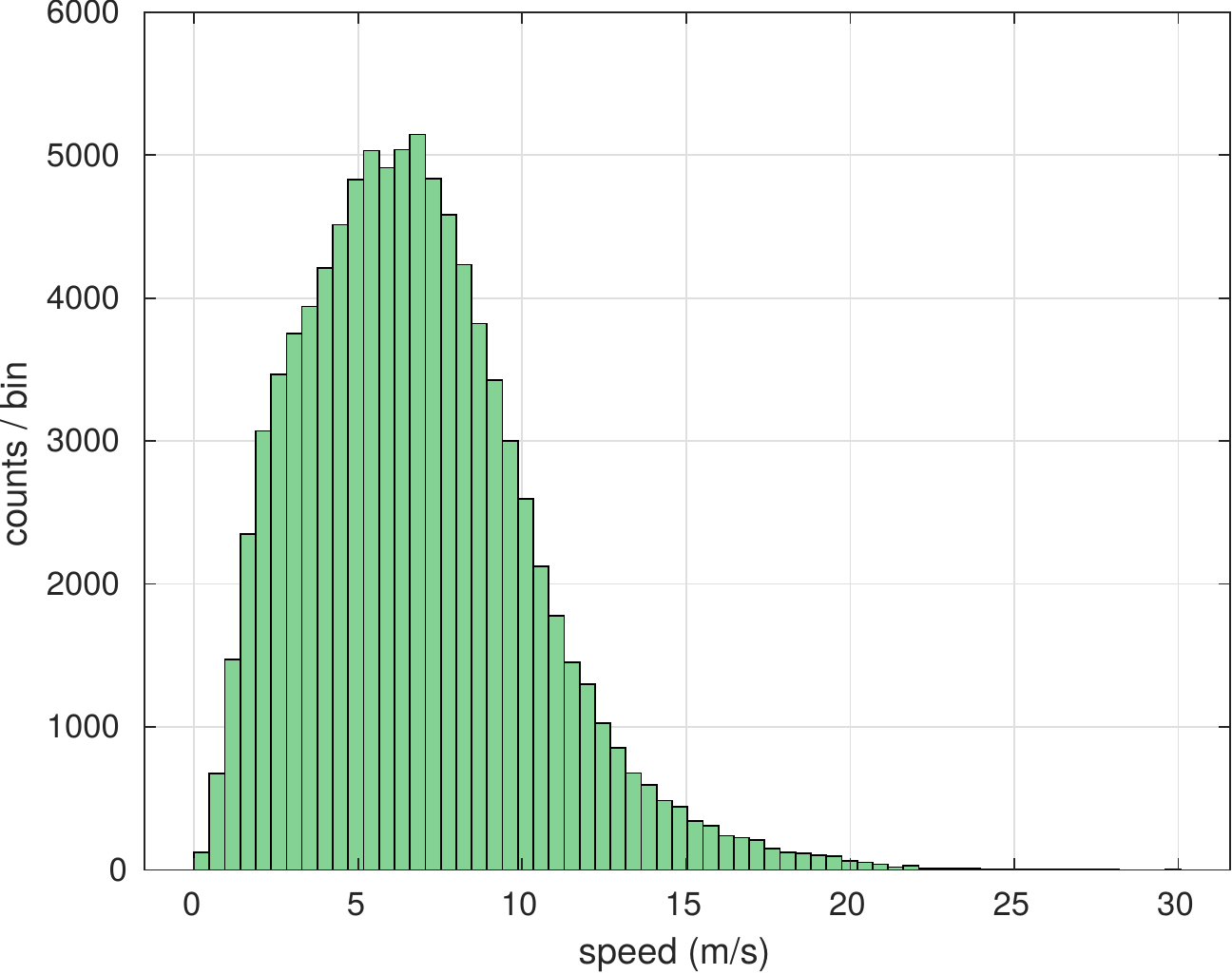}}
      \subfigure[]{
        \label{Pena_mod}
        \includegraphics[width=0.31\textwidth]{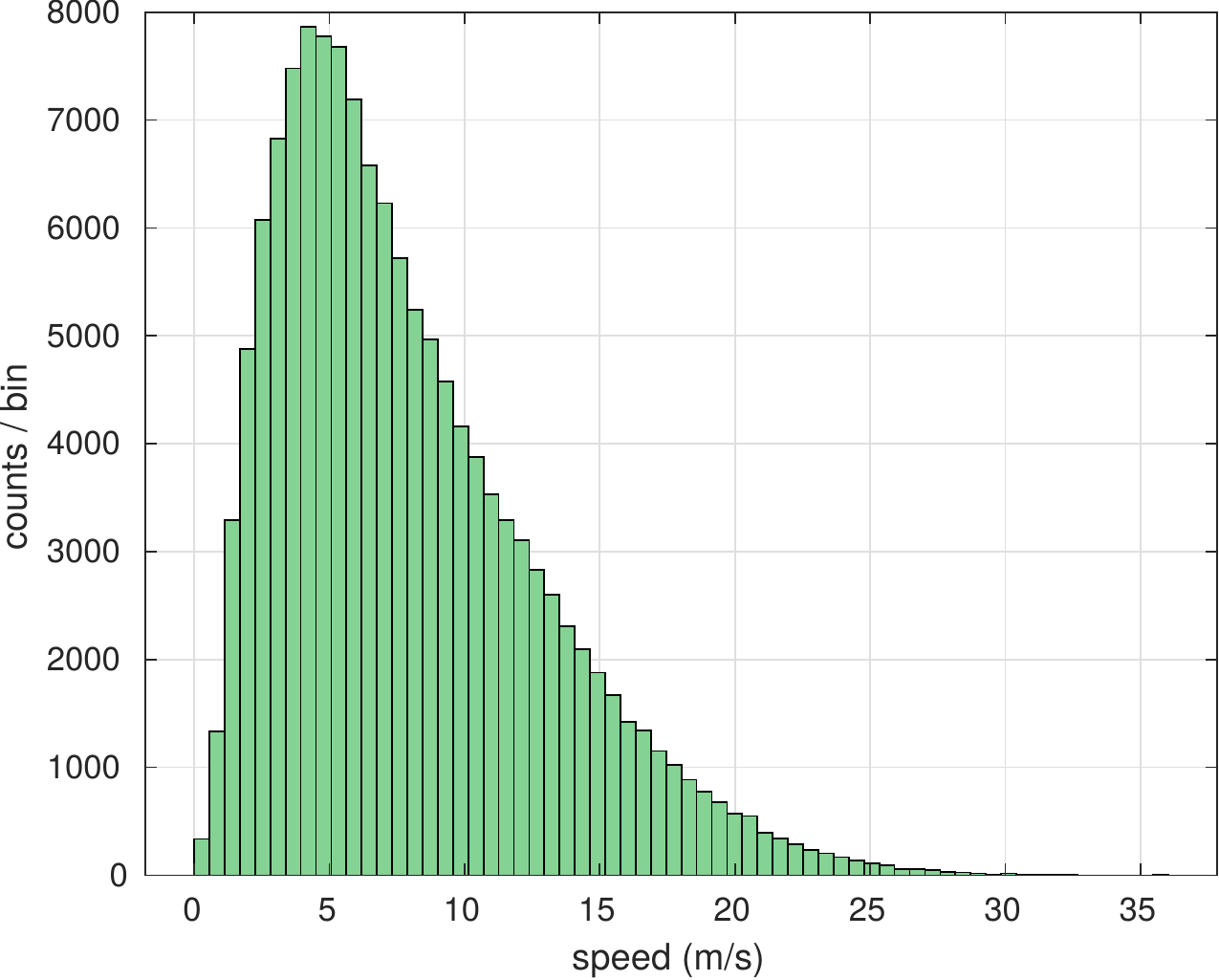}}
      \subfigure[]{
        \label{Jara_mod}
        \includegraphics[width=0.31\textwidth]{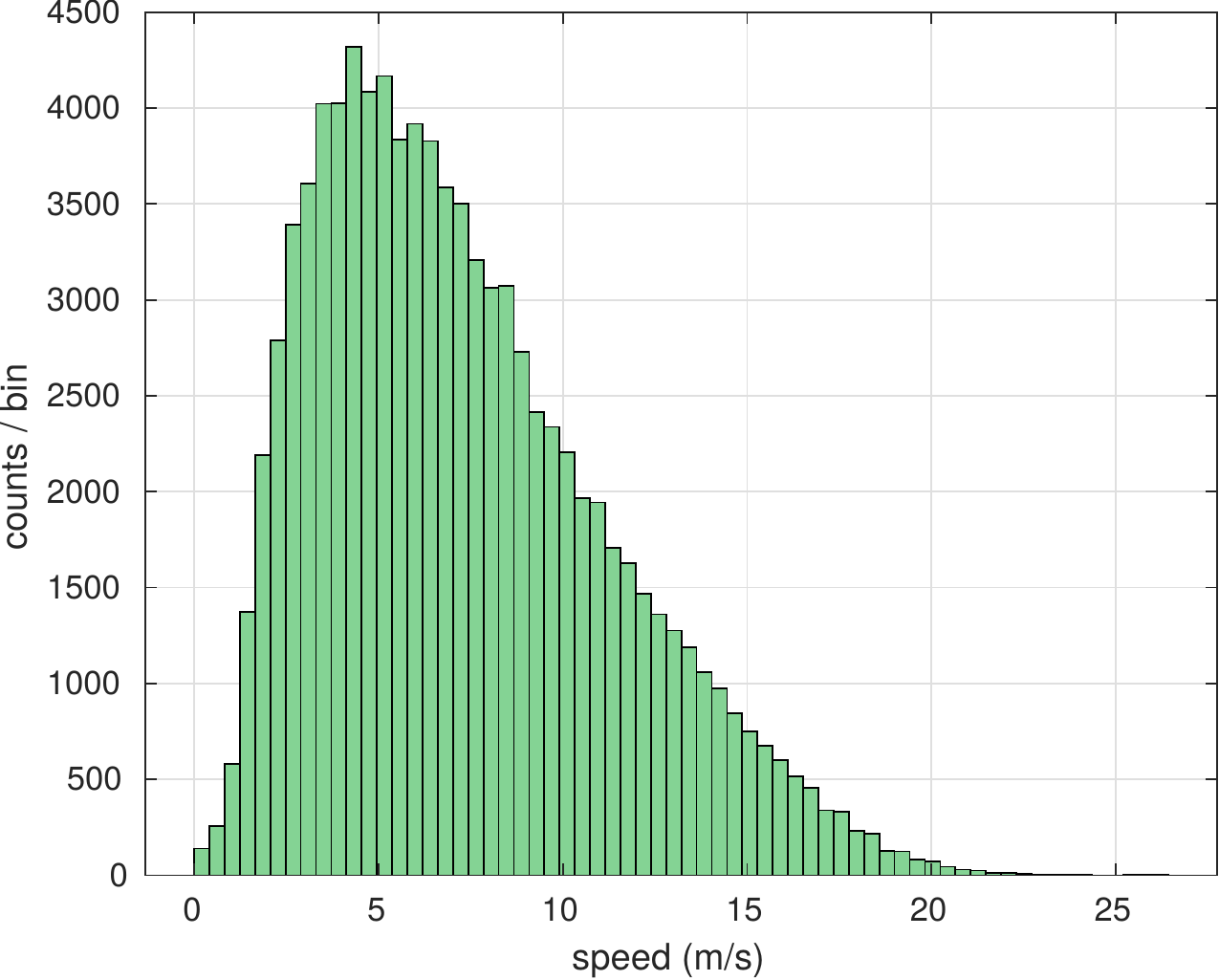}}
    \caption{Histograms of wind speed module for the three wind farms: A, B and C. Observe that the wind speed module can be modeled by means of a Weibull distribution.}
    \label{fig:histMod}
\end{figure}

Figure \ref{fig:histMod} shows the histograms of the wind speed module for the three wind farms considered. As can be seen, they can be approximated by Weibull distributions, with parameters given in Table \ref{tab:W_wind-param}. 
\begin{table}[!ht]
    \centering
    \caption{\label{tab:W_wind-param} Weibull parameters ($\lambda$ and $k$) estimation for the three wind farms considered.}

    \begin{tabular}{ccc}
        \toprule
        Wind Farm & scale ($\lambda$) & shape ($k$) \\ \midrule
         A & $7.6553$ & $2.0494$ \\ 
         B & $8.8942$ & $1.7132$ \\ 
         C & $8.2415$ & $1.9473$ \\ \bottomrule
    \end{tabular}
\end{table}

\subsubsection{Experimental results}

In this case study we tackle a problem of wind speed prediction with different prediction time-horizons (hourly, 4 hours and 8 hours) on the data described above. As we consider prediction time horizons lower than 8 hours, we will base the prediction only on previous wind speed data, without taken into account the atmospheric state or variables. For longer prediction time-horizons, the inclusion of exogenous variables to take into account the atmospheric state and dynamics is necessary. 

Tables \ref{wspred_A}, \ref{wspred_B} and \ref{wspred_C} show the results obtained in the wind speed prediction problem of wind farms A, B and C, respectively, for all the prediction time-horizons considered (hourly, 4 hours and 8 hours). These tables include a full comparison of the results by the different randomization-based techniques, and alternative ML regression such as SVR, ML or kNN. We have also included the results of the persistence operator ($x[n+1]=x[n]$), which in short-term prediction problems is usually considered as the baseline to be beaten. As can be seen, the prediction time-horizon of 1 hour is clearly dominated by persistence, where this operator is able to obtain $R^2$ of 0.832, 0.888 and 0.914, for wind farms A, B and C, respectively. In such a short-term prediction time horizon the ML randomized-based algorithms are able to obtain slightly better results than persistence, such as in wind farm A, where The EDeep RVFL is able to obtain an $R^2=0.835$. Similar small improvements are found for wind farms B and C with the randomization-based techniques over the persistence operator. In all cases, the classical ML approaches, kNN, SVR and MLP are not able to improve the results of the persistence in 1 hour prediction time-horizons.

In the experiments with 4 hours time-horizon, the difference between the ML algorithms and the persistence is more evident. As can be seen, the persistence is still better than the classical ML approaches, but the differences in prediction errors and $R^2$ are now smaller. In this case, the randomization-based approaches are able to beat persistence more clearly, such as in wind farm A, $R^2=0.464$ by the edRVFL, versus $R^2=0.395$ obtained by the persistence operator. In wind farms B and C, the results are similar, $R^2=0.565$ by the edRVFL, versus $R^2=0.551$ obtained by the persistence, and $R^2=0.705$ by the edRVFL, versus $R^2=0.676$ obtained by the persistence.

Finally, in the case of 8-hours prediction time-horizon the persistence effect in the prediction is even smaller, and the differences with respect to ML randomization-based algorithms are much more evident. In this case, classical ML approaches are able to obtain results similar to the persistence operator, and the randomization-based approaches clearly outperform persistence operator. Note that, in general, the performance of all algorithms is low, since we are here at the limit where the prediction needs atmospheric information to be significant. There are also important differences between wind farms. In the case of wind farm A, the performance of all the algorithms is very poor, which means that the atmospheric information plays an important role in the wind speed prediction a this location. The best algorithm is again the edRVFL, but with a $R^2=0.15$, which means that the prediction is not accurate at all in this case. In wind farm B, the prediction is better. The best approach in this case is the ESN, with $R^2=0.325$ versus $R^2=0.218$ by the persistence operator. Finally, in wind farm C, the best algorithm is the dESN, with $R^2=0.534$ versus $R^2=0.454$ by the persistence operator. Note that the performance of the ML approaches is still poor, but much better than in wind farm A. 
\begin{table}[]
\caption{\label{wspred_A} Wind speed prediction results for wind farm A for different forecasting horizons.}
\vspace{2mm}
\centering
\resizebox{0.9\columnwidth}{!}{
\begin{tabular}{lccccc}
\toprule
Model & MBE & MAE & RMSE & $R^2$ & TIME \\
\midrule
\multicolumn{6}{c}{Horizon: 1 }\\
\midrule
LV & 0.007 & 0.823 & 1.135 & 0.832 & -- \\
kNN & $0.188 \pm 0.0$ & $1.49 \pm 0.0$ & $1.875 \pm 0.0$ & $0.543 \pm 0.0$ & $0.011 \pm 0.0$ \\
SVR & $0.241 \pm 0.0$ & $0.968 \pm 0.0$ & $1.271 \pm 0.0$ & $0.79 \pm 0.0$ & $0.91 \pm 0.003$ \\
MLP & $0.225 \pm 0.073$ & $0.917 \pm 0.028$ & $1.224 \pm 0.027$ & $0.805 \pm 0.008$ & $5.194 \pm 1.085$ \\
RFR & $0.189 \pm 0.008$ & $0.877 \pm 0.005$ & $1.17 \pm 0.005$ & $0.822 \pm 0.002$ & $2.999 \pm 0.088$ \\
AdaBoost & $0.159 \pm 0.011$ & $0.897 \pm 0.007$ & $1.193 \pm 0.008$ & $0.815 \pm 0.002$ & $0.339 \pm 0.004$ \\
ELM & $0.199 \pm 0.082$ & $1.187 \pm 0.036$ & $1.524 \pm 0.038$ & $0.698 \pm 0.015$ & $2.341 \pm 0.018$ \\
ML-ELM & $0.371 \pm 0.099$ & $1.181 \pm 0.035$ & $1.518 \pm 0.038$ & $0.7 \pm 0.015$ & $0.206 \pm 0.029$ \\
RVFL & $0.096 \pm 0.013$ & $0.821 \pm 0.005$ & $1.13 \pm 0.004$ & $0.834 \pm 0.001$ & $0.119 \pm 0.013$ \\
multiRVFL & $0.108 \pm 0.014$ & $0.825 \pm 0.007$ & $1.131 \pm 0.006$ & $0.834 \pm 0.002$ & $0.171 \pm 0.002$ \\
dRVFL & $0.088 \pm 0.016$ & $0.825 \pm 0.007$ & $1.133 \pm 0.006$ & $0.833 \pm 0.002$ & $0.817 \pm 0.002$ \\
edRVFL & $0.097 \pm 0.015$ & $0.82 \pm 0.005$ & $1.127 \pm 0.004$ & $0.835 \pm 0.001$ & $0.192 \pm 0.012$ \\
ESN & $0.16 \pm 0.023$ & $0.83 \pm 0.008$ & $1.135 \pm 0.007$ & $0.833 \pm 0.002$ & $3.573 \pm 0.042$ \\
dESN & $0.163 \pm 0.034$ & $0.84 \pm 0.012$ & $1.145 \pm 0.012$ & $0.83 \pm 0.004$ & $6.869 \pm 0.337$ \\
\midrule
\multicolumn{6}{c}{Horizon: 4 }\\
\midrule
LV & -0.001 & 1.623 & 2.148 & 0.395 & -- \\
kNN & $0.327 \pm 0.0$ & $2.04 \pm 0.0$ & $2.514 \pm 0.0$ & $0.17 \pm 0.0$ & $0.011 \pm 0.0$ \\
SVR & $0.522 \pm 0.0$ & $1.763 \pm 0.0$ & $2.188 \pm 0.0$ & $0.372 \pm 0.0$ & $0.515 \pm 0.002$ \\
MLP & $0.472 \pm 0.086$ & $1.846 \pm 0.069$ & $2.29 \pm 0.076$ & $0.311 \pm 0.046$ & $10.2 \pm 0.029$ \\
RFR & $0.518 \pm 0.02$ & $1.727 \pm 0.01$ & $2.135 \pm 0.009$ & $0.402 \pm 0.005$ & $1.333 \pm 0.077$ \\
AdaBoost & $0.521 \pm 0.019$ & $1.742 \pm 0.012$ & $2.152 \pm 0.01$ & $0.392 \pm 0.006$ & $0.339 \pm 0.005$ \\
ELM & $0.39 \pm 0.071$ & $1.885 \pm 0.062$ & $2.357 \pm 0.078$ & $0.27 \pm 0.047$ & $2.368 \pm 0.006$ \\
ML-ELM & $0.558 \pm 0.108$ & $1.873 \pm 0.08$ & $2.31 \pm 0.09$ & $0.299 \pm 0.056$ & $0.105 \pm 0.031$ \\
RVFL & $0.301 \pm 0.059$ & $1.625 \pm 0.037$ & $2.045 \pm 0.035$ & $0.451 \pm 0.019$ & $0.164 \pm 0.006$ \\
multiRVFL & $0.354 \pm 0.053$ & $1.628 \pm 0.032$ & $2.049 \pm 0.034$ & $0.449 \pm 0.018$ & $0.195 \pm 0.009$ \\
dRVFL & $0.318 \pm 0.082$ & $1.653 \pm 0.039$ & $2.075 \pm 0.04$ & $0.435 \pm 0.022$ & $0.982 \pm 0.005$ \\
edRVFL & $0.334 \pm 0.024$ & $1.614 \pm 0.026$ & $2.021 \pm 0.026$ & $0.464 \pm 0.014$ & $0.843 \pm 0.003$ \\
ESN & $0.57 \pm 0.097$ & $1.713 \pm 0.051$ & $2.134 \pm 0.043$ & $0.402 \pm 0.024$ & $3.16 \pm 0.448$ \\
dESN & $0.571 \pm 0.098$ & $1.699 \pm 0.048$ & $2.118 \pm 0.041$ & $0.411 \pm 0.023$ & $3.187 \pm 0.424$ \\
\midrule
\multicolumn{6}{c}{Horizon: 8 }\\
\midrule
LV & -0.002 & 2.258 & 2.856 & -0.071 & -- \\
kNN & $0.45 \pm 0.0$ & $2.267 \pm 0.0$ & $2.814 \pm 0.0$ & $-0.041 \pm 0.0$ & $0.011 \pm 0.0$ \\
SVR & $0.572 \pm 0.0$ & $2.053 \pm 0.0$ & $2.55 \pm 0.0$ & $0.146 \pm 0.0$ & $0.704 \pm 0.001$ \\
MLP & $0.661 \pm 0.12$ & $2.19 \pm 0.077$ & $2.708 \pm 0.069$ & $0.036 \pm 0.05$ & $5.22 \pm 1.087$ \\
RFR & $0.717 \pm 0.037$ & $2.232 \pm 0.022$ & $2.685 \pm 0.023$ & $0.052 \pm 0.016$ & $0.59 \pm 0.005$ \\
AdaBoost & $0.707 \pm 0.02$ & $2.243 \pm 0.021$ & $2.692 \pm 0.021$ & $0.048 \pm 0.015$ & $0.351 \pm 0.013$ \\
ELM & $0.371 \pm 0.08$ & $2.203 \pm 0.093$ & $2.731 \pm 0.098$ & $0.019 \pm 0.07$ & $1.431 \pm 0.005$ \\
ML-ELM & $0.68 \pm 0.069$ & $2.205 \pm 0.07$ & $2.702 \pm 0.073$ & $0.04 \pm 0.051$ & $0.198 \pm 0.038$ \\
RVFL & $0.571 \pm 0.056$ & $2.135 \pm 0.049$ & $2.583 \pm 0.058$ & $0.123 \pm 0.039$ & $0.15 \pm 0.006$ \\
multiRVFL & $0.526 \pm 0.076$ & $2.1 \pm 0.043$ & $2.558 \pm 0.038$ & $0.14 \pm 0.025$ & $0.422 \pm 0.003$ \\
dRVFL & $0.439 \pm 0.068$ & $2.132 \pm 0.068$ & $2.628 \pm 0.073$ & $0.092 \pm 0.05$ & $0.643 \pm 0.001$ \\
edRVFL & $0.447 \pm 0.061$ & $2.074 \pm 0.05$ & $2.542 \pm 0.051$ & $0.15 \pm 0.034$ & $0.571 \pm 0.002$ \\
ESN & $0.691 \pm 0.051$ & $2.193 \pm 0.039$ & $2.65 \pm 0.043$ & $0.077 \pm 0.03$ & $4.194 \pm 0.032$ \\
dESN & $0.764 \pm 0.181$ & $2.236 \pm 0.071$ & $2.689 \pm 0.059$ & $0.05 \pm 0.042$ & $6.611 \pm 0.512$ \\
\bottomrule
\end{tabular}}
\end{table}

\begin{table}[]
\caption{\label{wspred_B} Wind speed prediction results for wind farm B for different forecasting horizons.}
\vspace{2mm}
\centering
\resizebox{0.9\columnwidth}{!}{
\begin{tabular}{lccccc}
\toprule
Model & MBE & MAE & RMSE & $R^2$  & TIME \\
\midrule
\multicolumn{6}{c}{Horizon: 1 }\\
\midrule
LV & 0.026 & 1.345 & 1.989 & 0.888 & -- \\
kNN & $-0.448 \pm 0.0$ & $2.766 \pm 0.0$ & $3.646 \pm 0.0$ & $0.625 \pm 0.0$ & $0.017 \pm 0.0$ \\
SVR & $-0.387 \pm 0.0$ & $1.923 \pm 0.0$ & $2.801 \pm 0.0$ & $0.778 \pm 0.0$ & $2.84 \pm 0.002$ \\
MLP & $0.119 \pm 0.086$ & $1.431 \pm 0.009$ & $2.054 \pm 0.014$ & $0.881 \pm 0.002$ & $5.797 \pm 1.09$ \\
RFR & $-0.021 \pm 0.005$ & $1.374 \pm 0.005$ & $2.041 \pm 0.008$ & $0.882 \pm 0.001$ & $10.936 \pm 0.896$ \\
AdaBoost & $-0.048 \pm 0.022$ & $1.55 \pm 0.017$ & $2.266 \pm 0.016$ & $0.855 \pm 0.002$ & $0.503 \pm 0.001$ \\
ELM & $-0.156 \pm 0.111$ & $2.032 \pm 0.043$ & $2.717 \pm 0.062$ & $0.791 \pm 0.009$ & $3.237 \pm 0.007$ \\
ML-ELM & $-0.33 \pm 0.047$ & $2.111 \pm 0.028$ & $2.879 \pm 0.049$ & $0.766 \pm 0.008$ & $0.514 \pm 0.084$ \\
RVFL & $-0.038 \pm 0.013$ & $1.353 \pm 0.002$ & $1.998 \pm 0.003$ & $0.887 \pm 0.0$ & $0.149 \pm 0.019$ \\
multiRVFL & $-0.022 \pm 0.014$ & $1.355 \pm 0.004$ & $2.0 \pm 0.006$ & $0.887 \pm 0.001$ & $0.933 \pm 0.003$ \\
dRVFL & $-0.02 \pm 0.027$ & $1.378 \pm 0.008$ & $2.011 \pm 0.013$ & $0.886 \pm 0.001$ & $1.238 \pm 0.012$ \\
edRVFL & $-0.038 \pm 0.011$ & $1.365 \pm 0.004$ & $2.004 \pm 0.008$ & $0.887 \pm 0.001$ & $0.897 \pm 0.002$ \\
ESN & $-0.02 \pm 0.016$ & $1.339 \pm 0.011$ & $2.002 \pm 0.021$ & $0.887 \pm 0.002$ & $4.338 \pm 1.486$ \\
dESN & $-0.027 \pm 0.021$ & $1.337 \pm 0.01$ & $1.995 \pm 0.012$ & $0.888 \pm 0.001$ & $4.769 \pm 0.513$ \\
\midrule
\multicolumn{6}{c}{Horizon: 4 }\\
\midrule
LV & -0.12 & 2.917 & 3.986 & 0.551 & -- \\
kNN & $-0.489 \pm 0.0$ & $3.766 \pm 0.0$ & $4.82 \pm 0.0$ & $0.344 \pm 0.0$ & $0.017 \pm 0.0$ \\
SVR & $-0.558 \pm 0.0$ & $3.373 \pm 0.0$ & $4.369 \pm 0.0$ & $0.461 \pm 0.0$ & $1.555 \pm 0.002$ \\
MLP & $-0.101 \pm 0.149$ & $3.077 \pm 0.026$ & $4.109 \pm 0.04$ & $0.523 \pm 0.009$ & $5.816 \pm 1.097$ \\
RFR & $-0.318 \pm 0.039$ & $2.958 \pm 0.026$ & $3.958 \pm 0.027$ & $0.557 \pm 0.006$ & $0.567 \pm 0.008$ \\
AdaBoost & $-0.333 \pm 0.076$ & $2.963 \pm 0.041$ & $3.966 \pm 0.045$ & $0.556 \pm 0.01$ & $0.502 \pm 0.001$ \\
ELM & $-0.541 \pm 0.135$ & $3.339 \pm 0.082$ & $4.394 \pm 0.091$ & $0.454 \pm 0.023$ & $1.965 \pm 0.007$ \\
ML-ELM & $-0.43 \pm 0.092$ & $3.297 \pm 0.05$ & $4.359 \pm 0.07$ & $0.463 \pm 0.017$ & $0.505 \pm 0.054$ \\
RVFL & $-0.381 \pm 0.041$ & $2.899 \pm 0.016$ & $3.923 \pm 0.013$ & $0.565 \pm 0.003$ & $0.16 \pm 0.014$ \\
multiRVFL & $-0.336 \pm 0.07$ & $2.92 \pm 0.035$ & $3.943 \pm 0.035$ & $0.561 \pm 0.008$ & $1.135 \pm 0.008$ \\
dRVFL & $-0.355 \pm 0.064$ & $2.922 \pm 0.026$ & $3.944 \pm 0.029$ & $0.56 \pm 0.007$ & $0.584 \pm 0.002$ \\
edRVFL & $-0.397 \pm 0.046$ & $2.898 \pm 0.023$ & $3.922 \pm 0.025$ & $0.565 \pm 0.005$ & $0.29 \pm 0.002$ \\
ESN & $-0.348 \pm 0.15$ & $2.937 \pm 0.11$ & $3.954 \pm 0.146$ & $0.558 \pm 0.033$ & $6.146 \pm 0.055$ \\
dESN & $-0.331 \pm 0.055$ & $2.87 \pm 0.007$ & $3.868 \pm 0.01$ & $0.577 \pm 0.002$ & $4.38 \pm 0.499$ \\
\midrule
\multicolumn{6}{c}{Horizon: 8 }\\
\midrule
LV & -0.224 & 3.926 & 5.26 & 0.218 & -- \\
kNN & $-0.618 \pm 0.0$ & $4.451 \pm 0.0$ & $5.634 \pm 0.0$ & $0.103 \pm 0.0$ & $0.017 \pm 0.0$ \\
SVR & $-1.135 \pm 0.0$ & $4.086 \pm 0.0$ & $5.413 \pm 0.0$ & $0.172 \pm 0.0$ & $1.661 \pm 0.002$ \\
MLP & $-0.479 \pm 0.126$ & $4.054 \pm 0.085$ & $5.376 \pm 0.135$ & $0.183 \pm 0.041$ & $12.554 \pm 0.399$ \\
RFR & $-0.695 \pm 0.028$ & $3.812 \pm 0.011$ & $5.006 \pm 0.016$ & $0.292 \pm 0.004$ & $0.566 \pm 0.01$ \\
AdaBoost & $-0.644 \pm 0.037$ & $3.816 \pm 0.014$ & $5.005 \pm 0.018$ & $0.292 \pm 0.005$ & $0.503 \pm 0.001$ \\
ELM & $-0.73 \pm 0.124$ & $4.044 \pm 0.057$ & $5.31 \pm 0.059$ & $0.203 \pm 0.018$ & $1.974 \pm 0.004$ \\
ML-ELM & $-0.817 \pm 0.074$ & $4.214 \pm 0.111$ & $5.43 \pm 0.124$ & $0.167 \pm 0.037$ & $0.412 \pm 0.057$ \\
RVFL & $-0.779 \pm 0.089$ & $3.819 \pm 0.025$ & $5.044 \pm 0.03$ & $0.281 \pm 0.009$ & $0.196 \pm 0.003$ \\
multiRVFL & $-0.682 \pm 0.097$ & $3.804 \pm 0.035$ & $5.039 \pm 0.05$ & $0.283 \pm 0.014$ & $0.25 \pm 0.002$ \\
dRVFL & $-0.68 \pm 0.064$ & $3.799 \pm 0.033$ & $5.023 \pm 0.04$ & $0.287 \pm 0.011$ & $0.343 \pm 0.007$ \\
edRVFL & $-0.746 \pm 0.059$ & $3.791 \pm 0.021$ & $5.02 \pm 0.028$ & $0.288 \pm 0.008$ & $0.546 \pm 0.002$ \\
ESN & $-0.548 \pm 0.075$ & $3.727 \pm 0.037$ & $4.886 \pm 0.05$ & $0.325 \pm 0.014$ & $5.705 \pm 0.041$ \\
dESN & $-0.567 \pm 0.065$ & $3.732 \pm 0.033$ & $4.911 \pm 0.044$ & $0.319 \pm 0.012$ & $5.711 \pm 0.03$ \\
\bottomrule
\end{tabular}}
\end{table}

\begin{table}[]
\caption{\label{wspred_C}Wind speed prediction results for wind farm C for different forecasting horizons.}
\vspace{2mm}
\centering
\resizebox{0.9\columnwidth}{!}{
\begin{tabular}{lccccc}
\toprule
Model & MBE & MAE & RMSE & $R^2$  & TIME \\
\midrule
\multicolumn{6}{c}{Horizon: 1 }\\
\midrule
LV & 0.096 & 0.875 & 1.241 & 0.914 & -- \\
kNN & $-0.344 \pm 0.0$ & $1.745 \pm 0.0$ & $2.208 \pm 0.0$ & $0.729 \pm 0.0$ & $0.011 \pm 0.0$ \\
SVR & $-0.128 \pm 0.0$ & $1.144 \pm 0.0$ & $1.555 \pm 0.0$ & $0.865 \pm 0.0$ & $1.321 \pm 0.021$ \\
MLP & $-0.053 \pm 0.071$ & $0.931 \pm 0.019$ & $1.257 \pm 0.026$ & $0.912 \pm 0.004$ & $5.151 \pm 1.074$ \\
RFR & $-0.097 \pm 0.008$ & $0.862 \pm 0.004$ & $1.143 \pm 0.004$ & $0.927 \pm 0.001$ & $5.428 \pm 0.115$ \\
AdaBoost & $-0.097 \pm 0.014$ & $0.981 \pm 0.01$ & $1.287 \pm 0.009$ & $0.908 \pm 0.001$ & $0.855 \pm 0.001$ \\
ELM & $-0.123 \pm 0.068$ & $1.345 \pm 0.058$ & $1.733 \pm 0.078$ & $0.833 \pm 0.015$ & $2.381 \pm 0.003$ \\
ML-ELM & $-0.287 \pm 0.054$ & $1.363 \pm 0.057$ & $1.76 \pm 0.076$ & $0.827 \pm 0.015$ & $0.402 \pm 0.059$ \\
RVFL & $-0.038 \pm 0.02$ & $0.879 \pm 0.006$ & $1.227 \pm 0.004$ & $0.916 \pm 0.001$ & $0.156 \pm 0.029$ \\
multiRVFL & $-0.035 \pm 0.013$ & $0.886 \pm 0.008$ & $1.231 \pm 0.007$ & $0.916 \pm 0.001$ & $0.252 \pm 0.0$ \\
dRVFL & $-0.048 \pm 0.02$ & $0.896 \pm 0.012$ & $1.242 \pm 0.012$ & $0.914 \pm 0.002$ & $0.576 \pm 0.003$ \\
edRVFL & $-0.016 \pm 0.015$ & $0.885 \pm 0.007$ & $1.234 \pm 0.006$ & $0.915 \pm 0.001$ & $0.239 \pm 0.005$ \\
ESN & $-0.053 \pm 0.014$ & $0.883 \pm 0.008$ & $1.214 \pm 0.009$ & $0.918 \pm 0.001$ & $4.151 \pm 0.024$ \\
dESN & $-0.062 \pm 0.023$ & $0.89 \pm 0.007$ & $1.231 \pm 0.007$ & $0.916 \pm 0.001$ & $3.895 \pm 0.023$ \\
\midrule
\multicolumn{6}{c}{Horizon: 4 }\\
\midrule
LV & 0.028 & 1.869 & 2.38 & 0.676 & -- \\
kNN & $-0.36 \pm 0.0$ & $2.261 \pm 0.0$ & $2.831 \pm 0.0$ & $0.541 \pm 0.0$ & $0.011 \pm 0.0$ \\
SVR & $-0.417 \pm 0.0$ & $2.009 \pm 0.0$ & $2.543 \pm 0.0$ & $0.63 \pm 0.0$ & $0.585 \pm 0.003$ \\
MLP & $-0.235 \pm 0.075$ & $1.982 \pm 0.082$ & $2.506 \pm 0.091$ & $0.64 \pm 0.026$ & $9.904 \pm 0.347$ \\
RFR & $-0.25 \pm 0.021$ & $1.851 \pm 0.009$ & $2.319 \pm 0.009$ & $0.692 \pm 0.002$ & $0.417 \pm 0.007$ \\
AdaBoost & $-0.225 \pm 0.013$ & $1.861 \pm 0.014$ & $2.333 \pm 0.015$ & $0.688 \pm 0.004$ & $0.291 \pm 0.004$ \\
ELM & $-0.343 \pm 0.113$ & $2.225 \pm 0.159$ & $2.768 \pm 0.204$ & $0.559 \pm 0.065$ & $0.146 \pm 0.02$ \\
ML-ELM & $-0.425 \pm 0.123$ & $2.228 \pm 0.109$ & $2.79 \pm 0.145$ & $0.553 \pm 0.047$ & $0.105 \pm 0.028$ \\
RVFL & $-0.175 \pm 0.045$ & $1.807 \pm 0.021$ & $2.267 \pm 0.023$ & $0.706 \pm 0.006$ & $0.111 \pm 0.013$ \\
multiRVFL & $-0.166 \pm 0.054$ & $1.87 \pm 0.057$ & $2.358 \pm 0.066$ & $0.682 \pm 0.018$ & $0.447 \pm 0.013$ \\
dRVFL & $-0.175 \pm 0.074$ & $1.82 \pm 0.036$ & $2.287 \pm 0.038$ & $0.7 \pm 0.01$ & $0.228 \pm 0.004$ \\
edRVFL & $-0.147 \pm 0.039$ & $1.808 \pm 0.02$ & $2.268 \pm 0.02$ & $0.705 \pm 0.005$ & $0.187 \pm 0.013$ \\
ESN & $-0.255 \pm 0.043$ & $1.83 \pm 0.029$ & $2.294 \pm 0.035$ & $0.699 \pm 0.009$ & $5.147 \pm 0.032$ \\
dESN & $-0.226 \pm 0.021$ & $1.827 \pm 0.022$ & $2.282 \pm 0.021$ & $0.702 \pm 0.006$ & $3.894 \pm 0.026$ \\
\midrule
\multicolumn{6}{c}{Horizon: 8 }\\
\midrule
LV & -0.006 & 2.471 & 3.096 & 0.454 & -- \\
kNN & $-0.448 \pm 0.0$ & $2.585 \pm 0.0$ & $3.252 \pm 0.0$ & $0.398 \pm 0.0$ & $0.012 \pm 0.001$ \\
SVR & $-0.591 \pm 0.0$ & $2.474 \pm 0.0$ & $3.098 \pm 0.0$ & $0.453 \pm 0.0$ & $0.749 \pm 0.001$ \\
MLP & $-0.201 \pm 0.16$ & $2.49 \pm 0.071$ & $3.126 \pm 0.09$ & $0.443 \pm 0.032$ & $5.214 \pm 1.088$ \\
RFR & $-0.316 \pm 0.039$ & $2.366 \pm 0.028$ & $2.915 \pm 0.032$ & $0.516 \pm 0.011$ & $0.413 \pm 0.008$ \\
AdaBoost & $-0.236 \pm 0.031$ & $2.384 \pm 0.021$ & $2.939 \pm 0.021$ & $0.508 \pm 0.007$ & $0.288 \pm 0.001$ \\
ELM & $-0.225 \pm 0.127$ & $2.511 \pm 0.127$ & $3.202 \pm 0.176$ & $0.414 \pm 0.063$ & $1.37 \pm 0.007$ \\
ML-ELM & $-0.523 \pm 0.074$ & $2.648 \pm 0.08$ & $3.287 \pm 0.108$ & $0.384 \pm 0.04$ & $0.062 \pm 0.0$ \\
RVFL & $-0.25 \pm 0.053$ & $2.331 \pm 0.046$ & $2.873 \pm 0.05$ & $0.53 \pm 0.017$ & $0.142 \pm 0.004$ \\
multiRVFL & $-0.256 \pm 0.081$ & $2.366 \pm 0.051$ & $2.922 \pm 0.062$ & $0.514 \pm 0.02$ & $0.598 \pm 0.002$ \\
dRVFL & $-0.339 \pm 0.097$ & $2.361 \pm 0.061$ & $2.91 \pm 0.07$ & $0.518 \pm 0.023$ & $0.66 \pm 0.003$ \\
edRVFL & $-0.239 \pm 0.051$ & $2.34 \pm 0.036$ & $2.889 \pm 0.04$ & $0.525 \pm 0.013$ & $0.512 \pm 0.002$ \\
ESN & $-0.276 \pm 0.091$ & $2.32 \pm 0.046$ & $2.883 \pm 0.055$ & $0.527 \pm 0.018$ & $8.613 \pm 0.091$ \\
dESN & $-0.343 \pm 0.06$ & $2.312 \pm 0.036$ & $2.862 \pm 0.035$ & $0.534 \pm 0.011$ & $5.13 \pm 0.04$ \\
\bottomrule
\end{tabular}}
\end{table}

Table \ref{tab:Wilcoxon_wind} shows Wilcoxon signed rank p-values for wind farms A, B and C. As in the case of the solar prediction, values shaded in gray correspond to those cases where statistical significance cannot be declared at $\alpha=0.05$. As can be seen, in wind farm A, the traditional ML algorithms works statistically similar, whereas the randomization-based approaches seem to work different (better) in the majority of cases. In wind farm B, the randomization-based techniques work statistically better than the alternative approaches, and finally, in the wind farm C, the performance of the edRVFL, ESN and dESN is statistically similar, and better than the rest of approaches. 
\begin{table}[ht]
\caption{\label{tab:Wilcoxon_wind} Wilcoxon signed rank p-values for wind farms A, B and C. Cells shaded in gray correspond to those cases where statistical significance cannot be declared at $\alpha=0.05$.}
\centering
\resizebox{\columnwidth}{!}{
\begin{tabular}{ccccccccccccc}
\toprule
p-value &
SVR & MLP & RFR & AdaBoost & ELM & ML-ELM & RVFL & multiRVFL & dRVFL & edRVFL & ESN\\
\midrule
\multicolumn{13}{c}{Wind farm A}\\
\midrule
MLP & 1.691e-06 & --- & --- & --- & --- & --- & --- & --- & --- & --- & --- & \\
RFR & 1.691e-06 & \cellcolor[gray]{0.8}2.531e-01 & --- & --- & --- & --- & --- & --- & --- & --- & --- & \\
AdaBoost & 1.691e-06 & \cellcolor[gray]{0.8}5.995e-01 & \cellcolor[gray]{0.8}2.054e-01 & --- & --- & --- & --- & --- & --- & --- & --- & \\
ELM & 1.691e-06 & \cellcolor[gray]{0.8}3.383e-01 & \cellcolor[gray]{0.8}5.957e-02 & \cellcolor[gray]{0.8}1.016e-01 & --- & --- & --- & --- & --- & --- & --- & \\
ML-ELM & 3.106e-06 & \cellcolor[gray]{0.8}8.288e-01 & \cellcolor[gray]{0.8}2.531e-01 & \cellcolor[gray]{0.8}2.531e-01 & \cellcolor[gray]{0.8}7.341e-01 & --- & --- & --- & --- & --- & --- & \\
RVFL & 1.036e-02 & 1.762e-05 & 1.691e-06 & 1.691e-06 & 3.106e-06 & 1.691e-06 & --- & --- & --- & --- & --- & \\
multiRVFL & \cellcolor[gray]{0.8}4.043e-01 & 1.691e-06 & 1.691e-06 & 1.691e-06 & 1.691e-06 & 3.106e-06 & 3.852e-02 & --- & --- & --- & --- & \\
dRVFL & 1.861e-04 & 3.020e-04 & 1.467e-03 & 6.771e-05 & 1.454e-04 & 1.183e-03 & 2.832e-02 & 3.826e-04 & --- & --- & --- & \\
edRVFL & \cellcolor[gray]{0.8}9.262e-01 & 1.691e-06 & 1.691e-06 & 1.691e-06 & 1.691e-06 & 3.106e-06 & 2.054e-02 & \cellcolor[gray]{0.8}2.054e-01 & 1.861e-04 & --- & --- & \\
ESN & 3.106e-06 & 5.982e-03 & 3.020e-04 & 2.375e-04 & 8.766e-05 & 7.209e-03 & 3.106e-06 & 1.691e-06 & \cellcolor[gray]{0.8}8.938e-02 & 3.106e-06 & --- & \\
dESN & 1.691e-06 & \cellcolor[gray]{0.8}2.797e-01 & \cellcolor[gray]{0.8}1.016e-01 & 4.469e-02 & \cellcolor[gray]{0.8}7.810e-01 & \cellcolor[gray]{0.8}7.341e-01 & 1.691e-06 & 1.691e-06 & 3.826e-04 & 3.106e-06 & 1.861e-04 & \\
\midrule
\multicolumn{13}{c}{Wind farm B}\\
\midrule
MLP & \cellcolor[gray]{0.8}8.938e-02 & --- & --- & --- & --- & --- & --- & --- & --- & --- & --- & \\
RFR & 1.691e-06 & 1.691e-06 & --- & --- & --- & --- & --- & --- & --- & --- & --- & \\
AdaBoost & 1.691e-06 & 1.691e-06 & \cellcolor[gray]{0.8}3.704e-01 & --- & --- & --- & --- & --- & --- & --- & --- & \\
ELM & 1.691e-06 & 1.469e-02 & 1.691e-06 & 1.691e-06 & --- & --- & --- & --- & --- & --- & --- & \\
ML-ELM & \cellcolor[gray]{0.8}1.152e-01 & 1.236e-02 & 1.691e-06 & 1.691e-06 & 6.075e-04 & --- & --- & --- & --- & --- & --- & \\
RVFL & 1.691e-06 & 1.691e-06 & 3.053e-05 & 3.106e-06 & 1.691e-06 & 1.691e-06 & --- & --- & --- & --- & --- & \\
multiRVFL & 1.691e-06 & 3.106e-06 & 1.469e-02 & 2.231e-03 & 1.691e-06 & 1.691e-06 & \cellcolor[gray]{0.8}6.880e-01 & --- & --- & --- & --- & \\
dRVFL & 1.691e-06 & 1.691e-06 & \cellcolor[gray]{0.8}5.168e-02 & 1.740e-02 & 1.691e-06 & 1.691e-06 & 1.812e-03 & \cellcolor[gray]{0.8}1.842e-01 & --- & --- & --- & \\
edRVFL & 1.691e-06 & 1.691e-06 & 8.659e-03 & 1.812e-03 & 1.691e-06 & 1.691e-06 & 7.613e-04 & 3.308e-02 & \cellcolor[gray]{0.8}7.810e-01 & --- & --- & \\
ESN & 1.691e-06 & 1.691e-06 & 1.691e-06 & 1.691e-06 & 1.691e-06 & 1.691e-06 & 1.691e-06 & 1.691e-06 & 1.691e-06 & 3.106e-06 & --- & \\
dESN & 1.691e-06 & 1.691e-06 & 8.766e-05 & 8.766e-05 & 1.691e-06 & 1.691e-06 & 1.762e-05 & 3.106e-06 & 3.826e-04 & 6.771e-05 & 7.613e-04 & \\
\midrule
\multicolumn{13}{c}{Wind farm C}\\
\midrule
MLP & 3.852e-02 & --- & --- & --- & --- & --- & --- & --- & --- & --- & --- & \\
RFR & 1.691e-06 & 1.691e-06 & --- & --- & --- & --- & --- & --- & --- & --- & --- & \\
AdaBoost & 1.691e-06 & 1.691e-06 & 2.231e-03 & --- & --- & --- & --- & --- & --- & --- & --- & \\
ELM & 7.209e-03 & 3.852e-02 & 7.522e-06 & 7.522e-06 & --- & --- & --- & --- & --- & --- & --- & \\
ML-ELM & 3.106e-06 & 3.053e-05 & 1.691e-06 & 1.691e-06 & \cellcolor[gray]{0.8}4.775e-01 & --- & --- & --- & --- & --- & --- & \\
RVFL & 1.691e-06 & 1.691e-06 & 4.075e-03 & 7.522e-06 & 3.106e-06 & 1.691e-06 & --- & --- & --- & --- & --- & \\
multiRVFL & 1.691e-06 & 3.106e-06 & \cellcolor[gray]{0.8}5.573e-01 & \cellcolor[gray]{0.8}4.775e-01 & 7.522e-06 & 1.691e-06 & 2.416e-02 & --- & --- & --- & --- & \\
dRVFL & 1.691e-06 & 1.691e-06 & \cellcolor[gray]{0.8}7.810e-01 & \cellcolor[gray]{0.8}8.938e-02 & 3.106e-06 & 1.691e-06 & \cellcolor[gray]{0.8}1.016e-01 & \cellcolor[gray]{0.8}3.081e-01 & --- & --- & --- & \\
edRVFL & 1.691e-06 & 1.691e-06 & 4.075e-03 & 7.522e-06 & 3.106e-06 & 1.691e-06 & \cellcolor[gray]{0.8}5.166e-01 & 1.036e-02 & \cellcolor[gray]{0.8}4.043e-01 & --- & --- & \\
ESN & 1.691e-06 & 1.691e-06 & 3.308e-02 & 9.507e-04 & 3.106e-06 & 1.691e-06 & \cellcolor[gray]{0.8}5.573e-01 & 4.075e-03 & \cellcolor[gray]{0.8}1.466e-01 & \cellcolor[gray]{0.8}3.704e-01 & --- & \\
dESN & 1.691e-06 & 3.106e-06 & \cellcolor[gray]{0.8}8.288e-01 & \cellcolor[gray]{0.8}1.646e-01 & 7.522e-06 & 1.691e-06 & \cellcolor[gray]{0.8}1.016e-01 & \cellcolor[gray]{0.8}3.383e-01 & \cellcolor[gray]{0.8}6.880e-01 & \cellcolor[gray]{0.8}6.842e-02 & \cellcolor[gray]{0.8}1.646e-01 & \\
\bottomrule
\end{tabular}}

\end{table}

\begin{figure}[!ht]
	\vspace{6pt}
	\begin{center}
		\includegraphics[width=\columnwidth]{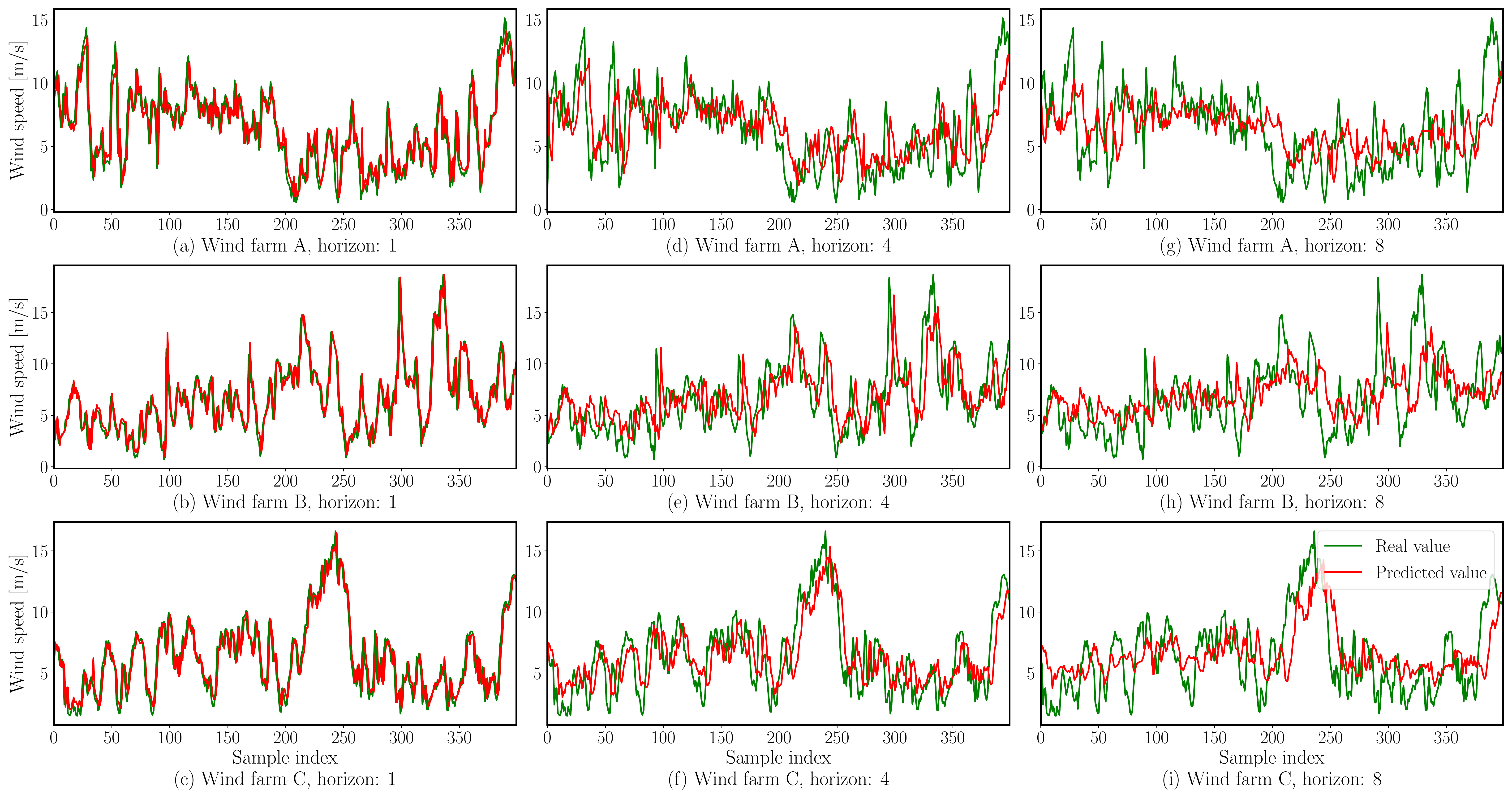}
	\end{center}
	\caption[]{ \label{fig:plots_wind} Predicted versus real time series for different prediction horizons and wind farm A and RVFL (a, b and c), wind farm B and dRVFL (d, e and f); and wind farm C and dESN (g, h and i).}
\end{figure}

Figure \ref{fig:plots_wind} shows the performance of the best randomization-based algorithms in this wind speed prediction problem, in the different prediction time-horizons considered. Figures (a, b and c) show the performance of a RVFL in wind farm A, Figures (d, e and f) show the performance of dRVFL in wind farm B, and Figures (g, h and i) show the performance of dESN in data from wind farm C. In all cases, the performance of the randomization-based algorithms for the 1 hour time horizon is excellent, and it is getting worse when the prediction time-horizon grows. The prediction for the 8-hours time-horizon is clearly poor in all cases without information from the atmospheric state. 

\subsection{Dammed water level prediction for hydro-power}

The last case study presented in this paper deals with a problem of water level prediction in a dammed water reservoir, with hydro-power application. We first describe the dataset and the input variables for the problem, and then we show the performance of the randomization-based algorithms in this problem, compared to classical ML approaches such as SVR, MLP and GPR.

\subsubsection{Dataset Description}
\begin{figure}[!ht]
	\vspace{6pt}
	\begin{center}
		\includegraphics[draft=false, angle=0,width=8cm]{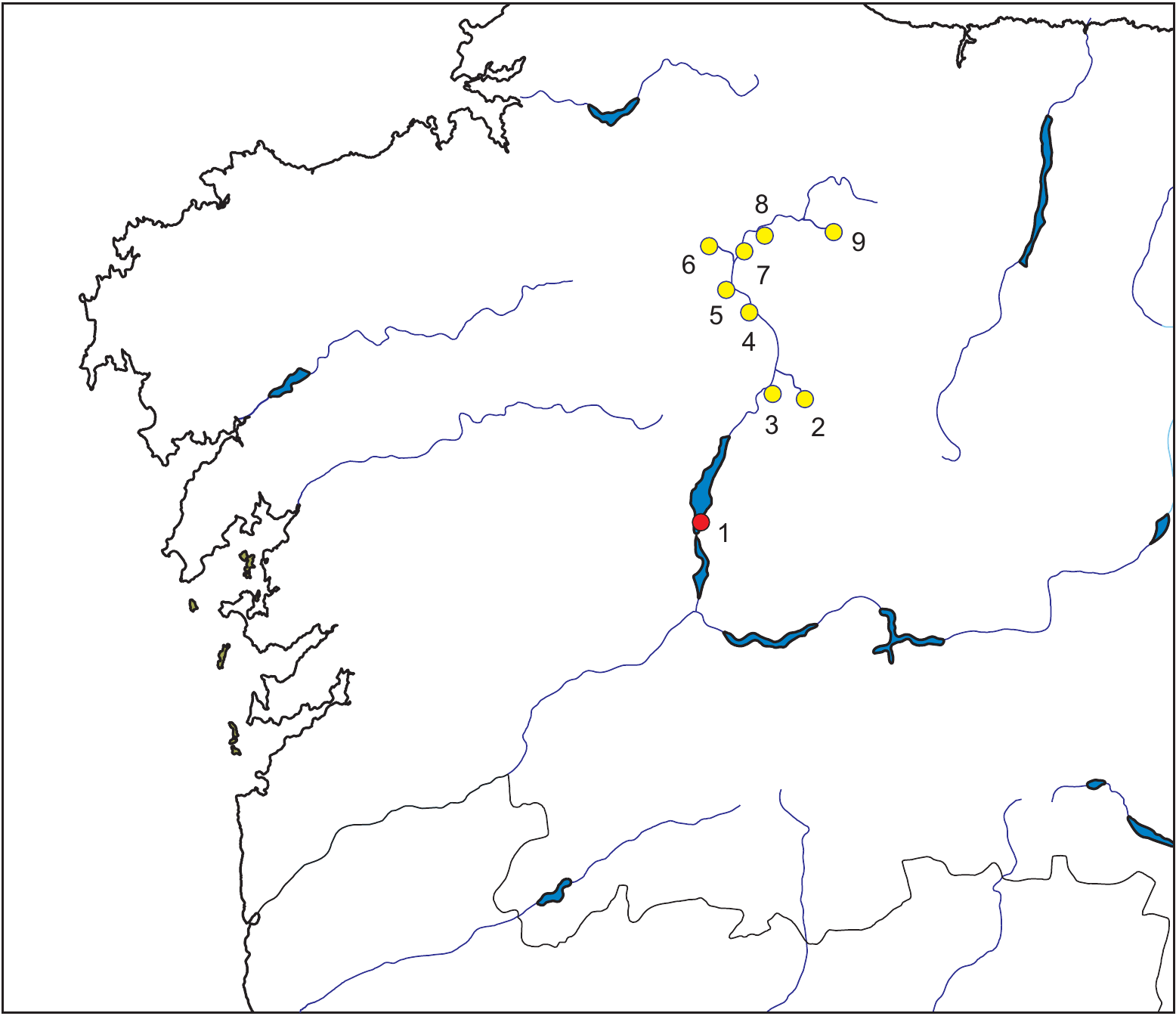}
	\end{center}
	\caption[]{ \label{fig:GeographicalMap} Geographical location of the meteorological stations (blue dots) and the Belesar reservoir (red dot). (1) Belesar reservoir; (2) Sarria river (Pobra de S. Xulian); (3) Neira river (O Paramo); (4) Mi\~no   River (Lugo); (5) Narla river (Gondai); (6) Ladra river (Begonte); (7) Mi\~no   River (Cela); (8) Mi\~no   River (Pontevilar); and (9) Az\'umara river (Reguntille).}
\end{figure}

Belesar is a dammed water reservoir on the Mi\~no River, Galicia, Spain (42.628889$^\circ$ N, 7.7125$^\circ$ W). Its maximum capacity is 655 hm$^3$, with a surface of 1910 ha. Belesar reservoir was created as part of a hydro-power project in 1963, together with a hydro-power station. This reservoir is also used for human consumption in the zone \cite{CHMinhoSil}. Figure \ref{fig:GeographicalMap} shows the reservoir location in Galicia, North-West of Spain. 

\begin{figure}[!ht]
\centering
\includegraphics[width=0.5\columnwidth]{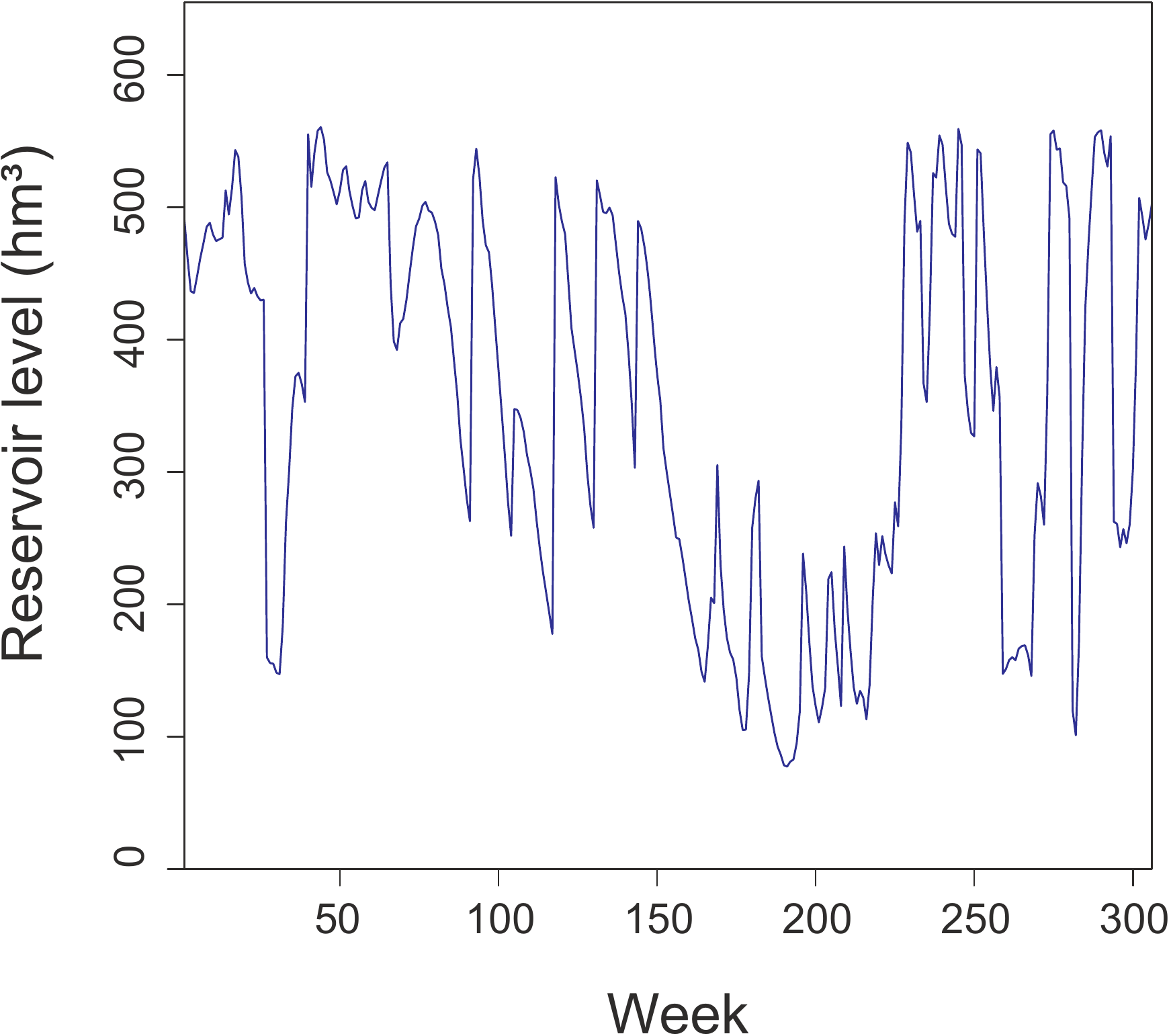}
\caption[]{ \label{fig:WaterLevel} Time series of water level at Belesar reservoir, since 2009.}
\end{figure}

The location of the measuring stations considered in this prediction problem is also shown in Figure \ref{fig:GeographicalMap}. These stations measure the height (m) and flow (m$^3$/s) upstream and on the tributaries, and also the precipitation amount (mm), both of which are taken into account as predictor (input) variables for water level prediction at the dam. The target variable is therefore the dammed water level at Belesar reservoir, measured in hm$^3$. Along with this variable, another one is provided with the dam's data: the amount of water used to generate electricity, measured in m$^3$/s. In addition, we also consider the depth of snow accumulated and the level of snow melt in spring months as input variable. The snow data were obtained from the reanalysis ERA5 model from the Climate Data Store API \cite{SnowData}. We have obtained weekly data of the objective and input variables from 1st January 2009 until 30th September 2015. After the cleaning and pre-processing the data, there were 306 weekly data samples for this prediction problem as shown in \cite{castillo2020analysis}.

Following the methodology in \cite{castillo2020analysis}, four different datasets were created by considering different groups of input variables in the ML regressors. Table \ref{fig:VariablesDataset} shows the different groups of input variables considered. A hold-out scheme was used in all the experiments for short-term analysis and prediction with ML algorithms, with a $80$/$20$ split, where $80\%$ of the data were used to train and $20\%$ for testing.  With this hold-out scheme, a temporal division of the data has been considered in this case, taking the first $80\%$ of the weeks for training and the last $20\%$ for testing the models. A comparison of the randomized-based techniques discussed in this work with alternative regression methods analyzed in \cite{castillo2020analysis} is carried out in the next subsections.

\begin{table}[H]
	\caption{Input variables included in each of the datasets used in the experimental evaluation of the ML regression techniques for short-term analysis and prediction of dammed water level at Belesar.}
\centering
\resizebox{0.9\columnwidth}{!}{
\begin{tabular}{ccc}
			\toprule
			 \textbf{Dataset} & \textbf{Variables Included} & \textbf{Number of Variables} \\
			\midrule
			A & Upstream and tributaries' flow & 19 \\
			\midrule
			B & Upstream and tributaries' flow \& Precipitations & 28 \\
			\midrule
			C & \makecell{Upstream and tributaries' flow,\\Precipitations \& Reservoir output} & 29 \\
			\midrule
			D & \makecell{Upstream and tributaries' flow, \\Precipitations, Reservoir output \& Snow data} & 31 \\
			\bottomrule
		\end{tabular}}
	\label{fig:VariablesDataset}
\end{table}%

\subsubsection{Experimental results}

Table \ref{tab:results_Belesar} shows the results obtained by the different ML randomization-based approaches tested in the four datasets considered at Belesar. As can be seen, the best over all prediction error is provided by the dRVFL algorithm in dataset D (31 input variables), with a MAE of $9.14$ $Hm^3$, an extremely accurate result, which improves over 20\% previous results of classical SVR, MLP and GPR. The second best approach in this problem is the multi-level RVFL, with a best MAE of $10.26$ $Hm^3$, closely followed by other versions of the algorithm such as edRVFL or the basic RVFL. In this problem, other types of randomization-based approaches such as those based on ELM, Adaboost or RF, do not work so well, obtaining results slightly worse than those by the classical ML approaches, and far away from the RVFL family performance. 

\begin{table}[h!]
\caption{\label{tab:results_Belesar} Results obtained by the ML randomization-based approaches and alternative ML algorithms in each of the datasets considered in the prediction of dammed water level.}
\centering
\resizebox{0.85\columnwidth}{!}{
\begin{tabular}{lccccccccccc}
\toprule
& \multicolumn{11}{c}{Datasets} \\
\cmidrule{2-12}
 & \multicolumn{2}{c}{A} & & \multicolumn{2}{c}{B} & & \multicolumn{2}{c}{C} & & \multicolumn{2}{c}{D} \\
\cmidrule{2-3}
\cmidrule{5-6}
\cmidrule{8-9}
\cmidrule{11-12}
& \makecell{RMSE\\ (hm$^3$)} & \makecell{MAE\\ (hm$^3$)} & & \makecell{RMSE\\ (hm$^3$)} & \makecell{MAE\\ (hm$^3$)} & & \makecell{RMSE\\ (hm$^3$)} & \makecell{MAE\\ (hm$^3$)} & & \makecell{RMSE\\ (hm$^3$)} & \makecell{MAE\\ (hm$^3$)} \\
\midrule
SVR (lin) \cite{castillo2020analysis} & 19.96 & 16.18 & & 25.35 & 20.79 & & 16.44 & 11.28 & & 18.60 & 12.74 \\
SVR (rbf) \cite{castillo2020analysis} & 19.34 & 14.55 & & 22.56 & 16.46 & & 19.28 & 12.09 & & 21.88 & 14.25 \\
MLP \cite{castillo2020analysis} & 21.66 & 17.24 & & 23.42 & 17.38 & & 21.12 & 15.17 & & 20.37 & 15.19 \\
GPR \cite{castillo2020analysis} & 21.74 & 18.28 & & 24.43 & 19.28 & & 17.00 & 11.99 & & 18.65 & 13.40 \\
\midrule 
RFR & 27.77 & 20.81 & & 25.14 & 18.90 & & 25.60 & 19.80 & & 23.42 & 18.28 \\
AdaBoost & 26.09 & 20.96 & & 24.60 & 19.61 & & 24.87 & 19.57 & & 24.08 & 18.95 \\
ELM \cite{castillo2020analysis} & 24.27 & 17.48 & & 24.86 & 18.31 & & 21.59 & 14.56 & & 22.87 & 15.10 \\
ML-ELM & 17.80 & 13.51 & & 19.11 & 15.00 & & 17.93 & 13.18 & & 14.64 & 12.02 \\
RVFL & 18.16 & 14.36 & & 20.18 & 15.55 & & 15.11 & 10.98 & & 13.51 & 10.45 \\
multiRVFL & 16.89 & 12.71 & & 19.64 & 14.96 & & 15.07 & 10.38 & & 13.59 & 10.26 \\
dRVFL & 17.40 & 13.16 & & 18.87 & 14.53 & & 14.45 & 9.93 & & 11.06 & 9.14 \\
edRVFL & 17.69 & 13.49 & & 19.65 & 15.01 & & 14.58 & 10.03 & & 13.66 & 10.36 \\
\bottomrule
\end{tabular}}
\end{table}

Table \ref{Wilcoxon_Belesar}  shows the Wilcoxon signed rank p-values for the different datasets in the water level prediction problem at Belesar. As can be seen, somo algorithms of the RFVL family works similar, but in the majority of cases the test confirm statistically the superiority of the RVFL family approaches over the rest of algorithms tested in this problem.

\begin{table}[]
\caption{\label{Wilcoxon_Belesar} Wilcoxon signed rank p-values for the different datasets in the water level prediction problem at Belesar. Cells shaded in gray correspond to those cases where statistical significance cannot be declared at $\alpha=0.05$.}
\centering
\resizebox{\columnwidth}{!}{
\begin{tabular}{ccccccccc}
\toprule
p-value &
RFR & AdaBoost & ELM & ML-ELM & RVFL & multiRVFL & dRVFL\\
\midrule
\multicolumn{9}{c}{Dataset A}\\
\midrule
AdaBoost & 5.792e-05 & --- & --- & --- & --- & --- & --- & \\
ELM & 1.734e-06 & 1.734e-06 & --- & --- & --- & --- & --- & \\
ML-ELM & 1.734e-06 & 1.734e-06 & \cellcolor[gray]{0.8}1.109e-01 & --- & --- & --- & --- & \\
RVFL & 1.734e-06 & 1.734e-06 & \cellcolor[gray]{0.8}5.304e-01 & 4.950e-02 & --- & --- & --- & \\
multiRVFL & 1.734e-06 & 1.734e-06 & 1.175e-02 & 2.105e-03 & 2.849e-02 & --- & --- & \\
dRVFL & 1.734e-06 & 1.734e-06 & 1.477e-04 & 6.339e-06 & 3.317e-04 & \cellcolor[gray]{0.8}1.306e-01 & --- & \\
edRVFL & 1.734e-06 & 1.734e-06 & 2.353e-06 & 2.879e-06 & 4.286e-06 & 4.449e-05 & \cellcolor[gray]{0.8}6.268e-02 & \\
\midrule
\multicolumn{9}{c}{Dataset B}\\
\midrule
AdaBoost & \cellcolor[gray]{0.8}3.086e-01 & --- & --- & --- & --- & --- & --- & \\
ELM & 1.734e-06 & 1.734e-06 & --- & --- & --- & --- & --- & \\
ML-ELM & \cellcolor[gray]{0.8}2.802e-01 & \cellcolor[gray]{0.8}1.915e-01 & 4.286e-06 & --- & --- & --- & --- & \\
RVFL & 1.734e-06 & 1.734e-06 & 1.319e-02 & 2.127e-06 & --- & --- & --- & \\
multiRVFL & 1.734e-06 & 1.734e-06 & \cellcolor[gray]{0.8}3.493e-01 & 3.182e-06 & 1.852e-02 & --- & --- & \\
dRVFL & 1.734e-06 & 1.734e-06 & 4.277e-02 & 2.353e-06 & \cellcolor[gray]{0.8}5.440e-01 & 3.872e-02 & --- & \\
edRVFL & 1.734e-06 & 1.734e-06 & 7.157e-04 & 2.603e-06 & \cellcolor[gray]{0.8}4.405e-01 & 9.842e-03 & \cellcolor[gray]{0.8}8.774e-01 & \\
\midrule
\multicolumn{9}{c}{Dataset C}\\
\midrule
AdaBoost & 2.597e-05 & --- & --- & --- & --- & --- & --- & \\
ELM & 1.734e-06 & 1.734e-06 & --- & --- & --- & --- & --- & \\
ML-ELM & 1.477e-04 & \cellcolor[gray]{0.8}1.915e-01 & 1.734e-06 & --- & --- & --- & --- & \\
RVFL & 1.734e-06 & 1.734e-06 & 6.892e-05 & 1.734e-06 & --- & --- & --- & \\
multiRVFL & 1.734e-06 & 1.734e-06 & \cellcolor[gray]{0.8}6.871e-02 & 1.921e-06 & 2.225e-04 & --- & --- & \\
dRVFL & 1.734e-06 & 1.734e-06 & 1.245e-02 & 1.734e-06 & 1.319e-02 & \cellcolor[gray]{0.8}1.204e-01 & --- & \\
edRVFL & 1.734e-06 & 1.734e-06 & 8.919e-05 & 1.734e-06 & \cellcolor[gray]{0.8}3.820e-01 & 1.114e-03 & 5.706e-04 & \\
\midrule
\multicolumn{9}{c}{Dataset D}\\
\midrule
AdaBoost & \cellcolor[gray]{0.8}1.779e-01 & --- & --- & --- & --- & --- & --- & \\
ELM & 1.734e-06 & 1.734e-06 & --- & --- & --- & --- & --- & \\
ML-ELM & 8.466e-06 & 1.639e-05 & 3.182e-06 & --- & --- & --- & --- & \\
RVFL & 1.734e-06 & 1.734e-06 & 4.196e-04 & 1.921e-06 & --- & --- & --- & \\
multiRVFL & 1.734e-06 & 1.734e-06 & \cellcolor[gray]{0.8}7.036e-01 & 2.879e-06 & 6.320e-05 & --- & --- & \\
dRVFL & 1.734e-06 & 1.734e-06 & \cellcolor[gray]{0.8}6.268e-02 & 1.734e-06 & 1.965e-03 & 8.730e-03 & --- & \\
edRVFL & 1.734e-06 & 1.734e-06 & 1.245e-02 & 1.734e-06 & 3.379e-03 & 1.382e-03 & \cellcolor[gray]{0.8}2.536e-01 & \\
\bottomrule
\end{tabular}}
\end{table}

\begin{figure}[!ht]
\centering
\includegraphics[width=\columnwidth]{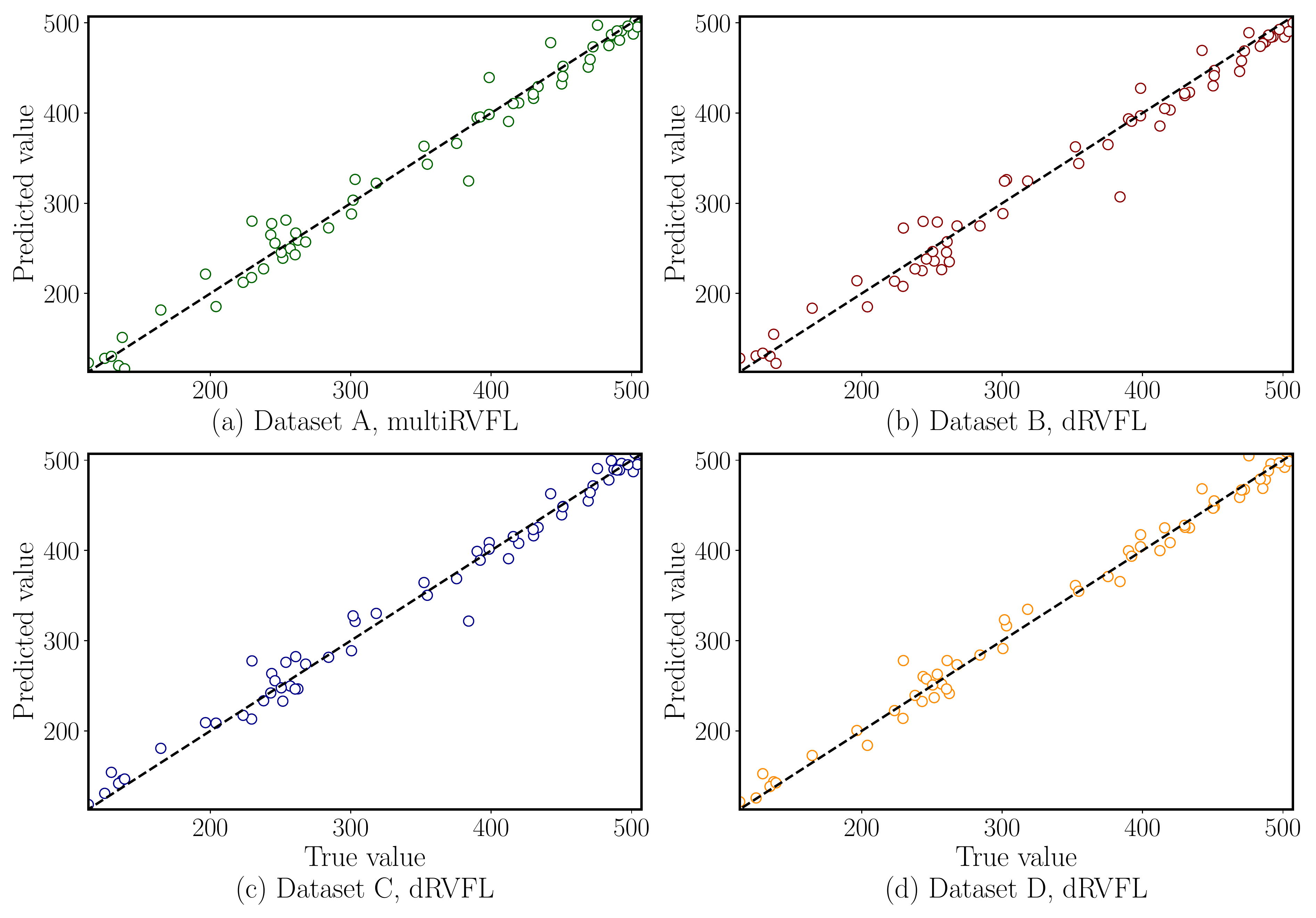}
\caption{\label{fig:scatterWater} Scatter plots of true versus predicted values corresponding to the best RMSE models and the 4 datasets considered in the third case of study.}
\end{figure}

Figure \ref{fig:scatterWater} shows the scatter plots (true vs. predicted values) for the RMSE of the four datasets considered. As can be seen, in the first dataset (19 input variables), the best performance corresponds to the multiRVFL, and in the rest of datasets (28, 29 and 31 input variables, respectively), the dRVFL is the best algorithm, showing an excellent performance, as can be seen in scatter plots depicted. Note that the best overall performance is obtained by the dRVFL algorithm in the dataset D (31 input variables).

\section{Conclusions, challenges and research directions}\label{Challenges}

Randomization-based ML algorithms are currently a hot topic in Artificial Intelligence, due to their excellent performance in prediction problems, both in terms of classification or regression approaches. Furthermore, these learning algorithms offer a bounded computation time, what makes them perfect for constructing hybrid and ensemble-based approaches. In the last years, there has been a boom in the application of these approaches to renewable energy prediction problems. In this manuscript we have discussed the most important randomization-based ML approaches, delving into their main algorithmic properties. We have also revised the most important works in the literature, focusing on recent contributions that amount to the large majority published in the last five years. Many different studies in solar, wind, marine/ocean and hydro-power have been addressed in the bibliographic review carried out in our study. We have also provided experimental evidence of the good performance on some randomization-based ML approaches, over different real prediction problems, namely, solar radiation estimation, wind speed prediction and water level prediction for hydro-power. In all the prediction problems tackled in our experiments, randomization-based ML algorithms have elicited a superior performance, improving the results of alternative existing algorithms in the literature with statistical significance in most cases. Remarkably, training latencies were in the order of a few seconds for most cases, reducing dramatically the time required by other models in the benchmarks under consideration.

Despite the evidence of their superior modeling capability and training efficiency showcased in our study, randomization based models is a research field that enjoys a vibrant momentum in the research community. Indeed, many challenges remain still open by the community, part of which permeate to their application to renewable energy problems. We briefly pause at these challenges (depicted schematically in Figure \ref{fig:challenges}), outlining research opportunities and directions stemming from each of them:
\begin{itemize}[leftmargin=*]
\item On the interplay between randomness and uncertainty: as was previously highlighted in our critical analysis of the literature (Subsection \ref{critical_literature_analysis}), the way randomization based machine learning models are trained suggests that the output prediction of these models should be subject to a degree of epistemic uncertainty \cite{hullermeier2019aleatoric,der2009aleatory}, thereby hindering the trustworthiness of practical solutions embracing them at their core. Indeed, the user consuming the model's output could eventually ask him/herself to which extent the predicted output can be trusted given that the knowledge captured by the model in use is partly governed at random. This concern is particularly relevant when modeling sequences via Echo State Networks, as the recurrence through which the input sequence is processed can catastrophically amplify the effect of small initialization differences of their random reservoir weights in the output of the model. Therefore, more studies are needed to verify the circumstances under which the effect of randomness in the model's output is counteracted by e.g. robust readout layers, an oversized set of random neural features, or any other modeling choice alike. 
\begin{figure}[h]
\begin{center}
\includegraphics[width=\columnwidth]{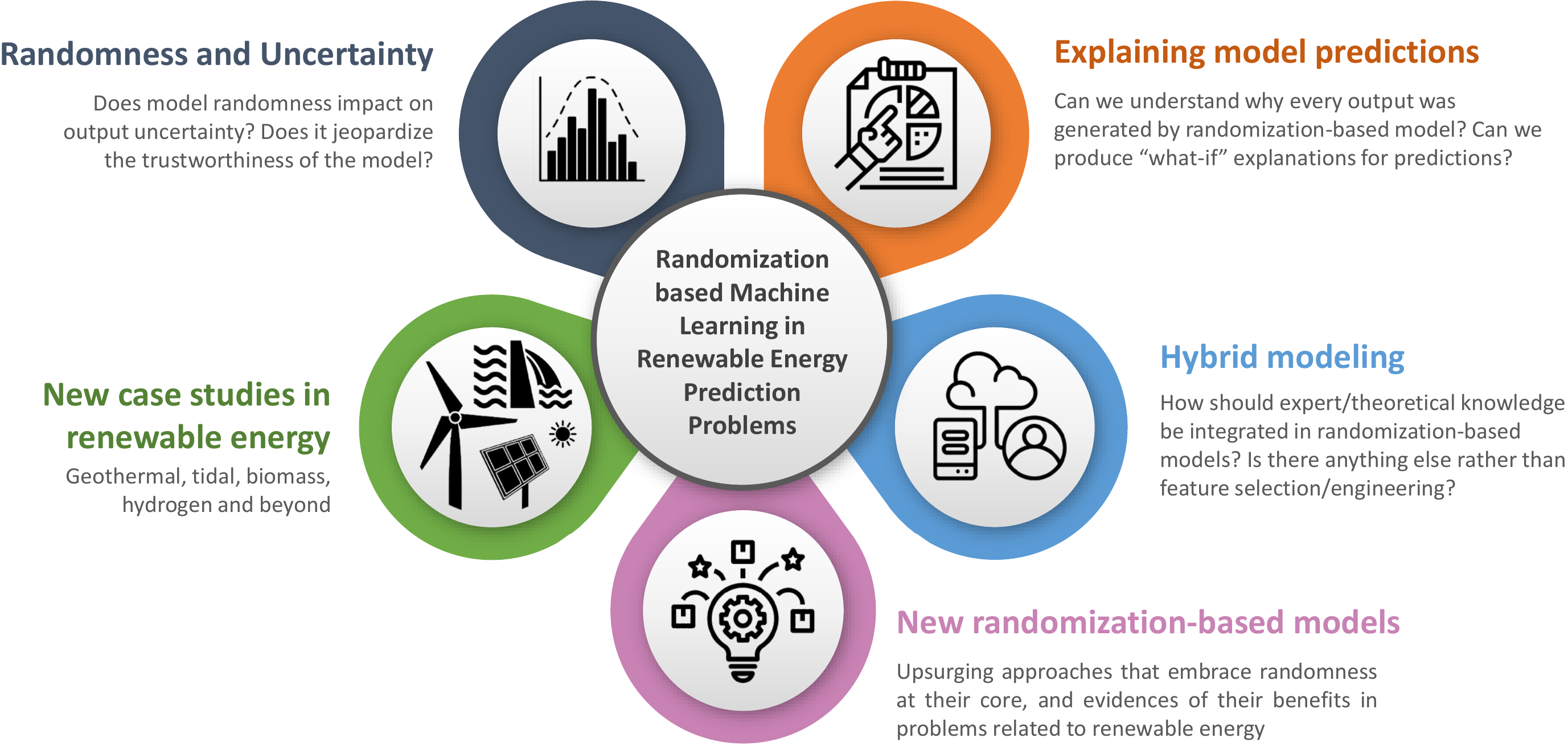}
\end{center}
\caption{Challenges and open questions related to randomization-based machine learning models applied to renewable energy prediction problems.}
\label{fig:challenges}
\end{figure}

\item Explainability of randomization based machine learning models: in partial connection to the above and our prior insights in Subsection \ref{critical_literature_analysis}, the need for explaining predictions also arises when dealing with randomization based models. It is not only their randomness what clashes with the acceptability of the prediction issued by such models by the end user: the lack of understanding why an output was produced by a model also impedes that it is trusted, assumed and used in practical settings. Unlike other methods not relying on randomization (e.g. K-nearest neighbors or decision trees), most randomization based models fall within the category of machine learning algorithms whose lack of transparency demand for explanations \cite{adadi2018peeking,arrieta2020explainable}. This is also the case of ensembles where, even if composed by human-readable tree models, their composition and aggregation to form the ensemble yields a computational structure difficult to be assimilated by non-expert audience. This has motivated several studies dealing with the interpretability of bagging and boosting ensembles \cite{lundberg2019explainable,sagi2020explainable}, but attempts at explaining modern forms of randomization-based models are still in their infancy \cite{arrieta2021post}. When conceived within the renewable energy domain, explainability comprises not only feature attribution studies, but also techniques that allow unveiling which plausible modifications of the input variables can influence most in the output. Counterfactual generation can be useful to understand e.g. which input features can give rise to wind speed regimes that can damage the structural integrity of wind turbines. The use of eXplainable Artificial Intelligence (XAI) techniques can definitely bolster the adoption of randomization-based models in practical renewable energy prediction systems and processes.

\item Hybrid modeling: while setting part of the model at random is the essence of the models under our focus, much theoretical knowledge has been collected over decades from the study of the complex phenomena participating in renewable energy scenarios. For instance, wind generation has been largely approached by resorting to computational fluid dynamics, paving the way to extended knowledge about the variables that most significantly control the wind regimes at different scales. We advocate for the incorporation of such background knowledge in data-based models as the ones investigated in this study. A number of assorted strategies can be designed for that purpose: from the most straightforward one (engineered features for the model that find their rationale in known facts and results from previous studies) to more elaborated ways (correspondingly, biasing the random initialization as a function of such knowledge, or incorporating the output of numerical simulators as an additional input to the randomization based model). In other words, rephrasing Epicurus' Principle of Multiple Explanations: ``if several theories are consistent with the observed data, retain them all''. 

\item New models based on randomization: the portfolio of models reviewed in this survey is a representative set of the distinct strategies by which data-based models can benefit from randomness: ensembles, wherein randomness emerges from the bootstraps fed to every learner, and randomization based neural networks, where a diversity of random features is extracted by the neural part, and later mapped to the target variable via a readout layer. Notwithstanding this intended representativeness of the models considered in our survey, many other proposals are reported nowadays, in which randomness again appears as one of their foundational algorithmic ingredients. To cite a few, oblique random forests \cite{menze2011oblique,katuwal2020heterogeneous} generalize the splitting criterion of the trees within the ensemble to multiple variables, hence producing an oblique hyperplane instead of the axis-parallel cut that characterizes univariate splitting rules. On the other hand, isolation forests extend the aggregation of bootstrapped trees to the task of detecting outliers in data \cite{liu2008isolation}. Furthermore, the usage of random features and random tensor products is at the heart of kernel methods when seeking to use them over large-scale datasets \cite{rahimi2007random}. Even brand new concepts and studies in Deep Learning such as network overparametrization \cite{allen2018learning}, random feature mappings \cite{saxe2011random} and self-distillation \cite{allen2020towards} are linked to the effect of randomization through the initialization of the neural weights or by assembling models learned from random unstructured inputs. There is a vast area of opportunity for the community working in renewable energy prediction problems to ascertain the potential of these newly formulated alternatives.

\item More renewable energy case studies: the heterogeneous set of scenarios comprising the experimental part of this survey has exposed quantitatively the outperforming behavior of randomization-based models with respect to other models in use by the community, specially in scenarios subject to long-term persistence. However, the need for data-based models is rising sharply for all sources of renewable energy, mainly by virtue of the increasing digitization of their assets, systems and processes. Therefore, it is our firm belief that randomization based machine learning models can also be massively used in scenarios involving other renewable energy sources, including tidal, hydrogen, geothermal and biomass energy. We foresee that the inherent persistence and seasonality of these complex systems unleash a magnificent opportunity for randomization based machine learning, which can contribute to the environmental sustainability of data-based solutions for renewable energy prediction to levels out of the reach of other modeling counterparts (e.g. deep neural networks). 
\end{itemize}

On a closing note, we accord with critical voices in the community claiming that predictive modeling should also consider design factors beyond the accuracy/precision of their outputs. Randomization based models step aside the current research mainstream, in which overly complex models are proposed to achieve negligible performance gains of questionable practical value. Other aspects such as power consumption, memory requirements, training time or inference latency are decisive for realizing the benefits of predictive models in many practical scenarios. Renewable energy is in no way an exception to this statement. We hope that this survey enshrines as a reference material for researchers and practitioners willing to examine renewable energy prediction problems through the lenses of performance, efficiency and environmental sustainability. 

\section*{Acknowledgments}
This research has been partially supported by the Ministerio de Econom\'{i}a y Competitividad of Spain (Grant Ref. TIN2017-85887-C2-2-P). This research has also been partially supported by Comunidad de Madrid, PROMINT-CM project (grant ref: P2018/EMT-4366). J. Del Ser would like to thank the Basque Government for its funding support through the EMAITEK and ELKARTEK programs (3KIA project, KK-2020/00049), as well as the consolidated research group MATHMODE (ref. T1294-19). 

\bibliographystyle{model1-num-names}
\bibliography{sample}

\end{document}